\documentclass{article}

\usepackage{microtype}
\usepackage{graphicx}
\usepackage{subfigure}
\usepackage{booktabs}

\usepackage{hyperref}

\usepackage[accepted]{icml2024}

\usepackage{amsmath}
\usepackage{amssymb}
\usepackage{mathtools}
\usepackage{amsthm}

\usepackage[capitalize,noabbrev]{cleveref}

\theoremstyle{plain}

\theoremstyle{definition}

\theoremstyle{remark}

\usepackage[textsize=tiny]{todonotes}

\usepackage{csquotes}
\usepackage{tikz}
\usepackage{listings}
\usepackage{xcolor,soul}

\newcommand\rightangle{%
  \mathord{%
    \mspace{1mu}%
    \text{\radot}%
    \mspace{1mu}%
  }%
}

\newcommand\radot{%
    \tikz[line cap=round,x=1ex,y=1ex,line width=0.3pt]
    {\draw (0,1) |- (1,0) (0.55,0) arc(0:90:0.55); \fill (0.23,0.23) circle (0.05);}%
}

\colorlet{punct}{red!60!black}
\definecolor{background}{HTML}{EEEEEE}
\definecolor{delim}{RGB}{20,105,176}
\colorlet{keyword}{magenta!60!black}

\lstdefinelanguage{lean}{
    basicstyle=\footnotesize\ttfamily,
    numbers=left,
    numberstyle=\scriptsize,
    stepnumber=1,
    numbersep=8pt,
    showstringspaces=false,
    breaklines=true,
    frame=lines,
    backgroundcolor=\color{background},
    literate=
     *{axiom}{{{\color{keyword}axiom}}}{5}
      {theorem}{{{\color{keyword}theorem}}}{7}
      {lemma}{{{\color{keyword}lemma}}}{5}
      {by}{{{\color{keyword}by}}}{2}
      {Lemma}{{{\color{keyword}Lemma}}}{5}
      {Proof}{{{\color{keyword}Proof}}}{5}
      {Qed}{{{\color{keyword}Qed}}}{3}
      {:}{{{\color{punct}{:}}}}{1}
      {,}{{{\color{punct}{,}}}}{1}
      {\{}{{{\color{delim}{\{}}}}{1}
      {\}}{{{\color{delim}{\}}}}}{1}
      {[}{{{\color{delim}{[}}}}{1}
      {]}{{{\color{delim}{]}}}}{1},
}

\newcommand{\smallsec}[1]{\paragraph{#1.}}

\newcommand{\toolName}{\ensuremath{\mathsf{E}{\tt {3}}}}
\newcommand\dataname{LeanEuclid}

\definecolor{greenforassumption}{RGB}{226, 240, 217}
\definecolor{blueforgoal}{RGB}{222, 235, 247}
\definecolor{orangeforrule}{RGB}{251, 229, 214}

\DeclareRobustCommand{\hlorange}[1]{{\sethlcolor{orangeforrule}\hl{#1}}}

\begin{document}

\twocolumn[
\icmltitle{Autoformalizing Euclidean Geometry}

\icmlsetsymbol{equal}{*}

\begin{icmlauthorlist}
\icmlauthor{Logan Murphy}{uoft,equal}
\icmlauthor{Kaiyu Yang}{caltech,equal}
\icmlauthor{Jialiang Sun}{uoft}
\icmlauthor{Zhaoyu Li}{uoft}
\icmlauthor{Anima Anandkumar}{caltech}
\icmlauthor{Xujie Si}{uoft}
\end{icmlauthorlist}

\icmlaffiliation{caltech}{Caltech}
\icmlaffiliation{uoft}{University of Toronto}

\icmlcorrespondingauthor{Logan Murphy}{lmurphy@cs.toronto.edu}
\icmlcorrespondingauthor{Kaiyu Yang}{kaiyuy@caltech.edu}
\icmlcorrespondingauthor{Xujie Si}{six@cs.toronto.edu}

\icmlkeywords{Euclidean Geometry, Autoformalization, Theorem Proving, Machine Learning, ICML}

\vskip 0.3in
]

\printAffiliationsAndNotice{\icmlEqualContribution}

\newcommand{\lm}[1]{\textcolor{orange}{LM: #1}}

\newcommand{\zl}[1]{\textcolor{green}{zl: #1}}
\newcommand{\js}[1]{\textcolor{red}{JS: #1}}
\newcommand{\aacomment}[1]{\textcolor{red}{{\bf Anima:} #1}}

\begin{abstract}
Autoformalization involves automatically translating informal math into formal theorems and proofs that are machine-verifiable. Euclidean geometry provides an interesting and controllable domain for studying autoformalization. In this paper, we introduce a neuro-symbolic framework for autoformalizing Euclidean geometry, which combines domain knowledge, SMT solvers, and large language models (LLMs). One challenge in Euclidean geometry is that informal proofs rely on diagrams, leaving gaps in texts that are hard to formalize. To address this issue, we use theorem provers to fill in such diagrammatic information automatically, so that the LLM only needs to autoformalize the explicit textual steps, making it easier for the model. We also provide automatic semantic evaluation for autoformalized theorem statements. We construct LeanEuclid, an autoformalization benchmark consisting of problems from Euclid's \emph{Elements} and the UniGeo dataset formalized in the Lean proof assistant. Experiments with GPT-4 and GPT-4V show the capability and limitations of state-of-the-art LLMs on autoformalizing geometry problems. The data and code are available at \url{https://github.com/loganrjmurphy/LeanEuclid}.
\end{abstract}

\section{Introduction}

\begin{figure*}[ht]
  \centering
  \includegraphics[width=1.0\linewidth]{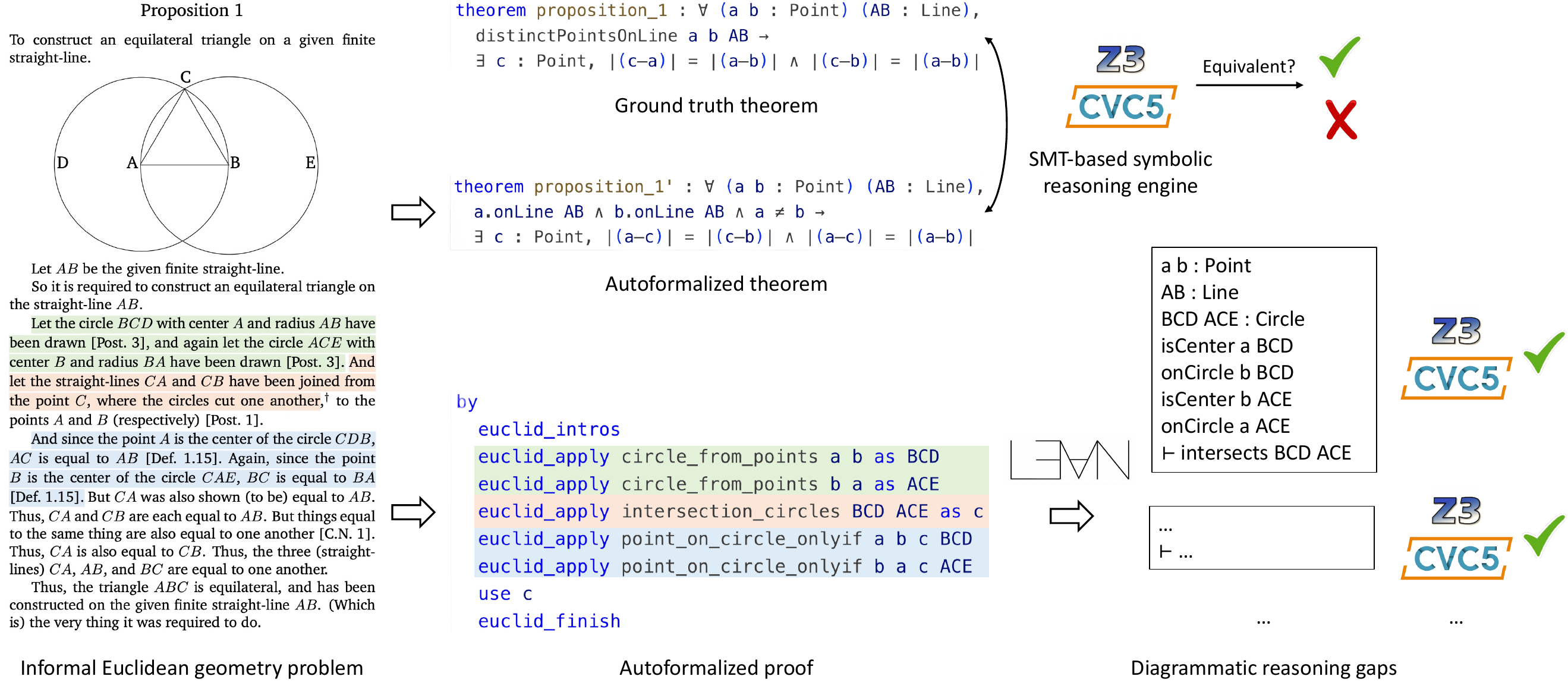}
  \caption{\textbf{Left}: Proposition 1 in Euclid's \emph{Elements} (Book I). The \hlorange{orange text} involves diagrammatic reasoning: Euclid did not explicitly prove the two circles actually intersect, but the reader can use the diagram to implicitly fill in the logical gap.
  \textbf{Top right}: The model autoformalizes the problem into a formal theorem (\texttt{proposition\_1'}), which is evaluated by checking its logical equivalence with the ground truth (\texttt{proposition\_1}), leveraging domain knowledge and a symbolic automated reasoning engine based on SMT (satisfiability modulo theories) solvers.
  \textbf{Bottom right}: A proof autoformalized by the model. Like Euclid's proofs, it does not need to handle diagrammatic reasoning explicitly. Lean can check the proof to identify a list of diagrammatic reasoning gaps, e.g., ``\texttt{intersects BCD ACE}''. Then, it attempts to fill in all gaps automatically using the symbolic reasoning engine based on SMT solvers.
  }
  \label{fig:overall}
\end{figure*}

Euclidean geometry is one of the oldest branches of mathematics. It has served as a test of human intelligence for more than two millennia and has recently been used to test AI. Substantial work has focused on solving geometry problems automatically~\citep{wu2008decision}, e.g., AlphaGeometry~\citep{trinh2023solving} can solve some of the IMO geometry problems. These methods consume problems and produce solutions in structured formats. In this work, we address a complementary task, \emph{autoformalization}: Can AI understand human-written problems/solutions and translate them automatically into formal theorems/proofs? Specifically, we focus on the setting where formal theorems/proofs can be verified by the Lean proof assistant~\citep{moura2021lean}. Lean provides a language for writing formal proofs. It is popular among mathematicians and has a growing ecosystem of integration with large language models (LLMs), e.g., LeanDojo~\citep{yang2023leandojo} and Lean Copilot~\citep{song2024towards}.

We demonstrate that Euclidean geometry provides an interesting and controllable domain for autoformalization. First, an automatic evaluation of autoformalized theorems is difficult in general but feasible in Euclidean geometry. Second, the logical gaps in informal proofs are well understood in Euclidean geometry, making it easier to faithfully formalize the proofs. Third, combining text-based and diagrammatic reasoning makes Euclidean geometry a natural domain to study multimodal reasoning models. Therefore, autoformalizing Euclidean geometry is an attractive target for AI.

\smallsec{Evaluating Autoformalized Theorem Statements}
Despite the promise of machine learning and LLMs in autoformalizing theorems~\cite{wu2022autoformalization}, a major roadblock is the lack of reliable and automatic evaluation.  Comparing the model output verbatim with ground truth would be too rigid since there are many valid ways to formalize a theorem. Checking the logical equivalence between two theorems is generally intractable. Researchers have resorted to proxy metrics such as the BLEU score~\cite{papineni2002bleu}. However, LLMs can score high on such metrics without generating correct formalization~\citep{jiang2023multilingual}. Alternatively, human evaluation is widely used as a last resort, but it is costly, especially if we want to use the results to improve the method iteratively.

\smallsec{Our Approach to Evaluating Autoformalization} To overcome the evaluation bottleneck, we introduce a new automatic approach for evaluating the semantics of autoformalized theorems. The key insight is that equivalence checking can be made feasible in specific domains (such as Euclidean geometry) by combining domain knowledge with automated reasoning tools, such as satisfiability modulo theories (SMT) solvers~\citep{barrett2018satisfiability}.

To evaluate the autoformalized theorems, we develop a symbolic reasoning engine based on SMT solvers. As Fig.~\ref{fig:overall} (\emph{Top right}) shows, given a ground-truth formal theorem $T_{gt}$ and the autoformalized theorem $T_{pred}$ produced by a language model, we use the symbolic engine to try to prove their equivalence ($T_{gt} \Leftrightarrow T_{pred}$). If successful, their logical gap is small enough to conclude that $T_{pred}$ is correct. Even if the symbolic engine cannot prove $T_{gt} \Leftrightarrow T_{pred}$, it can provide partial results useful for a more fine-grained analysis. We validate this evaluation protocol by showing it correlates well with human evaluation.

\smallsec{\dataname: Formalizing Proofs and Diagrams}
We construct \emph{\dataname}, a benchmark for testing machine learning on autoformalizing Euclidean geometry. As in Fig~\ref{fig:overall} (\emph{Left}), each example in {\dataname } has an informal theorem, proof, and diagram in \LaTeX, as well as a formal theorem and proof in Lean. Data examples in {\dataname } are manually formalized into Lean from Euclid's \emph{Elements}~\citep{heiberg2007euclid} and the UniGeo dataset~\cite{chen2022unigeo}.

{\dataname } serves as a benchmark for autoformalizing not only theorems but also proofs. Geometric proofs are challenging to formalize faithfully. Humans (ancient or modern, including Euclid himself) use diagrams to license proof steps without making every detail explicit. Fig.~\ref{fig:overall} shows an example of diagrammatic reasoning from Euclid's \emph{Elements}. Euclid uses the intersection of two circles ($C$) without proving its existence. Most readers would not find the proof problematic, as the two circles intersect in the diagram. Such implicit diagrammatic reasoning is ubiquitous in informal geometric proofs but needs to be handled explicitly in formal proofs~\cite {beeson2019proof}. Therefore, a naive attempt to autoformalize the proofs would be difficult, as it requires the model to fill in many diagrammatic reasoning gaps, with nothing to reference in the informal texts.

To mitigate diagrammatic gaps, {\dataname} adopts a formal system named E~\citep{avigad2009formal}, introduced by philosophers for modeling diagrammatic reasoning in Euclid's \emph{Elements}. It teases out a set of diagrammatic rules so that diagrammatic reasoning can be modeled as logical deductions. We implement E in Lean and provide proof automation to fill in diagrammatic reasoning gaps, using the same symbolic reasoning engine developed for equivalence checking. Our system enables formalizing all 48 theorems and proofs from \emph{Elements} (Book I), following Euclid's original proofs as closely as possible, with diagrammatic reasoning carried out implicitly and automatically (see Fig.~\ref{fig:overall}). The data is included in {\dataname}, making autoformalizing Euclid's proofs feasible. The language model now only needs to autoformalize the explicit textual proof steps, leaving the ``obvious'' implicit reasoning to the symbolic engine.

\smallsec{LLMs for Autoformalization}
{
Using {\dataname}, we experimentally investigate the capability of GPT-4~\citep{openai2023gpt4} and GPT-4V (with vision) on autoformalizing theorems and proofs in Euclidean geometry. With 5-shot prompting, GPT-4 can autoformalize only 18.9\% of the theorems correctly, demonstrating the difficulty of {\dataname}. GPT-4V performs slightly better (21.0\%), showing the potential benefit of multimodal models. Furthermore, GPT-4V can autoformalize 23.1\% of {\dataname} proofs correctly, generating the entire proof without searching the space of individual steps. Finally, our study shows that imperfect autoformalized proofs can often be manually repaired into correct proofs through a few simple modifications.}

\smallsec{Contributions}
We make three main contributions: First, we introduce an SMT-based symbolic engine for automatically evaluating the \emph{semantics} of theorem statements in Euclidean geometry. Second, we demonstrate the feasibility of autoformalizing only explicit geometric proof steps, relying on the symbolic engine for implicit diagrammatic reasoning. Third, our experiments reveal the capability and limitations of state-of-the-art LLMs on autoformalizing Euclidean geometry. Overall, our approach demonstrates Euclidean geometry as a promising target for autoformalization. It may facilitate autoformalization research to move from human inspection to automatic evaluation. Such a transition is essential for accelerating the development of machine learning methods on this task and ultimately easing the human effort required by formalization.

\section{Related Work}
\label{sec:related}

\smallsec{Neural Theorem Proving and Autoformalization}
Machine learning and LLMs can prove theorems using proof assistants~\citep{polu2020generative,yang2019learning,yang2023leandojo,first2023baldur,lample2022hypertree,han2022proof}. In theorem proving, everything is formal: The model generates a formal proof given a formal theorem. In contrast, we focus on autoformalization, which aims to translate informal math into formal theorems and proofs. \citet{li2024dl4tp} provides a comprehensive survey on deep learning for theorem proving and autoformalization.

Machine learning for autoformalization is strained by the lack of informal-formal pairs for training. Manually annotating such a dataset would be prohibitively expensive. Researchers have explored two approaches to overcome data scarcity. First, we can generate synthetic training data either procedurally~\citep{wang2018first,cunningham2023towards} or using LLMs~\citep{chen2023nl2tl,jiang2023multilingual}. Second, we can leverage the few-shot capability of LLMs such as GPT-4~\citep{openai2023gpt4}. We only need to annotate a few examples for in-context demonstration and a relatively small testing set~\citep{wu2022autoformalization,agrawal2022towards,jiang2023draft,azerbayev2023proofnet}. {\dataname } follows the latter approach and is intended as a benchmark for testing only.

Autoformalized theorems are difficult to evaluate, so all existing works have resorted to human evaluation. In addition, they have used proxy metrics such as compilation rates~\cite{jiang2023multilingual,azerbayev2023proofnet} and the BLEU score~\cite{wu2022autoformalization,azerbayev2023proofnet}, none of which is sufficiently accurate. In contrast, we are the first to evaluate the semantics of autoformalized statements reliably and automatically.

\smallsec{Formalizing Euclidean Geometry}
To our knowledge, no machine learning method has attempted to autoformalize Euclidean geometry. Tarski's~\citep{schwabhauser2013metamathematische} and Hilbert's axioms~\citep{hilbert2013grundlagen} are two classical axiomatic systems that put Euclidean geometry on a rigorous foundation. \citet{beeson2019proof} uses a variant of Tarski's system to manually formalize Euclid's \emph{Elements} (Book I) in the Coq proof assistant~\cite{barras1997coq}. \citet{hernandezespiet2023e} is an ongoing work formalizing the book in Lean. However, they require formal proofs to explicitly handle all gaps left by diagrammatic reasoning. As a result, formal proofs in their systems are much more verbose than Euclid's proofs (examples in Appendix~\ref{appendix:comparison}), making their systems unsuitable for studying autoformalization.

Proofs in Euclidean geometry have gaps due to diagrammatic reasoning. However, for more than two millennia, they have seldom troubled \emph{Elements}' readers around the world. Philosophers have posited many theories aiming at a systematic understanding of diagrammatic reasoning~\citep{miller2001diagrammatic,mumma2010proofs}. In particular, we build upon the formal system E~\citep{avigad2009formal} and are the first to turn this framework into a practical system that includes an automated procedure for diagrammatic reasoning.

\smallsec{Geometric Problem Solving}
Automatic geometry problem solvers have been studied extensively~\citep{nevins1975plane,zhang1995automated}. Wu's method~\citep{wu2008decision} and  Gr{\"o}bner bases method~\citep{buchberger2005applications} transform geometric problems into algebraic equations. AlphaGeometry~\citep{trinh2023solving} is a concurrent work that successfully solves many IMO geometry problems. Similar to us, it is a hybrid system using neural networks for difficult proof steps (e.g., constructing new geometric objects) and a symbolic engine for deducing simple facts. However, our method focuses on autoformalizing human-written theorems/proofs instead of finding new proofs. AlphaGeometry consumes problems and produces solutions in structured representations, without handling informal mathematical language. Furthermore, they represent theorems and proofs in a domain-specific language tailored for Euclidean geometry, whereas we use Lean: a general-purpose proof assistant. Lean provides a highly trusted foundation for the soundness of our system,\footnote{Our symbolic reasoning engine (Sec.~\ref{subsec:smt}) relies on SMT solvers, whose results currently cannot be fully certified by Lean.} and it opens up opportunities to interoperate with other branches of mathematics formalized in Lean's mathematical library~\citep{mathlib}.

Euclidean geometry has also served as benchmarks in machine learning and NLP for multimodal understanding of geometric problems and diagrams~\citep{lu2021inter,chen2021geoqa,gao2023g,peng2023geodrl,kazemi2023geomverse,liang2023unimath}. These methods produce informal solutions given informal problem statements, whereas we formalize both problems and solutions. UniGeo~\citep{chen2022unigeo} is a dataset consisting of 9,543 informal theorems and proofs. We formalize 125 random examples from UniGeo and include them in {\dataname}, which shows the generality of our formalization beyond Euclid's \emph{Elements}. 

\section{Formalizing Euclidean Geometry in Lean}

To construct the {\dataname } benchmark, we implement a variant of system E~\citep{avigad2009formal} in Lean and use it to formalize Euclidean geometry problems. {\dataname } has two key features: First, it supports implicit diagrammatic reasoning, which is critical to faithfully formalizing geometric proofs. Second, it can check the logical equivalence between theorems, which is essential for automatically evaluating the autoformalized theorems.

\subsection{The Lean Proof Assistant}
Proof assistants, a.k.a. interactive theorem provers, are languages and IDEs for humans to write formal proofs that can be checked by computers~\citep{leroy2016compcert,hales2017formal}. Widely used proof assistants include Coq, Isabelle, and Lean. They are similar from a machine learning perspective. Fig.~\ref{fig:overall} (\emph{Middle}) shows examples of theorems and proofs in Lean. Each theorem has a name and a theorem statement. A proof consists of multiple steps, and each step is called a ``tactic''. For example, ``\texttt{use c}'' and ``\texttt{euclid\_apply circle\_from\_points b a as ACE}'' are tactics.
 
Given a theorem and its proof, Lean can check the proof and output an error message if it is incorrect. Lean can also check if the theorem is well-formed. The percentage of well-formed theorems 
autoformalized by the model is widely used as a proxy metric in prior works~\citep{jiang2023multilingual,azerbayev2023proofnet}. However, well-formedness does not guarantee correctness, since a predicted theorem statement can be well-formed but have a different meaning from the ground truth.

\subsection{The Formal System E}
E is a formal system for modeling theorems and proofs in Euclidean geometry inspired by Euclid's \emph{Elements}. It defines basic geometric objects (e.g., points and lines) and relations between them (e.g., a point being on a line). It also prescribes how to formally state and prove theorems. E is a conceptual framework that can be implemented within any proof assistant. We explain our Lean implementation using \texttt{proposition\_1} in Fig.~\ref{fig:overall} as a running example. For a complete specification of E, please refer to Appendix~\ref{appendix:e} and E's paper~\citep{avigad2009formal}.

\smallsec{Geometric Objects, Relations, and Theorems}
E has six types of objects: points, lines, segments, circles, angles, and triangles. They can form various relations, e.g., ``\texttt{onLine a AB}'' means the point \texttt{a} is on the line \texttt{AB}. Some relations are defined as syntactic sugar of more basic relations, e.g., ``\texttt{distinctPointsOnLine a b AB}'' is the conjunction of ``\texttt{onLine a AB}'', ``\texttt{onLine b AB}'', and ``\texttt{a $\neq$ b}''. A segment \texttt{(a--b)} is specified by its two endpoints \texttt{a} and \texttt{b}, and it has a length \texttt{|(a--b)|} $\in \mathbb{R}$. Using these concepts, it is straightforward to understand the theorem statement of \texttt{proposition\_1}: 
\begin{lstlisting}[language=lean,numbers=none]
theorem proposition_1 : (*@$\forall$@*) (a b : Point) (AB : Line), distinctPointsOnLine a b AB (*@$\rightarrow$@*) (*@$\exists$@*) c : Point, |(c(*@--@*)a)| = |(a(*@--@*)b)| (*@$\land$@*) |(c(*@--@*)b)| = |(a(*@--@*)b)|
\end{lstlisting}
Like above, most theorems in Euclidean geometry take the form of \emph{given existing objects satisfying certain properties, constructing new objects satisfying certain properties}. Below is the general form of theorems in E:
\begin{multline*}
    \forall~(\overrightarrow{P} : Point)~(\overrightarrow{L} : Line)~(\overrightarrow{C} : Circle),~\bigwedge_{i=1}^{m} a_i \rightarrow\\
    \exists~(\overrightarrow{P'} : Point)~(\overrightarrow{L'} : Line)~(\overrightarrow{C'} : Circle),~\bigwedge_{j=1}^{n} a'_j
\end{multline*}
That is, given points $\overrightarrow{P}$, lines $\overrightarrow{L}$, and circles $\overrightarrow{C}$ satisfying $\bigwedge_{i=1}^{m} a_i$, one can construct points $\overrightarrow{P'}$, lines $\overrightarrow{L'}$, and circles $\overrightarrow{C'}$ satisfying $\bigwedge_{j=1}^{n} a'_j$. Each clause $a_i$ is called a ``precondition''. Logically, it is a disjunction of literals with variables from $\overrightarrow{P}$, $\overrightarrow{L}$, and $\overrightarrow{C}$. Each $a'_j$ is called a ``postcondition'', which is a disjunction of literals with variables from $\overrightarrow{P}$, $\overrightarrow{L}$, $\overrightarrow{C}$, $\overrightarrow{P'}$, $\overrightarrow{L'}$, and $\overrightarrow{C'}$. Variable sets can be empty. For example, in \texttt{proposition\_1}, we have $\overrightarrow{C} = \overrightarrow{L'} = \overrightarrow{C'} = \varnothing$.

\smallsec{Axioms}
Axioms take the same form as theorems, but they are built into the system and do not need to be proved. Appendix~\ref{appendix:e} contains the complete list of axioms in our instantiation of the formal system E. They fall into two categories: construction rules and non-construction rules.\footnote{``Non-construction rules'' include diagrammatic, metric, and transfer rules in E's original terminology~\citep{avigad2009formal}.} Construction rules construct new objects, i.e., at least one of $\overrightarrow{P'}$, $\overrightarrow{L'}$, and $\overrightarrow{C'}$ must be non-empty. For example, \texttt{circle\_from\_points} constructs a circle \texttt{$\alpha$} given a point \texttt{a} as its center and another point \texttt{b} on it:
\begin{lstlisting}[language=lean,numbers=none]
axiom circle_from_points : (*@$\forall$@*) (a b : Point), a (*@$\neq$@*) b (*@$\rightarrow$@*) (*@$\exists$@*) (*@$\alpha$@*) : Circle, (isCenter a (*@$\alpha$@*)) (*@$\land$@*) (onCircle b (*@$\alpha$@*))
\end{lstlisting}
Given two intersecting circles $\alpha$ and $\beta$, \texttt{intersection\_circles} constructs their intersection:
\begin{lstlisting}[language=lean,numbers=none]
axiom intersection_circles : (*@$\forall$@*) ((*@$\alpha$@*) (*@$\beta$@*) : Circle), intersectsCircle (*@$\alpha$@*) (*@$\beta$@*) (*@$\rightarrow$@*) (*@$\exists$@*) c : Point, (onCircle c (*@$\alpha$@*)) (*@$\land$@*) (onCircle c (*@$\beta$@*))
\end{lstlisting}
Non-construction rules deduce properties of existing objects without creating new objects. For examples:
\begin{lstlisting}[language=lean,numbers=none]
axiom center_inside_circle : (*@$\forall$@*) (a : Point) ((*@$\alpha$@*) : Circle), isCenter a (*@$\alpha$@*) (*@$\rightarrow$@*) insideCircle a (*@$\alpha$@*)

axiom intersection_circle_circle_2: (*@$\forall$@*) (a b : Point) ((*@$\alpha$@*) (*@$\beta$@*) : Circle), (onCircle a (*@$\alpha$@*)) (*@$\land$@*) (insideCircle b (*@$\alpha$@*)) (*@$\land$@*) (insideCircle a (*@$\beta$@*)) (*@$\land$@*) (onCircle b (*@$\beta$@*)) (*@$\rightarrow$@*) intersectsCircle (*@$\alpha$@*) (*@$\beta$@*)
\end{lstlisting}
The distinction between construction and non-construction rules is important for automated reasoning in Sec.~\ref{subsec:smt}.

\begin{figure}[ht]
  \centering
  \includegraphics[width=1.0\linewidth]{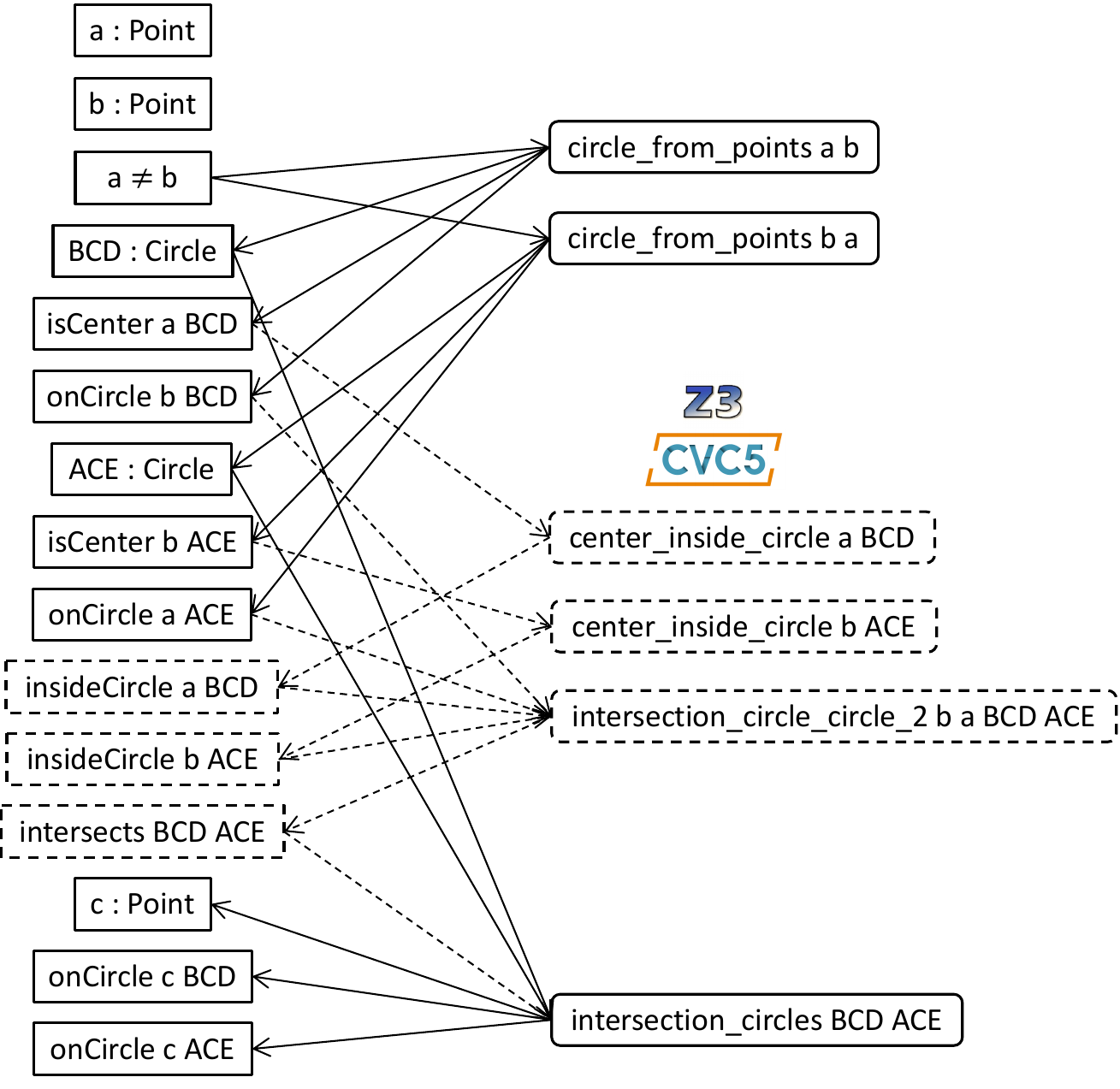}
  \caption{Three steps in the proof in Fig.~\ref{fig:overall}. \textbf{Left}: Geometric objects and facts. \textbf{Right}: Rules applied to construct new objects and deduce new facts. Each rule has inbound arrows from its preconditions and outbound arrows to its postconditions. \textbf{Dashed lines}: When applying a rule with missing preconditions, we try to fill the gap using a symbolic reasoning engine based on SMT solvers. These implicit steps performed by the symbolic engine can only use non-construction rules, whereas explicit steps performed by humans (or machine learning models) can use any rules.}
  \label{fig:proof}
\end{figure}

\smallsec{Proofs}
We implement the formal system E in Lean. Therefore, a proof can be made of arbitrary tactics as long as the entire proof can be checked by Lean. Nevertheless, we provide several customized tactics that can cover most steps used by Euclid (complete list in Appendix~\ref{appendix:e}). The most important tactic we provide is \texttt{euclid\_apply}, which applies a rule (either an axiom or an existing theorem) in the forward direction. Taking \texttt{intersection\_circles} as an example, when using this rule, one should explicitly specify two circles, say, \texttt{BCD} and \texttt{ACE}:
\begin{lstlisting}[language=lean,numbers=none]
intersection_circles BCD ACE : intersectsCircle BCD ACE (*@$\rightarrow$@*) (*@$\exists$@*) c : Point, (onCircle c BCD) (*@$\land$@*) (onCircle c ACE)
\end{lstlisting}
``\texttt{euclid\_apply intersection\_circles BCD ACE as c}'' instructs Lean to check if the precondition ``\texttt{intersectsCircle BCD ACE}'' is a known fact. If so, it introduces a new point \texttt{c} and two new facts, ``\texttt{onCircle c BCD}'' and ``\texttt{onCircle c ACE}''. If not, it triggers diagrammatic reasoning, trying to prove ``\texttt{intersectsCircle BCD ACE}'' using a symbolic automated reasoning engine (Sec.~\ref{subsec:smt}). If successful, it proceeds as normal. Otherwise, the tactic fails. Fig.~\ref{fig:proof} shows three proof steps using \texttt{euclid\_apply} with different rules. The last step triggers diagrammatic reasoning.

\subsection{Symbolic Reasoning Engine Based on SMT}
\label{subsec:smt}
We introduce a symbolic reasoning engine for deducing ``obvious'' facts from existing geometric objects and relations. It has two important uses in {\dataname }: (1) diagrammatic reasoning, as shown in the previous example, and (2) checking the equivalence between theorems.

In Euclidean geometry, a natural way to measure obviousness is whether construction rules are used. Applying construction rules (e.g., drawing auxiliary lines) requires creativity, which is non-obvious and hard to automate. In contrast, non-construction rules are mechanic and more amenable to automated reasoning. Intuitively, they do not introduce new objects and will not explode the search space. Therefore, we design the symbolic engine to be an automated theorem prover equipped with non-construction rules in E.

Instead of generic first-order provers~\citep{schulz2013system}, our symbolic engine builds on top of SMT solvers. They have efficient decision procedures for handling real numbers, which is important for reasoning about geometric quantities such as lengths, angles, and areas. Given a target conclusion (e.g., ``\texttt{intersectsCircle BCD ACE}'' in Fig.~\ref{fig:proof}), our symbolic engine proves it by refutation: It generates a list of SMT formulas, including all non-construction rules in E, all existing objects/relations, and the \emph{negation} of the conclusion. Then, it runs off-the-shelf SMT solvers to check if the list of formulas is satisfiable. The target conclusion is successfully proved if the solver returns UNSAT (not satisfiable).

\subsection{\dataname: Overview and Dataset Statistics}

In constructing {\dataname }, we manually formalized 173 theorems/proofs in Lean. 48 examples are from Euclid's \emph{Elements} (Book I),\footnote{We use an open-source version of Euclid's \emph{Elements} at \url{https://github.com/rfitzp/Elements}.} and 125 are adapted from UniGeo~\citep{chen2022unigeo}. Our formal proofs of \emph{Elements} vindicate Euclid's use of diagrams, as they follow Euclid's texts closely and leave diagrammatic reasoning implicit. We identified several small gaps in \emph{Elements}. Euclid occasionally used a stronger version of a theorem he had proved. Furthermore, when there were multiple cases, Euclid would discuss only one and omit others. Most omissions are acceptable in modern mathematics, e.g., when multiple cases are symmetric. The only omission we find questionable is in the proof of Proposition~24, where Euclid discussed only a simple case but omitted a more challenging case (details in Appendix~\ref{appendix:prop_24}). 

UniGeo contains problems in five categories: triangle, congruent, similarity, quadrilateral, and parallel. {\dataname} includes 25 random problems from each category. Problems from UniGeo are generally easier than those from \emph{Elements}. Each problem comes with a diagram and a textual question. Unlike \emph{Elements}, UniGeo's text does not include complete information about the problem, so we manually add missing diagrammatic details to the text.

\section{Autoformalizing Euclidean Geometry with Large Language Models}
Next, we describe our approach to autoformalizing (and evaluating) theorems and proofs targeting {\dataname}. 

\subsection{Autoformalizing and Evaluating Theorems}
Autoformalizing a theorem statement requires taking natural language input and generating the corresponding formal theorem in Lean. We use LLMs' in-context learning capability for autoformalization. Our prompt template contains an overview of {\dataname} syntax, some examples of well-formed formulas, and various guidelines for the task. For few-shot learning, we also include $k$ examples of informal-formal pairs as in-context demonstrations. The complete prompt template is in Appendix~\ref{appendix:prompts}.

When using our symbolic engine for checking the equivalence between theorems, we call it {\toolName} (\textbf{E}uclidean \textbf{E}quivalence \textbf{E}ngine). We also provide a Python wrapper for {\toolName}, supporting various configuration options and easy integration with different LLMs. We provide a high-level overview of {\toolName}'s two primary evaluation procedures: logical equivalence and approximate equivalence. 

\smallsec{Logical Equivalence}
The logical equivalence procedure takes a predicted formula $T_{pred}$ and ground truth formula $T_{gt}$ and attempts to prove $T_{pred} \iff T_{gt}$ using SMT solvers. {\toolName} performs a separate satisfiability check for each direction, each of which can succeed or fail.\footnote{{\toolName} uses not only non-construction rules but also a handful of construction rules (details in Appendix~\ref{appendix:e}).} Of course, $T_{pred}$ may actually be unsatisfiable (meaning it contains a contradiction), in which case we will always be able to prove $T_{pred} \implies T_{gt}$. As such, {\toolName} also performs a separate check to see whether a contradiction can be found in $T_{pred}$.

\smallsec{Approximate Equivalence} If logical equivalence cannot be proven, we may still want to gauge whether $T_{pred}$  is a semantically ``close'' formalization of $T_{gt}$. For instance, $T_{pred}$ may differ from $T_{gt}$ by only a single extra precondition or postcondition. To this end, {\toolName} also supports \emph{approximate equivalence checking}, which is a low-level semantic comparison between the two formulas. 
Specifically, {\toolName} will attempt to separately prove each of the preconditions and the postconditions of each formula in a clause-by-clause fashion.
However, this requires choosing a unification of the bound variables of the two formulas. Since trying each possible unification is infeasible, we instead choose the best $n$ unifications using a string similarity heuristic. For each candidate unification, we check how many pre/postconditions can be proved from one proposition to the other. \footnote{{\toolName} currently only performs approximate equivalence checking when the predicted formula has the same number of bound variables as the ground truth formula.}

\subsection{Autoformalizing Proofs}
To autoformalize proofs, we provide the LLM with a natural language description of the proposition, its natural language proof, and a formalization of the proposition in {\dataname}. Its task is to produce a formal proof consisting of a sequence of tactics. We provide a prompt template similar to the one used to autoformalize theorem statements. In addition to a brief overview of system E, we describe the custom {\dataname} tactics (e.g., ${\tt euclid\_apply}$), and provide a list of axioms and theorems the model can use in its proof. Beyond the core axioms of E, the list of helper theorems available to the model depends on the proposition being proven. For example, if we are formalizing the proof of Proposition 10 in \emph{Elements}, then the set of available helper theorems will include all versions of Propositions 1--9. Once the autoformalized proof is generated, we check its correctness using Lean and attempt to fill in diagrammatic reasoning gaps using the symbolic reasoning engine. The complete prompt template is given in Appendix~\ref{appendix:prompts}.

\section{Experiments}

\subsection{Autoformalizing Theorem Statements}

\smallsec{Experimental Setup}
We evaluated the efficacy of LLMs on autoformalizing theorem statements using GPT-4 and GPT-4V. The input modality for GPT-4 was restricted to textual questions, while GPT-4V extended this by including diagrams as well. To implement few-shot learning, we randomly selected five propositions from Euclid's \emph{Elements} and five problems from each category in the UniGeo dataset, serving as in-context learning examples. We then use {\toolName} to automatically evaluate the results of each round. To see how well {\toolName} correlates with human evaluation, we manually evaluate a sample of formalized theorems from \emph{Elements} to identify any false negatives/positives.

\begin{table}[ht]
    \centering
    \resizebox{\linewidth}{!}{
    \begin{tabular}{c|ccc|ccc}\toprule
     &   \multicolumn{3}{c|}{GPT-4} & \multicolumn{3}{c}{GPT-4V}\\
       Dataset & 0-shot &  1-shot & 5-shot & 0-shot & 1-shot & 5-shot \\
        \midrule
        \emph{Elements} & 2.3\% & 4.7\% & 16.3\% & 2.3\% & 4.7\% & 20.9\% \\
        UniGeo  & 3.0\% & 9.0\% & 20.0\% & 5.0\% & 10.0\% & 21.0\% \\
        \midrule
        Overall & 2.8\% & 7.7\% & 18.9\% & 4.2\% & 8.4\% & 21.0\% \\
        \bottomrule
    \end{tabular}
    }
    \caption{Percentage of proved semantic equivalences from autoformalized theorem statements from \emph{Elements} and UniGeo. Experiments were conducted in January 2024 using \texttt{gpt-4-1106-preview} and \texttt{gpt-4-1106-vision-preview}.}
    \label{tab:propFormalization}
\end{table}

\smallsec{Results}
Table~\ref{tab:propFormalization} shows the accuracies of autoformalizing theorem statements, where correctness is measured by using {\toolName} to check logical equivalence. Results on the \emph{Elements} and UniGeo parts of LeanEuclid are also shown separately. Across both parts, few-shot learning with 5 shots produces correct formalizations at a rate of 21\%. We also see a small improvement when visual inputs are included. We emphasize that Table~\ref{tab:propFormalization} was generated automatically by {\toolName}. While some autoformalized theorem statements are syntactic matches of their ground truth and, therefore, easily identified as correct, others are not. Manual evaluation of these instances is expensive and tedious, whereas {\toolName} is able to identify these cases easily.

For cases where equivalence fails, {\toolName} reports whether either of $T_{gt} \implies T_{pred}$ or $T_{pred} \implies T_{gt}$ can be proved, allowing us to partition the failed cases into different classes. For instance, two rounds of autoformalization with GPT-4 (1-shot) yielded 67 rejected predictions, for 31 of these we could prove only $T_{gt} \implies T_{pred}$, for 15 we could prove only $T_{pred} \implies T_{gt}$, and for 21 we could prove neither.

\begin{figure}[ht]
    \centering
    \includegraphics[width=\columnwidth]{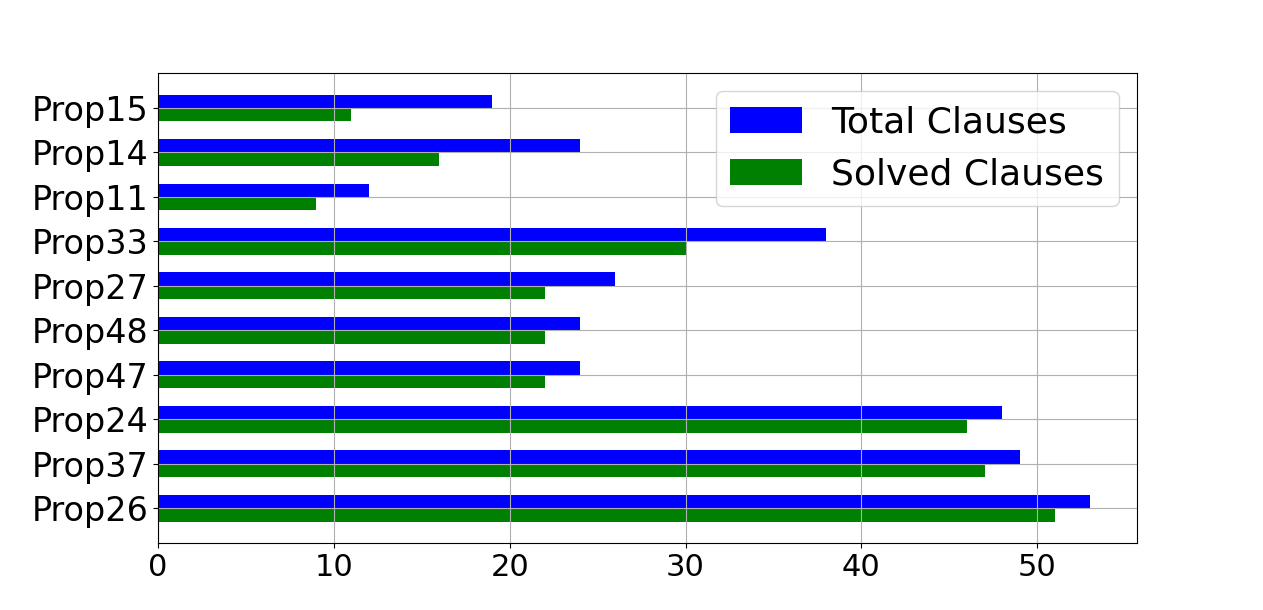}
    \caption{Approximate equivalence checking results for theorems from \emph{Elements} formalized by GPT-4 (5-shots).}
    \label{fig:approxDraft}
\end{figure}

Given that only a minority of predictions are provably correct, we can send the remainder to {\toolName}'s approximate equivalence checker to glean ``close'' formalizations. For brevity, we only showcase this evaluation on the results of GPT-4 (5-shot) on Euclid's \emph{Elements}. Of the 36 rejected predictions, 10 possess the correct quantity and type of bound variables, and so are amenable to approximate analysis. The results of this analysis are shown in Fig.~\ref{fig:approxDraft}. For each proposition, we show the total number of clauses (preconditions and postconditions of both formulas) and the number of clauses that could be proved. We can see that, in addition to the 7 propositions that were provably equivalent, this round produced 5 propositions for which more than 90\% of all proof obligations can be resolved. Some interesting examples produced during our experiments are in Appendix~\ref{appendix:props}.

\smallsec{Comparison with Manual Evaluation}
To compare the accuracy of {\toolName} to manual evaluation, we took a sample of 86 formalizations and investigated them for false negatives. The examples were taken from the results of GPT-4 and GPT-4V (both 5-shot).  Among the 86 autoformalized theorem statements, 16 were proved equivalent to the ground truth, and all 16 were judged to be correct by humans. The remaining 70 could not be proved equivalent, but human inspection revealed that 3 should be equivalent to the ground truth. This gives us an estimated false negative rate of 15.8\%. These instances are shown in Appendix~\ref{appendix:props}.

With respect to false positives,  we did not find indications of soundness bugs in {\toolName}. However, a false positive can occur when an autoformalized theorem \emph{happens} to be provably equivalent to the ground truth, but is unlikely to be identified by a human as a ``faithful'' formalization of the given proposition. We identified only one such case from our experiments, which is also shown in Appendix~\ref{appendix:props}.

\paragraph{Why is Formalizing Theorem Statements Hard?}
Compared to previous autoformalization results~\cite{wu2022autoformalization}, our experiments show that the models struggle to correctly formalize most of the theorems in our dataset. 
We suspect this is primarily a result of using the formal system E as a specification language. E is designed primarily as a proof system, and not as a specification language; as noted by Avigad, many basic relation constructs in E (e.g., ${\tt between}$, ${\tt sameSide}$) are almost never mentioned explicitly in Euclid's actual writing~\cite{avigad2009formal}. Furthermore, the language can only refer to composite structures (angles, triangles, etc.) in terms of their atomic components (points, lines, etc.). This makes theorem statements in E relatively verbose, and this verbosity introduces more room for the model to make small mistakes. 

In summary, while state-of-the-art models struggle to successfully autoformalize many of the theorems in our dataset, we see that {\toolName} can successfully identify and quantify the correctness of autoformalized theorem statements; in particular, despite being incomplete by design, {\toolName} only produces a small number of false negatives. We believe that {\toolName} can significantly facilitate the training and validation of autoformalization models targeting Euclidean geometry. Moreover, we believe that similar tools can be developed for other domains if an appropriate formal theory can be defined.

\subsection{Autoformalizing Proofs}
To check whether {\dataname} is a suitable target for autoformalizing proofs, we attempted to autoformalize proofs from \emph{Elements} and UniGeo using GPT-4 and GPT-4V.

\smallsec{Experimental Setup}
We tested each model against 43 proofs from \emph{Elements} and 100 proofs from UniGeo. To demonstrate concretely the capabilities and limitations of the model in writing formal {\dataname} proofs, we attempted to formalize entire proofs from single queries, rather than using an iterative or search-based autoformalization procedure. We evaluated the formalized UniGeo proofs based on whether it is verified by Lean as-is, and experimented with 0-shot, 1-shot, and 5-shot prompts. {The proofs from \emph{Elements} are more complex, so we did not anticipate many proofs to be completely correct. Instead, we measured how much effort is required to \emph{repair} the autoformalized proofs into proofs that are accepted by Lean. 

We manually repaired each incorrectly autoformalized proof from \emph{Elements}, attempting to make as few alterations as required. Using our ground truth proof as a reference point, we modified invalid tactics that could be repaired (e.g., by rearranging the order of its arguments), added missing tactics, and removed tactics that could not easily be repaired. Unnecessary but valid tactics were left unchanged.

\begin{table}[t]
    \centering
    \resizebox{0.8\linewidth}{!}{
    \begin{tabular}{c|c c| c c}\toprule
     &   \multicolumn{2}{c|}{GPT-4} & \multicolumn{2}{c}{GPT-4V}\\
       Category &  1-shot & 5-shot & 1-shot & 5-shot \\
        \midrule
        Triangle &  35\% & 45\%  & 45\% & 70\% \\
        Similarity  & 5\% & 15\% & 10\% & 15\% \\
        Congruent  & 5\% & 25\% & 15\% & 25\%  \\
        Quadrilateral  & 35\% & 25\% &  20\% & 30\%  \\
        Parallel & 5\% & 15\%  &  5\% & 15\%\\
        \midrule
        Overall &  17\% & 25\% & 19\% & 31\% \\
        \bottomrule
    \end{tabular}
    }
    \caption{Percentage of successfully autoformalized proofs from UniGeo. Experiments were conducted in January 2024 using \texttt{gpt-4-1106-preview} and \texttt{gpt-4-1106-vision-preview}.}
    \label{tab:unigeoProof}
\end{table}

\smallsec{Results}
Table~\ref{tab:unigeoProof} shows the results of autoformalizing proofs from UniGeo proofs. Models with 0-shot prompts are not included since they failed to autoformalize any proofs. In general, the performance of the models significantly depends on the type of geometry problems and the number of few-shot demonstrations. Compared to theorem statements, we see a more significant improvement in the success rate when visual inputs are provided for autoformalizing proofs.

When autoformalizing proofs from \emph{Elements} with 5-shot prompts, we found that GPT-4 and GPT-4V were both only able to completely formalize the same two proofs (Propositions 1 and Proposition 17). That is to say, when combined with the UniGeo results, GPT-4 formalized correct proofs at a rate of 18.8\% on {\dataname}, while GPT-4V achieved a rate of 23.1\%. The remaining 41 proofs from \emph{Elements} required some degree of repair to be accepted by Lean. 

While it is difficult to precisely measure the quality of imperfect proofs, we can gain a rough approximation by computing the Levenshtein ratio between the original and repaired proofs. Doing so reveals that, for GPT-4, the autoformalized proofs had a median similarity ratio of 61.7\% compared to their repaired versions, with proofs in the 75th percentile scoring at least 75.2\%. For GPT-4V, the median similarity ratio was 64.0\%, and the proofs in the 75h percentile proofs scored at least 72.9\%. Moreover, we found that many of the modifications required to repair the proofs are very simple, such as strengthening a theorem slightly or rearranging tactics arguments; in general, the models are good at choosing relevant theorems, even if they do not invoke them correctly. Additional data and examples are in Appendix~\ref{appendix:proofs}.

We believe that these results reflect well on {\dataname} as a target language for autoformalizing Euclidean proofs. In particular, our tactic language and proof automation allow the model to focus only on explicit reasoning steps in the input text. This means the resulting proofs are much shorter than they would be if all reasoning steps were made explicit (and, as a result, they are easier to repair). Given that these results were obtained from standalone queries, we expect that neural theorem provers or proof repair tools targeting {\dataname} could be used to successfully formalize a significant portion of the proofs in our dataset.

\section{Conclusion}
We have demonstrated Euclidean geometry as an attractive target for autoformalization. With our SMT-based symbolic engine, it is feasible to (1) automatically evaluate autoformalized theorems statements and (2) have the model autoformalize only explicit proof steps, leaving diagrammatic reasoning implicit. We have constructed the {\dataname } benchmark to facilitate future research on autoformalizing Euclidean geometry.

\section{Limitations and Open Problems}
{\dataname }'s size (173 examples) is reasonable as a benchmark for testing, e.g., considering HumanEval~\citep{chen2021evaluating}, the most widely used benchmark for code generation, which has only 164 examples. Nevertheless, more data examples are necessary for finetuning-based approaches. A promising way of collecting examples inexpensively is to automatically translate existing geometry datasets, e.g., the synthetic dataset in AlphaGeometry, into our formalism.

Even though we use GPT-4 in experiments, data contamination is unlikely since our test data, {\dataname }, was not publicly available during the experiments. GPT-4 may have seen other formalizations of \emph{Elements} on the Internet, but they are fundamentally different from {\dataname} (Sec.~\ref{sec:related}).

We focus on Euclidean geometry, but findings related to LeanEuclid may be helpful to autoformalization in general. First, LeanEuclid provides a benchmark for developing domain-agnostic ideas for autoformalization, e.g., visual modality, retrieval, and learning from feedback. Second, we choose Euclidean geometry because its domain knowledge enables automated reasoning that can (1) check the equivalence between theorems and (2) fill in implicit reasoning gaps. Other domains with similar characteristics may also benefit from our methodology for constructing LeanEuclid.

Our work focuses on constructing a benchmark for autoformalization rather than introducing a novel model. Nevertheless, the LeanEuclid benchmark paves the way for future modeling efforts, such as having LLMs incorporate feedback from Lean to improve autoformalization.

\section*{Acknowledgements}
This work is partially supported by the Natural Sciences and Engineering Research Council of Canada (NSERC) through the Discovery Grants (individual) program. Kaiyu Yang is partially supported by Caltech's Computing, Data, and Society Postdoctoral Fellowship. Xujie Si is partially supported by the Canada CIFAR AI Chairs program. We thank Jia Deng for his contributions to the conceptualization of this project when Kaiyu Yang was a Ph.D. student under his mentorship. We are also thankful to Jeremy Avigad for his invaluable guidance on the formal system E. Special thanks go to Marsha Chechik and Kuldeep Meel for their insightful feedback on the initial version of this paper.

\section*{Impact Statement}

This work aims to advance autoformalization by studying it in the controlled domain of Euclidean geometry. Autoformalization can be applied to formal mathematics, verification, and autonomous systems~\citep{chen2023nl2tl,cosler2023nl2spec,pan2023data}. There are many potential societal consequences of these downstream applications, none of which we feel must be specifically highlighted here.

\bibliography{references}
\bibliographystyle{icml2024}

\newpage
\appendix
\onecolumn

\setcounter{table}{0}
\renewcommand{\thetable}{\Alph{table}}
\setcounter{figure}{0}
\renewcommand{\thefigure}{\Alph{figure}}
\setcounter{section}{0}
\renewcommand{\thesection}{\Alph{section}}

\section{The Formal System E}
\label{appendix:e}

We describe our variant of the formal system E in Lean-like pseudocode, including the complete list of axioms. Our design mostly follows E's original specification~\citep{avigad2009formal} with a few minor deviations.

\subsection{Geometry Objects}

E has six types of geometric objects: points, lines, circles, segments, angles, and triangles.\footnote{Triangles are called ``areas'' in \citet{avigad2009formal}.} Points, lines, and circles are basic types, whereas segments, angles, and triangles are parameterized by points. For example, points \texttt{a}, \texttt{b}, \texttt{c} can form segment \texttt{(a--b)}, angle \texttt{$\angle$ a:b:c}, and triangle \texttt{$\bigtriangleup$ a:b:c}. 
\begin{lstlisting}[language=lean,numbers=none]
Point : Type
Line : Type
Circle : Type
Segment : Point (*@$\rightarrow$@*) Point (*@$\rightarrow$@*) Type
Angle : Point (*@$\rightarrow$@*) Point (*@$\rightarrow$@*) Point (*@$\rightarrow$@*) Type
Triangle : Point (*@$\rightarrow$@*) Point (*@$\rightarrow$@*) Point (*@$\rightarrow$@*) Type
\end{lstlisting}

\subsection{Functions, Constants, and Relations}

Segments, angles, and triangles are associated with metric properties such as length, degree, and size.
\begin{lstlisting}[language=lean,numbers=none]
length : Segment (*@$\rightarrow$@*) (*@$\mathbb{R}$@*)
degree : Angle (*@$\rightarrow$@*) (*@$\mathbb{R}$@*)
size : Triangle (*@$\rightarrow$@*) (*@$\mathbb{R}$@*)
\end{lstlisting}

Right angle ($\rightangle$) is a special constant, denoting 90 degrees. 
\begin{lstlisting}[language=lean,numbers=none]
(*@$\rightangle$@*) : (*@$\mathbb{R}$@*)
\end{lstlisting}

We have nine basic predicates denoting the geometric relations between points, lines, and circles. For example, ``\texttt{sameSide a b L}'' means points \texttt{a} and \texttt{b} are on the same side of the line \texttt{L}. Also, ``\texttt{between a b c}'' means points \texttt{a}, \texttt{b}, and \texttt{c} are collinear, and \texttt{b} lies between \texttt{a} and \texttt{c}.
\begin{lstlisting}[language=lean,numbers=none]
onLine : Point (*@$\rightarrow$@*) Line (*@$\rightarrow$@*) Prop
sameSide : Point (*@$\rightarrow$@*) Point (*@$\rightarrow$@*) Line (*@$\rightarrow$@*) Prop
between : Point (*@$\rightarrow$@*) Point (*@$\rightarrow$@*) Point (*@$\rightarrow$@*) Prop
onCircle : Point (*@$\rightarrow$@*) Circle (*@$\rightarrow$@*) Prop
insideCircle : Point (*@$\rightarrow$@*) Circle (*@$\rightarrow$@*) Prop
isCentre : Point (*@$\rightarrow$@*) Circle (*@$\rightarrow$@*) Prop
intersectsLine : Line (*@$\rightarrow$@*) Line (*@$\rightarrow$@*) Prop
Line.intersectsCircle : Line (*@$\rightarrow$@*) Circle (*@$\rightarrow$@*) Prop
Circle.intersectsCircle : Circle (*@$\rightarrow$@*) Circle (*@$\rightarrow$@*) Prop
\end{lstlisting}

In addition to the basic predicates, we have four predicates defined as syntactic sugars. For example, ``\texttt{opposingSides a b L}'' is the conjunction of ``\texttt{$\lnot$(onLine a L)}'', ``\texttt{$\lnot$(onLine b L)}'', and ``\texttt{$\lnot$(sameSide a b L)}''.
\begin{lstlisting}[language=lean,numbers=none]
opposingSides (a : Point) (b : Point) (l : Line) : Prop := (*@$\lnot$@*)(onLine a l) (*@$\land$@*) (*@$\lnot$@*)(onLine b l) (*@$\land$@*) (*@$\lnot$@*)(sameSide a b l)

outsideCircle (p : Point) (c : Circle) : Prop := (*@$\lnot$@*)(insideCircle p c) (*@$\land$@*) (*@$\lnot$@*)(onCircle p c)

formTriangle (a b c : Point) (AB BC CA : Line) : Prop := (distinctPointsOnLine a b AB) (*@$\land$@*) (onLine b BC) (*@$\land$@*) (onLine c BC) (*@$\land$@*) (onLine c CA) (*@$\land$@*) (onLine a CA) (*@$\land$@*) AB (*@$\neq$@*) BC (*@$\land$@*) BC (*@$\neq$@*) CA (*@$\land$@*) CA (*@$\neq$@*) AB

formParallelogram (a b c d : Point) (AB CD AC BD : Line) : Prop := (onLine a AB) (*@$\land$@*) (onLine b AB) (*@$\land$@*) (onLine c CD) (*@$\land$@*) (onLine d CD) (*@$\land$@*) (onLine a AC) (*@$\land$@*) (onLine c AC) (*@$\land$@*) (distinctPointsOnLine b d BD) (*@$\land$@*) (sameSide a c BD) (*@$\land$@*) (*@$\lnot$@*)(intersectsLine AB CD) (*@$\land$@*) (*@$\lnot$@*)(intersectsLine AC BD)
\end{lstlisting}

\subsection{Axioms}

Axioms in E fall into five categories: construction rules, diagrammatic rules, metric rules, transfer rules, and superposition rules (the last four categories are collectively referred to as ``non-construction rules'' in this paper). \citet{avigad2009formal} has proved E's axioms to be sound and complete through a bi-directional translation between E and Tarski's axioms~\citep{schwabhauser2013metamathematische}. E's axioms may have redundancy, as \citet{avigad2009formal} did not attempt to minimize the number of axioms.

\smallsec{Construction Rules}
We have 32 construction rules. Among them, two rules are not from E's paper but convenient for formalizing some proofs: \texttt{point\_between\_points\_shorter\_than}, \texttt{extend\_point\_longer}. One rule is stronger than its original version: \texttt{point\_on\_line\_same\_side}.
\begin{lstlisting}[language=lean,numbers=none]
arbitrary_point : (*@$\exists$@*) a : Point, true

distinct_points : (*@$\forall$@*) a : Point, (*@$\exists$@*) b : Point, a (*@$\neq$@*) b

line_nonempty : (*@$\forall$@*) l : Line, (*@$\exists$@*) p : Point, onLine p l

exists_distincts_points_on_line : (*@$\forall$@*) l : Line, (*@$\forall$@*) a : Point, (*@$\exists$@*) b : Point, a (*@$\neq$@*) b (*@$\land$@*) onLine b l

exists_point_between_points_on_line : (*@$\forall$@*) (L : Line) (b c : Point), distinctPointsOnLine b c L (*@$\rightarrow$@*) (*@$\exists$@*) a : Point, (onLine a L) (*@$\land$@*) (between b a c)

exists_point_between_points_not_on_line : (*@$\forall$@*) (L M : Line) (b c : Point), (distinctPointsOnLine b c L) (*@$\land$@*) L (*@$\neq$@*) M (*@$\rightarrow$@*) (*@$\exists$@*) a : Point, (onLine a L) (*@$\land$@*) (between b a c) (*@$\land$@*) (*@$\lnot$@*)(onLine a M)

point_between_points_shorter_than : (*@$\forall$@*) (L : Line) (b c : Point) (s : Segment), distinctPointsOnLine b c L (*@$\land$@*) (|s| > 0) (*@$\rightarrow$@*) (*@$\exists$@*) a : Point, (onLine a L) (*@$\land$@*) (between b a c) (*@$\land$@*) (|b(*@--@*)a| < |s|)

extend_point : (*@$\forall$@*) (L : Line) (b c : Point), distinctPointsOnLine b c L (*@$\rightarrow$@*) (*@$\exists$@*) a : Point, (onLine a L) (*@$\land$@*) (between b c a)

extend_point_not_on_line : (*@$\forall$@*) (L M : Line) (b c : Point), (distinctPointsOnLine b c L) (*@$\land$@*) L (*@$\neq$@*) M (*@$\rightarrow$@*) (*@$\exists$@*) a : Point, (onLine a L) (*@$\land$@*) (between b c a) (*@$\land$@*) (*@$\lnot$@*)(onLine a M)

extend_point_longer : (*@$\forall$@*) (L : Line) (b c : Point) (s : Segment), distinctPointsOnLine b c L (*@$\rightarrow$@*) (*@$\exists$@*) a : Point, (onLine a L) (*@$\land$@*) (between b c a) (*@$\land$@*) (|c(*@--@*)a| > |s|)

point_same_side : (*@$\forall$@*) (L : Line) (b : Point), (*@$\lnot$@*)(onLine b L) (*@$\rightarrow$@*) (*@$\exists$@*) a : Point, sameSide a b L

distinct_point_same_side: (*@$\forall$@*) (L : Line) (b c : Point), (*@$\lnot$@*)(onLine b L) (*@$\rightarrow$@*) (*@$\exists$@*) a : Point, a (*@$\neq$@*) c (*@$\land$@*) (sameSide a b L)

point_on_line_same_side : (*@$\forall$@*) (L M : Line) (b : Point), (*@$\lnot$@*)(onLine b L) (*@$\land$@*) (intersectsLine L M) (*@$\rightarrow$@*) (*@$\exists$@*) a : Point, (onLine a M) (*@$\land$@*) (sameSide a b L)

exists_point_opposite : (*@$\forall$@*) (L : Line) (b : Point), (*@$\lnot$@*)(onLine b L) (*@$\rightarrow$@*) (*@$\exists$@*) a : Point, opposingSides a b L

exists_distinct_point_opposite_side : (*@$\forall$@*) (L : Line) (b c : Point), (*@$\lnot$@*)(onLine b L) (*@$\rightarrow$@*) (*@$\exists$@*) a : Point, a (*@$\neq$@*) c (*@$\land$@*) (opposingSides a b L)

exists_point_on_circle : (*@$\forall$@*) ((*@$\alpha$@*) : Circle), (*@$\exists$@*) a : Point, onCircle a (*@$\alpha$@*)

exists_distinct_point_on_circle : (*@$\forall$@*) ((*@$\alpha$@*) : Circle) (b : Point), (*@$\exists$@*) a : Point, a (*@$\neq$@*) b (*@$\land$@*) (onCircle a (*@$\alpha$@*))

exists_point_inside_circle : (*@$\forall$@*) ((*@$\alpha$@*) : Circle), (*@$\exists$@*) a : Point, insideCircle a (*@$\alpha$@*)

exists_distinct_point_inside_circle : (*@$\forall$@*) ((*@$\alpha$@*) : Circle) (b : Point), (*@$\exists$@*) a : Point, a (*@$\neq$@*) b (*@$\land$@*) (insideCircle a (*@$\alpha$@*))

exists_point_outside_circle : (*@$\forall$@*) ((*@$\alpha$@*) : Circle), (*@$\exists$@*) a : Point, outsideCircle a (*@$\alpha$@*)

exists_distinct_point_outside_circle : (*@$\forall$@*) ((*@$\alpha$@*) : Circle) (b : Point),  (*@$\exists$@*) a : Point, a (*@$\neq$@*) b (*@$\land$@*) (outsideCircle a (*@$\alpha$@*))

line_from_points : (*@$\forall$@*) (a b : Point), a (*@$\neq$@*) b (*@$\rightarrow$@*) (*@$\exists$@*) L : Line, (onLine a L) (*@$\land$@*) (onLine b L)

circle_from_points : (*@$\forall$@*) (a b : Point), a (*@$\neq$@*) b (*@$\rightarrow$@*) (*@$\exists$@*) (*@$\alpha$@*) : Circle, (isCentre c (*@$\alpha$@*)) (*@$\land$@*) (onCircle b (*@$\alpha$@*))

intersection_lines : (*@$\forall$@*) (L M : Line), intersectsLine L M (*@$\rightarrow$@*) (*@$\exists$@*) a : Point, (onLine a L) (*@$\land$@*) (onLine a M)

intersection_circle_line : (*@$\forall$@*) ((*@$\alpha$@*) : Circle) (L : Line), intersectsCircle L (*@$\alpha$@*) (*@$\rightarrow$@*) (*@$\exists$@*) a : Point, (onCircle a (*@$\alpha$@*)) (*@$\land$@*) (onLine a L)

intersections_circle_line : (*@$\forall$@*) ((*@$\alpha$@*) : Circle) (L : Line), intersectsCircle L (*@$\alpha$@*) (*@$\rightarrow$@*) (*@$\exists$@*) (a b : Point), (onCircle a (*@$\alpha$@*)) (*@$\land$@*) (onLine a L) (*@$\land$@*) (onCircle b (*@$\alpha$@*)) (*@$\land$@*) (onLine b L) (*@$\land$@*) a (*@$\neq$@*) b

intersection_circle_line_between_points : (*@$\forall$@*) ((*@$\alpha$@*) : Circle) (L : Line) (b c :Point), (insideCircle b (*@$\alpha$@*)) (*@$\land$@*) (onLine b L) (*@$\land$@*) (outsideCircle c (*@$\alpha$@*)) (*@$\land$@*) (onLine c L) (*@$\rightarrow$@*) (*@$\exists$@*) a : Point, (onCircle a (*@$\alpha$@*)) (*@$\land$@*) (onLine a L) (*@$\land$@*) (between b a c)

intersection_circle_line_extending_points : (*@$\forall$@*) ((*@$\alpha$@*) : Circle) (L : Line) (b c :Point), (insideCircle b (*@$\alpha$@*)) (*@$\land$@*) distinctPointsOnLine b c L (*@$\rightarrow$@*) (*@$\exists$@*) a : Point, (onCircle a (*@$\alpha$@*)) (*@$\land$@*) (onLine a L) (*@$\land$@*) (between a b c)

intersection_circles : (*@$\forall$@*) ((*@$\alpha$@*) (*@$\beta$@*) : Circle), intersectsCircle (*@$\alpha$@*) (*@$\beta$@*) (*@$\rightarrow$@*) (*@$\exists$@*) a : Point, (onCircle a (*@$\alpha$@*)) (*@$\land$@*) (onCircle a (*@$\beta$@*))

intersections_circles : (*@$\forall$@*) ((*@$\alpha$@*) (*@$\beta$@*) : Circle), intersectsCircle (*@$\alpha$@*) (*@$\beta$@*) (*@$\rightarrow$@*) (*@$\exists$@*) (a b : Point), (onCircle a (*@$\alpha$@*)) (*@$\land$@*) (onCircle a (*@$\beta$@*)) (*@$\land$@*) (onCircle b (*@$\alpha$@*)) (*@$\land$@*) (onCircle b (*@$\beta$@*)) (*@$\land$@*) a (*@$\neq$@*) b

intersection_same_side : (*@$\forall$@*) ((*@$\alpha$@*) (*@$\beta$@*) : Circle) (b c d : Point) (L : Line), (intersectsCircle (*@$\alpha$@*) (*@$\beta$@*)) (*@$\land$@*) (isCentre c (*@$\alpha$@*)) (*@$\land$@*) (isCentre d (*@$\beta$@*)) (*@$\land$@*) (onLine c L) (*@$\land$@*) (onLine d L) (*@$\land$@*) (*@$\lnot$@*)(onLine b L) (*@$\rightarrow$@*) (*@$\exists$@*) a : Point, (onCircle a (*@$\alpha$@*)) (*@$\land$@*) (onCircle a (*@$\beta$@*)) (*@$\land$@*) (sameSide a b L)

intersection_opposite_side : (*@$\forall$@*) ((*@$\alpha$@*) (*@$\beta$@*) : Circle) (b c d : Point) (L : Line), (intersectsCircle (*@$\alpha$@*) (*@$\beta$@*)) (*@$\land$@*) (isCentre c (*@$\alpha$@*)) (*@$\land$@*) (isCentre d (*@$\beta$@*)) (*@$\land$@*) (onLine c L) (*@$\land$@*) (onLine d L) (*@$\land$@*) (*@$\lnot$@*)(onLine b L) (*@$\rightarrow$@*) (*@$\exists$@*) a : Point, (onCircle a (*@$\alpha$@*)) (*@$\land$@*) (onCircle a (*@$\beta$@*)) (*@$\land$@*) (opposingSides a b L)
\end{lstlisting}

\smallsec{Diagrammatic Rules}
We have 36 diagrammatic rules for deducing diagrammatic facts, i.e., relations among objects. They include four rules not in E's paper: \texttt{intersection\_lines\_common\_point}, \texttt{parallel\_line\_unique}, \texttt{intersection\_symm}, and \texttt{parallelogram\_same\_side}.
\begin{lstlisting}[language=lean,numbers=none]
two_points_determine_line : (*@$\forall$@*) (a b : Point) (L M : Line), distinctPointsOnLine a b L (*@$\land$@*) (onLine a M) (*@$\land$@*) (onLine b M) (*@$\rightarrow$@*) L = M

centre_unique : (*@$\forall$@*) (a b : Point) ((*@$\alpha$@*) : Circle), (isCentre c (*@$\alpha$@*)) (*@$\land$@*) (isCentre b (*@$\alpha$@*)) (*@$\rightarrow$@*) a = b

center_inside_circle : (*@$\forall$@*) (a : Point) ((*@$\alpha$@*) : Circle), isCentre c (*@$\alpha$@*) (*@$\rightarrow$@*) insideCircle a (*@$\alpha$@*)

inside_not_on_circle : (*@$\forall$@*) (a : Point) ((*@$\alpha$@*) : Circle), insideCircle a (*@$\alpha$@*) (*@$\rightarrow$@*) (*@$\lnot$@*)(onCircle a (*@$\alpha$@*))

between_symm : (*@$\forall$@*) (a b c : Point), between a b c (*@$\rightarrow$@*) (between c b a) (*@$\land$@*) (a (*@$\neq$@*) b) (*@$\land$@*) (a (*@$\neq$@*) c) (*@$\land$@*) (*@$\lnot$@*)(between b a c)

between_same_line_out : (*@$\forall$@*) (a b c : Point) (L : Line), (between a b c) (*@$\land$@*) (onLine a L) (*@$\land$@*) (onLine b L) (*@$\rightarrow$@*) onLine c L

between_same_line_in : (*@$\forall$@*) (a b c : Point) (L : Line), (between a b c) (*@$\land$@*) (onLine a L) (*@$\land$@*) (onLine c L) (*@$\rightarrow$@*) onLine b L

between_trans_in : (*@$\forall$@*) (a b c d : Point), (between a b c) (*@$\land$@*) (between a d b) (*@$\rightarrow$@*) between a d c

between_trans_out : (*@$\forall$@*) (a b c d : Point), (between a b c) (*@$\land$@*) (between b c d) (*@$\rightarrow$@*) between a b d

between_points : (*@$\forall$@*) (a b c : Point) (L : Line), (a (*@$\neq$@*) b) (*@$\land$@*) (b (*@$\neq$@*) c) (*@$\land$@*) (c (*@$\neq$@*) a) (*@$\land$@*) (onLine a L) (*@$\land$@*) (onLine b L) (*@$\land$@*) (onLine c L) (*@$\rightarrow$@*) (between a b c) (*@$\lor$@*) (between b a c) (*@$\lor$@*) (between a c b)

between_not_trans : (*@$\forall$@*) (a b c d : Point), (between a b c) (*@$\land$@*) (between a b d) (*@$\rightarrow$@*) (*@$\lnot$@*)(between c b d)

same_side_rfl : (*@$\forall$@*) (a : Point) (L : Line), (*@$\lnot$@*)(onLine a L) (*@$\rightarrow$@*) sameSide a a L

same_side_symm : (*@$\forall$@*) (a b : Point) (L : Line), sameSide a b L (*@$\rightarrow$@*) sameSide b a L

same_side_not_on_line : (*@$\forall$@*) (a b : Point) (L : Line), sameSide a b L (*@$\rightarrow$@*) (*@$\lnot$@*)(onLine a L)

same_side_trans : (*@$\forall$@*) (a b c : Point) (L : Line), (sameSide a b L) (*@$\land$@*) (sameSide a c L) (*@$\rightarrow$@*) sameSide b c L

same_side_pigeon_hole : (*@$\forall$@*) (a b c : Point) (L : Line), (*@$\lnot$@*)(onLine a L) (*@$\land$@*) (*@$\lnot$@*)(onLine b L) (*@$\land$@*) (*@$\lnot$@*)(onLine c L) (*@$\rightarrow$@*) (sameSide a b L) (*@$\lor$@*) (sameSide a c L) (*@$\lor$@*) (sameSide b c L)

pasch_1: (*@$\forall$@*) (a b c : Point) (L : Line), (between a b c) (*@$\land$@*) (sameSide a c L) (*@$\rightarrow$@*) sameSide a b L

pasch_2: (*@$\forall$@*) (a b c : Point) (L : Line), (between a b c) (*@$\land$@*) (onLine a L) (*@$\land$@*) (*@$\lnot$@*)(onLine b L) (*@$\rightarrow$@*) sameSide b c L

pasch_3: (*@$\forall$@*) (a b c : Point) (L : Line), (between a b c) (*@$\land$@*) (onLine b L) (*@$\rightarrow$@*) (*@$\lnot$@*)(sameSide a c L)

pasch_4: (*@$\forall$@*) (a b c : Point) (L M : Line), (L (*@$\neq$@*) M) (*@$\land$@*) (onLine b L) (*@$\land$@*) (onLine b M) (*@$\land$@*) distinctPointsOnLine a c M (*@$\land$@*) (a (*@$\neq$@*) b) (*@$\land$@*) (c (*@$\neq$@*) b) (*@$\land$@*) (*@$\lnot$@*)(sameSide a c L) (*@$\rightarrow$@*) between a b c

triple_incidence_1 : (*@$\forall$@*) (L M N : Line) (a b c d : Point), (onLine a L) (*@$\land$@*) (onLine a M) (*@$\land$@*) (onLine a N) (*@$\land$@*) (onLine b L) (*@$\land$@*) (onLine c M) (*@$\land$@*) (onLine d N) (*@$\land$@*) (sameSide c d L) (*@$\land$@*) (sameSide b c N) (*@$\rightarrow$@*) (*@$\lnot$@*)(sameSide b d M)

triple_incidence_2 : (*@$\forall$@*) (L M N : Line) (a b c d : Point), (onLine a L) (*@$\land$@*) (onLine a M) (*@$\land$@*) (onLine a N) (*@$\land$@*) (onLine b L) (*@$\land$@*) (onLine c M) (*@$\land$@*) (onLine d N) (*@$\land$@*) (sameSide c d L) (*@$\land$@*) (*@$\lnot$@*)(sameSide b d M) (*@$\land$@*) (*@$\lnot$@*)(onLine d M) (*@$\land$@*) (b (*@$\neq$@*) a) (*@$\rightarrow$@*) sameSide b c N

triple_incidence_3 : (*@$\forall$@*) (L M N : Line) (a b c d e : Point), (onLine a L) (*@$\land$@*) (onLine a M) (*@$\land$@*) (onLine a N) (*@$\land$@*) (onLine b L) (*@$\land$@*) (onLine c M) (*@$\land$@*) (onLine d N) (*@$\land$@*) (sameSide c d L) (*@$\land$@*) (sameSide b c N) (*@$\land$@*) (sameSide d e M) (*@$\land$@*) (sameSide c e N) (*@$\rightarrow$@*) sameSide c e L

circle_line_intersections : (*@$\forall$@*) (a b c : Point) (L : Line) ((*@$\alpha$@*) : Circle), (onLine a L) (*@$\land$@*) (onLine b L) (*@$\land$@*) (onLine c L) (*@$\land$@*) (insideCircle a (*@$\alpha$@*)) (*@$\land$@*) (onCircle b (*@$\alpha$@*)) (*@$\land$@*) (onCircle c (*@$\alpha$@*)) (*@$\land$@*) (b (*@$\neq$@*) c) (*@$\rightarrow$@*) between b a c

circle_points_between : (*@$\forall$@*) (a b c : Point) ((*@$\alpha$@*) : Circle), (*@$\lnot$@*)(outsideCircle a (*@$\alpha$@*)) (*@$\land$@*) (*@$\lnot$@*)(outsideCircle b (*@$\alpha$@*)) (*@$\land$@*) (between a c b) (*@$\rightarrow$@*) insideCircle c (*@$\alpha$@*)

circle_points_extend : (*@$\forall$@*) (a b c : Point) ((*@$\alpha$@*) : Circle), (*@$\lnot$@*)(outsideCircle a (*@$\alpha$@*)) (*@$\land$@*) (*@$\lnot$@*)(insideCircle c (*@$\alpha$@*)) (*@$\land$@*) (between a c b) (*@$\rightarrow$@*) (outsideCircle b (*@$\alpha$@*))

circles_intersections_diff_side : (*@$\forall$@*) (a b c d : Point) ((*@$\alpha$@*) (*@$\beta$@*) : Circle) (L : Line), ((*@$\alpha$@*) (*@$\neq$@*) (*@$\beta$@*)) (*@$\land$@*) (onCircle c (*@$\alpha$@*)) (*@$\land$@*) (onCircle c (*@$\beta$@*)) (*@$\land$@*) (onCircle d (*@$\alpha$@*)) (*@$\land$@*) (onCircle d (*@$\beta$@*)) (*@$\land$@*) (c (*@$\neq$@*) d) (*@$\land$@*) (isCentre c (*@$\alpha$@*)) (*@$\land$@*) (isCentre b (*@$\beta$@*)) (*@$\land$@*) (onLine a L) (*@$\land$@*) (onLine b L) (*@$\rightarrow$@*) (*@$\lnot$@*)(sameSide c d L)

intersection_lines_opposing: (*@$\forall$@*) (a b : Point) (L M : Line), (opposingSides a b L) (*@$\land$@*) (onLine a M) (*@$\land$@*) (onLine b M) (*@$\rightarrow$@*) intersectsLine L M

intersection_lines_common_point : (*@$\forall$@*) (a : Point) (L M : Line), onLine a L (*@$\land$@*) (onLine a M) (*@$\land$@*) (L (*@$\neq$@*) M) (*@$\rightarrow$@*) intersectsLine L M

parallel_line_unique : (*@$\forall$@*) (a : Point) (L M N : Line), (*@$\lnot$@*)(onLine a L) (*@$\land$@*) (onLine a M) (*@$\land$@*) onLine a N (*@$\land$@*) (*@$\lnot$@*)(intersectsLine L M) (*@$\land$@*) (*@$\lnot$@*)(intersectsLine L N) (*@$\rightarrow$@*) M = N

intersection_symm : (*@$\forall$@*) (L M : Line), intersectsLine L M (*@$\rightarrow$@*) intersectsLine M L

intersection_circle_line_1: (*@$\forall$@*) (a b : Point) ((*@$\alpha$@*) : Circle) (L: Line), (*@$\lnot$@*)(outsideCircle a (*@$\alpha$@*)) (*@$\land$@*) (*@$\lnot$@*)(outsideCircle b (*@$\alpha$@*)) (*@$\land$@*) (opposingSides a b L) (*@$\rightarrow$@*) intersectsCircle L (*@$\alpha$@*)

intersection_circle_line_2: (*@$\forall$@*) (a : Point) ((*@$\alpha$@*) : Circle) (L: Line), (insideCircle a (*@$\alpha$@*)) (*@$\land$@*) (onLine a L) (*@$\rightarrow$@*) intersectsCircle L (*@$\alpha$@*)

intersection_circle_circle_1: (*@$\forall$@*) (a b : Point) ((*@$\alpha$@*) (*@$\beta$@*) : Circle), (*@$\lnot$@*)(outsideCircle a (*@$\alpha$@*)) (*@$\land$@*) (*@$\lnot$@*)(outsideCircle b (*@$\alpha$@*)) (*@$\land$@*) (insideCircle a (*@$\beta$@*)) (*@$\land$@*) (outsideCircle b (*@$\beta$@*)) (*@$\rightarrow$@*) intersectsCircle (*@$\alpha$@*) (*@$\beta$@*)

intersection_circle_circle_2: (*@$\forall$@*) (a b : Point) ((*@$\alpha$@*) (*@$\beta$@*) : Circle), (onCircle a (*@$\alpha$@*)) (*@$\rightarrow$@*) (insideCircle b (*@$\alpha$@*)) (*@$\rightarrow$@*) (insideCircle a (*@$\beta$@*)) (*@$\rightarrow$@*) (onCircle b (*@$\beta$@*)) (*@$\rightarrow$@*) intersectsCircle (*@$\alpha$@*) (*@$\beta$@*)

parallelogram_same_side : (*@$\forall$@*) (a b c d : Point) (AB CD AC BD : Line), formParallelogram a b c d AB CD AC BD (*@$\rightarrow$@*) sameSide b d AC (*@$\land$@*) sameSide c d AB (*@$\land$@*) sameSide a b CD
\end{lstlisting}

\smallsec{Metric Rules}
We have 11 metric rules. Unlike diagrammatic rules, metric rules deduce metric facts, e.g., about lengths, angles, and areas. In addition, since we use SMT solvers as the symbolic reasoning engine, their built-in theories on real numbers are also included implicitly as metric rules.
\begin{lstlisting}[language=lean,numbers=none]
zero_segment_if : (*@$\forall$@*) (a b : Point), |(a(*@--@*)b)| = 0 (*@$\rightarrow$@*) a = b

zero_segment_onlyif : (*@$\forall$@*) (a b : Point), a = b (*@$\rightarrow$@*) |(a(*@--@*)b)| = 0

segment_gte_zero : (*@$\forall$@*) (s : Segment), 0 (*@$\leq$@*) s.length

segment_symmetric : (*@$\forall$@*) (a b : Point), |(a(*@--@*)b)| = |(b(*@--@*)a)|

angle_symm : (*@$\forall$@*) (a b c : Point), (a (*@$\neq$@*) b) (*@$\land$@*) (b (*@$\neq$@*) c) (*@$\rightarrow$@*) ((*@$\angle$@*) a:b:c = (*@$\angle$@*) c:b:a)

angle_range : (*@$\forall$@*) (ang : Angle), (0 : (*@$\mathbb{R}$@*)) (*@$\leq$@*) ang (*@$\land$@*) ang (*@$\leq$@*) (*@$\rightangle$@*) + (*@$\rightangle$@*)

degenerated_area : (*@$\forall$@*) (a b : Point), Triangle.area (*@$\bigtriangleup$@*) a:a:b = 0

area_gte_zero : (*@$\forall$@*) (ar : Triangle), 0 (*@$\leq$@*) Triangle.area ar

area_symm_1 : (*@$\forall$@*) (a b c : Point), Triangle.area ((*@$\bigtriangleup$@*)a:b:c) = Triangle.area((*@$\bigtriangleup$@*)c:a:b)

area_symm_2 : (*@$\forall$@*) (a b c : Point), Triangle.area ((*@$\bigtriangleup$@*) a:b:c) = Triangle.area ((*@$\bigtriangleup$@*)a:c:b)

area_congruence : (*@$\forall$@*) (a b c a' b' c' : Point), (a(*@--@*)b) = (a'(*@--@*)b') (*@$\land$@*) (b(*@--@*)c) = (b'(*@--@*)c') (*@$\land$@*) (c(*@--@*)a) = (c'(*@--@*)a') (*@$\land$@*) ((*@$\angle$@*) a:b:c) = ((*@$\angle$@*) a':b':c') (*@$\land$@*) ((*@$\angle$@*) b:c:a) = ((*@$\angle$@*) b':c':a') (*@$\land$@*) ((*@$\angle$@*) c:a:b) = ((*@$\angle$@*) c':a':b') (*@$\rightarrow$@*) Triangle.area ((*@$\bigtriangleup$@*) a:b:c) = Triangle.area ((*@$\bigtriangleup$@*) a':b':c')
\end{lstlisting}

\smallsec{Transfer Rules}
We have 23 transfer rules acting as the bridge between diagrammatic facts and metric facts. They include 5 rules not in E's paper: \texttt{flat\_angle\_if}, \texttt{flat\_angle\_onlyif}, \texttt{parallelogram\_area}, \texttt{sum\_parallelograms\_area}, and \texttt{rectangle\_area}.
\begin{lstlisting}[language=lean,numbers=none]
between_if : (*@$\forall$@*) (a b c : Point), between a b c (*@$\rightarrow$@*) |(a(*@--@*)b)| + |(b(*@--@*)c)| = |(a(*@--@*)c)|

equal_circles : (*@$\forall$@*) (a b c : Point) ((*@$\alpha$@*) (*@$\beta$@*) : Circle), (isCentre c (*@$\alpha$@*)) (*@$\land$@*) (isCentre c (*@$\beta$@*)) (*@$\land$@*) (onCircle b (*@$\alpha$@*)) (*@$\land$@*) (onCircle c (*@$\beta$@*)) (*@$\land$@*) |(a(*@--@*)b)| = |(a(*@--@*)c)| (*@$\rightarrow$@*) (*@$\alpha$@*) = (*@$\beta$@*)

point_on_circle_if : (*@$\forall$@*) (a b c : Point) ((*@$\alpha$@*) : Circle), (isCentre c (*@$\alpha$@*)) (*@$\land$@*) (onCircle b (*@$\alpha$@*)) (*@$\land$@*) |(a(*@--@*)c)| = |(a(*@--@*)b)| (*@$\rightarrow$@*) onCircle c (*@$\alpha$@*)

point_on_circle_onlyif : (*@$\forall$@*) (a b c : Point) ((*@$\alpha$@*) : Circle), (isCentre c (*@$\alpha$@*)) (*@$\land$@*) (onCircle b (*@$\alpha$@*)) (*@$\land$@*) (onCircle c (*@$\alpha$@*)) (*@$\rightarrow$@*) |(a(*@--@*)c)| = |(a(*@--@*)b)|

point_in_circle_if : (*@$\forall$@*) (a b c : Point) ((*@$\alpha$@*) : Circle), (isCentre c (*@$\alpha$@*)) (*@$\land$@*) (onCircle b (*@$\alpha$@*)) (*@$\land$@*) (|(a(*@--@*)c)| < |(a(*@--@*)b)|) (*@$\rightarrow$@*) insideCircle c (*@$\alpha$@*)

point_in_circle_onlyif : (*@$\forall$@*) (a b c : Point) ((*@$\alpha$@*) : Circle), (isCentre c (*@$\alpha$@*)) (*@$\land$@*) (onCircle b (*@$\alpha$@*)) (*@$\land$@*) (insideCircle c (*@$\alpha$@*)) (*@$\rightarrow$@*) |(a(*@--@*)c)| < |(a(*@--@*)b)|

degenerated_angle_if : (*@$\forall$@*) (a b c : Point) (L : Line), (a (*@$\neq$@*) b) (*@$\land$@*) (a (*@$\neq$@*) c) (*@$\land$@*) (onLine a L) (*@$\land$@*) (onLine b L) (*@$\land$@*) (onLine c L) (*@$\land$@*) (*@$\lnot$@*)(between b a c) (*@$\rightarrow$@*) (*@$\angle$@*) b:a:c = 0

degenerated_angle_onlyif : (*@$\forall$@*) (a b c : Point) (L : Line), (a (*@$\neq$@*) b) (*@$\land$@*) (a (*@$\neq$@*) c) (*@$\land$@*) (onLine a L) (*@$\land$@*) (onLine b L) (*@$\land$@*) ((*@$\angle$@*) b:a:c = 0) (*@$\rightarrow$@*) (onLine c L) (*@$\land$@*) (*@$\lnot$@*)(between b a c)

sum_angles_if : (*@$\forall$@*) (a b c d : Point) (L M : Line), (onLine a L) (*@$\land$@*) (onLine a M) (*@$\land$@*) (onLine b L) (*@$\land$@*) (onLine c M) (*@$\land$@*) (a (*@$\neq$@*) b) (*@$\land$@*) (a (*@$\neq$@*) c) (*@$\land$@*) (*@$\lnot$@*)(onLine d L) (*@$\land$@*) (*@$\lnot$@*)(onLine d M) (*@$\land$@*) (L (*@$\neq$@*) M) (*@$\land$@*) ((*@$\angle$@*) b:a:c) = ((*@$\angle$@*) b:a:d) + ((*@$\angle$@*) d:a:c) (*@$\rightarrow$@*) (sameSide b d M) (*@$\land$@*) (sameSide c d L)

sum_angles_onlyif : (*@$\forall$@*) (a b c d : Point) (L M : Line), (onLine a L) (*@$\land$@*) (onLine a M) (*@$\land$@*) (onLine b L) (*@$\land$@*) (onLine c M) (*@$\land$@*) (a (*@$\neq$@*) b) (*@$\land$@*) (a (*@$\neq$@*) c) (*@$\land$@*) (*@$\lnot$@*)(onLine d L) (*@$\land$@*) (*@$\lnot$@*)(onLine d M) (*@$\land$@*) (L (*@$\neq$@*) M) (*@$\land$@*) (sameSide b d M) (*@$\land$@*) (sameSide c d L) (*@$\rightarrow$@*) ((*@$\angle$@*) b:a:c) = ((*@$\angle$@*) b:a:d) + ((*@$\angle$@*) d:a:c)

perpendicular_if : (*@$\forall$@*) (a b c d : Point) (L : Line), (onLine a L) (*@$\land$@*) (onLine b L) (*@$\land$@*) (between a c b) (*@$\land$@*) (*@$\lnot$@*)(onLine d L) (*@$\land$@*) ((*@$\angle$@*) a:c:d = (*@$\angle$@*) d:c:b) (*@$\rightarrow$@*) (*@$\angle$@*) a:c:d = (*@$\rightangle$@*)

perpendicular_onlyif : (*@$\forall$@*) (a b c d : Point) (L : Line), (onLine a L) (*@$\land$@*) (onLine b L) (*@$\land$@*) (between a c b) (*@$\land$@*) (*@$\lnot$@*)(onLine d L) (*@$\land$@*) ((*@$\angle$@*) a:c:d = (*@$\rightangle$@*)) (*@$\rightarrow$@*) (*@$\angle$@*) a:c:d = (*@$\angle$@*) d:c:b

flat_angle_if : (*@$\forall$@*) (a b c : Point), a (*@$\neq$@*) b (*@$\land$@*) b (*@$\neq$@*) c (*@$\land$@*) ((*@$\angle$@*) a:b:c = (*@$\rightangle$@*) + (*@$\rightangle$@*)) (*@$\rightarrow$@*) between a b c

flat_angle_onlyif : (*@$\forall$@*) (a b c : Point), between a b c (*@$\rightarrow$@*) (*@$\angle$@*) a:b:c = (*@$\rightangle$@*) + (*@$\rightangle$@*)

equal_angles : (*@$\forall$@*) (a b b' c c' : Point) (L M : Line), (onLine a L) (*@$\land$@*) (onLine b L) (*@$\land$@*) (onLine b' L) (*@$\land$@*) (onLine a M) (*@$\land$@*) (onLine c M) (*@$\land$@*) (onLine c' M) (*@$\land$@*) (b (*@$\neq$@*) a) (*@$\land$@*) (b' (*@$\neq$@*) a) (*@$\land$@*) (c (*@$\neq$@*) a) (*@$\land$@*) (c' (*@$\neq$@*) a) (*@$\land$@*) (*@$\lnot$@*)(between b a b') (*@$\land$@*) (*@$\lnot$@*)(between c a c') (*@$\rightarrow$@*) ((*@$\angle$@*) b:a:c = (*@$\angle$@*) b':a:c')

lines_intersect : (*@$\forall$@*) (a b c d : Point) (L M N : Line), (onLine a L) (*@$\land$@*) (onLine b L) (*@$\land$@*) (onLine b M) (*@$\land$@*) (onLine c M) (*@$\land$@*) (onLine c N) (*@$\land$@*) (onLine d N) (*@$\land$@*) (b (*@$\neq$@*) c) (*@$\land$@*) (sameSide a d M) (*@$\land$@*) ((*@$\angle$@*) a:b:c) + ((*@$\angle$@*) b:c:d) < (*@$\rightangle$@*) + (*@$\rightangle$@*) (*@$\rightarrow$@*) (*@$\exists$@*) e : Point, (onLine e L) (*@$\land$@*) (onLine e N) (*@$\land$@*) (sameSide e a M)

degenerated_area_if : (*@$\forall$@*) (a b c : Point) (L : Line),  distinctPointsOnLine a b L (*@$\land$@*) (Triangle.area (*@$\bigtriangleup$@*) a:b:c) = 0 (*@$\rightarrow$@*)  onLine c L

degenerated_area_onlyif : (*@$\forall$@*) (a b c : Point) (L : Line), distinctPointsOnLine a b L (*@$\land$@*) (onLine c L) (*@$\rightarrow$@*) (Triangle.area (*@$\bigtriangleup$@*) a:b:c) = 0

sum_areas_if : (*@$\forall$@*) (a b c d : Point) (L : Line),  (onLine a L) (*@$\land$@*) (onLine b L) (*@$\land$@*) (onLine c L) (*@$\land$@*) (a (*@$\neq$@*) b) (*@$\land$@*) (a (*@$\neq$@*) c) (*@$\land$@*) (b (*@$\neq$@*) c) (*@$\land$@*) (*@$\lnot$@*)(onLine d L) (*@$\land$@*) (between a c b) (*@$\rightarrow$@*) (Triangle.area (*@$\bigtriangleup$@*) a:c:d + Triangle.area (*@$\bigtriangleup$@*) d:c:b = Triangle.area (*@$\bigtriangleup$@*) a:d:b)

sum_areas_onlyif : (*@$\forall$@*) (a b c d : Point) (L : Line), (onLine a L) (*@$\land$@*) (onLine b L) (*@$\land$@*) (onLine c L) (*@$\land$@*) (a (*@$\neq$@*) b) (*@$\land$@*) (a (*@$\neq$@*) c) (*@$\land$@*) (b (*@$\neq$@*) c) (*@$\land$@*) (*@$\lnot$@*)(onLine d L) (*@$\land$@*) (Triangle.area (*@$\bigtriangleup$@*) a:c:d + Triangle.area (*@$\bigtriangleup$@*) d:c:b = Triangle.area (*@$\bigtriangleup$@*) a:d:b) (*@$\rightarrow$@*) between a c b

parallelogram_area : (*@$\forall$@*) (a b c d : Point) (AB CD AC BD : Line), (formParallelogram a b c d AB CD AC BD) (*@$\rightarrow$@*) Triangle.area (*@$\bigtriangleup$@*) a:c:d + Triangle.area (*@$\bigtriangleup$@*) a:d:b = Triangle.area (*@$\bigtriangleup$@*) b:a:c + Triangle.area (*@$\bigtriangleup$@*) b:c:d

sum_parallelograms_area : (*@$\forall$@*) (a b c d e f : Point) (AB CD AC BD : Line), (formParallelogram a b c d AB CD AC BD) (*@$\land$@*) (between a e b) (*@$\land$@*) (between c f d) (*@$\rightarrow$@*) Triangle.area (*@$\bigtriangleup$@*) a:c:f + Triangle.area (*@$\bigtriangleup$@*) a:f:e + Triangle.area (*@$\bigtriangleup$@*) e:f:d + Triangle.area (*@$\bigtriangleup$@*) e:d:b = Triangle.area (*@$\bigtriangleup$@*) a:c:d + Triangle.area (*@$\bigtriangleup$@*) a:d:b

rectangle_area : (*@$\forall$@*) (a b c d : Point) (AB CD AC BD : Line), (formParallelogram a b c d AB CD AC BD) (*@$\land$@*) ((*@$\angle$@*) a:c:d = (*@$\rightangle$@*)) (*@$\rightarrow$@*) (Triangle.area (*@$\bigtriangleup$@*) a:c:d + Triangle.area (*@$\bigtriangleup$@*) a:b:d = |(a(*@--@*)b)| * |(a(*@--@*)c)|) (*@$\land$@*) (Triangle.area (*@$\bigtriangleup$@*) b:a:c + Triangle.area (*@$\bigtriangleup$@*) b:d:c = |(a(*@--@*)b)| * |(a(*@--@*)c)|)
\end{lstlisting}

\smallsec{Superposition Rule}
Finally, we have a superposition rule to handle Euclid's method of superposition for proving Proposition~4 and Proposition~8 in the first book of \emph{Elements}. The rule follows the idea in \citet{avigadeuclidnotes}.
\begin{lstlisting}[language=lean,numbers=none]
superposition : (*@$\forall$@*) (a b c d g h : Point) (AB BC AC L : Line), (formTriangle a b c AB BC AC) (*@$\land$@*) (distinctPointsOnLine d g L) (*@$\land$@*) (*@$\lnot$@*)(onLine h L) (*@$\rightarrow$@*) (*@$\exists$@*) (b' c' : Point) (BC' AC' : Line), ((*@$\angle$@*) b:a:c = (*@$\angle$@*) b':d:c') (*@$\land$@*) ((*@$\angle$@*) a:c:b = (*@$\angle$@*) d:c':b') (*@$\land$@*) ((*@$\angle$@*) c:b:a = (*@$\angle$@*) c':b':d) (*@$\land$@*) |(a(*@--@*)b)| = |(d(*@--@*)b')| (*@$\land$@*) |(b(*@--@*)c)| = |(b'(*@--@*)c')| (*@$\land$@*) |(c(*@--@*)a)| = |(c'(*@--@*)d)| (*@$\land$@*) (onLine b' L) (*@$\land$@*) (*@$\lnot$@*)(between b' d g) (*@$\land$@*) (sameSide c' h L) (*@$\land$@*) (distinctPointsOnLine b' c' BC') (*@$\land$@*) (distinctPointsOnLine d c' AC')
\end{lstlisting}

\smallsec{Axioms Used by the Symbolic Engine}

Our symbolic reasoning engine (Sec.~\ref{subsec:smt}) is based on SMT solvers. Therefore, it always has access to all built-in theories in SMT, e.g., the theory of real numbers. When used for diagrammatic reasoning, it can also use all diagrammatic rules, metric rules, and transfer rules. When used for equivalence checking, it additionally has access to nine simple construction rules: \texttt{arbitrary\_point}, \texttt{distinct\_points}, \texttt{line\_nonempty}, \texttt{exists\_point\_between\_points\_on\_line}, \texttt{distinct\_point\_same\_side}, \texttt{exists\_point\_opposite}, \texttt{exists\_point\_on\_circle}, \texttt{line\_from\_points}, and \texttt{intersection\_lines}.

\section{An Example of Logical Gaps in Euclid's Proofs Found by Formalization}
\label{appendix:prop_24}

Below is Euclid's proof of Proposition~24 from \citet{heiberg2007euclid}:

\begin{figure*}[ht]
    \begin{center}
    \includegraphics[width=0.35\linewidth]{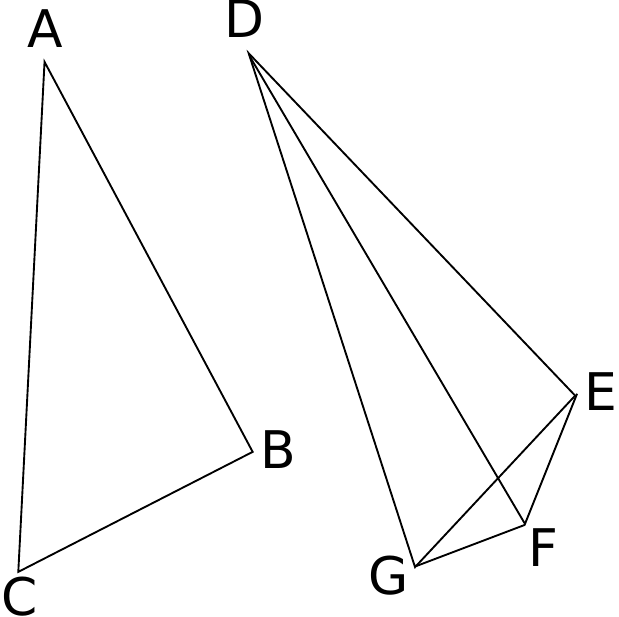}
    \end{center}
\end{figure*}

\begin{displayquote}
If two triangles have two sides equal to two sides, respectively,
but (one) has the angle encompassed by the equal straight-lines greater than the (corresponding)
angle (in the other), then (the former triangle) will also have a base greater than the base (of the latter).

Let  $ABC$ and $DEF$ be two triangles having the two sides $AB$ and $AC$
equal to the two sides $DE$ and $DF$, respectively. (That is), $AB$ (equal) to $DE$, and
$AC$ to $DF$.  Let them also have the angle at $A$ greater than the angle at $D$.
I say that the base $BC$ is also greater than the base $EF$.

For since angle $BAC$ is greater than angle $EDF$, let (angle) $EDG$, equal to
angle $BAC$,  have been constructed at the point $D$ on the straight-line $DE$ [Prop.~1.23]. And let $DG$ be made equal to either of $AC$ or $DF$ [Prop.~1.3], and let $EG$ and $FG$ have been joined.

Therefore, since $AB$ is equal to $DE$ and $AC$ to $DG$, the two (straight-lines)
$BA$, $AC$ are equal to the two (straight-lines) $ED$, $DG$, respectively.
Also the angle $BAC$ is equal to the angle $EDG$. Thus, the base $BC$ is equal
to the base $EG$ [Prop.~1.4]. Again, since $DF$ is equal to $DG$, angle $DGF$
is also equal to angle $DFG$ [Prop.~1.5]. Thus, $DFG$ (is) greater than $EGF$.
Thus, $EFG$ is much greater than $EGF$. And since triangle $EFG$ has angle $EFG$
greater than $EGF$, and the greater angle is subtended by the greater side [Prop.~1.19], side $EG$ (is) thus also greater than $EF$. But $EG$ (is) equal to
$BC$. Thus, $BC$ (is) also greater than $EF$.

Thus, if two triangles have two sides equal to two sides, respectively,
but (one) has the angle encompassed by the equal straight-lines greater than the 
(corresponding) angle (in the other), then (the former triangle) will also have a base greater than the base (of the latter).
(Which is) the very thing it was required to show.
\end{displayquote}

Euclid did not cover the case where $D$ and $G$ are on different sides of $EF$ (Fig.~\ref{fig:prop_24b}). For this case, we come up with the proof below, which is significantly more complicated than Euclid's proof. Therefore, we argue that Euclid's proof of Proposition~24 has a logical gap.

\begin{displayquote}
Similar to what Euclid did, let's construct $\triangle~EDG$ s.t., $\angle~EDG = \angle~BAC$, $|DG|=|DF|$, and $|BC|=|EG|$ (Fig.~\ref{fig:prop_24b}). By Proposition~5, we have $\angle~DGF = \angle~DFG$; let's denote it by $\alpha$. To prove $|BC| > |EF|$, we only need to prove $|EG| > |EF|$. By Proposition~19, this is further reduced to $\angle~EFG > \angle~EGF$. Let $x = \angle~EFG$ and $y = \angle~EGF$. We want to prove $x > y$.

Note that $\angle~DGE = \angle~DGF + \angle~EGF = \alpha + y$. $\angle~DFE = 2\pi - \angle~DFG - \angle~EFG = 2\pi - \alpha - x$. Furthermore, Proposition~17 states that the sum of any two angles in a triangle must be smaller than $\pi$. Therefore, any angle in a triangle must also be smaller than $\pi$, i.e.,
\begin{eqnarray*}
    \alpha + y & < & \pi \\
    2\pi - \alpha - x & < & \pi
\end{eqnarray*}
Simplifying these two inequalities leads to $x > y$. QED.
\end{displayquote}

\begin{figure*}[ht]
    \begin{center}
    \includegraphics[width=0.35\linewidth]{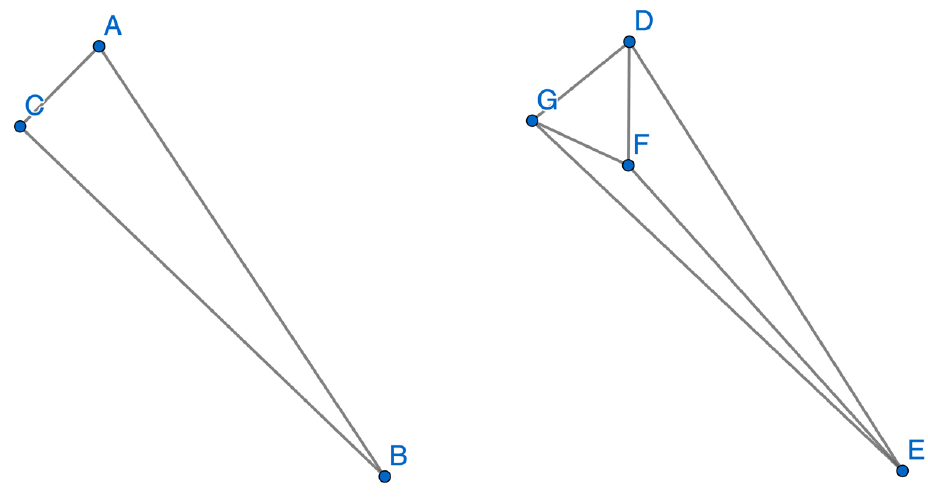}
    \caption{The case in Proposition~24 missed by Euclid.}
    \label{fig:prop_24b}
    \end{center}
\end{figure*}

Below is our formal proof of Proposition~24, which explicitly handles the case in Fig.~\ref{fig:prop_24b} following the informal proof above. 
\begin{lstlisting}[language=lean,numbers=none]
theorem proposition_24 : (*@$\forall$@*) (a b c d e f : Point) (AB BC AC DE EF DF : Line),
  formTriangle a b c AB BC AC (*@$\land$@*) formTriangle d e f DE EF DF (*@$\land$@*)
  (|(a(*@--@*)b)| = |(d(*@--@*)e)|) (*@$\land$@*) (|(a(*@--@*)c)| = |(d(*@--@*)f)|) (*@$\land$@*) ((*@$\angle$@*) b:a:c > (*@$\angle$@*) e:d:f) (*@$\rightarrow$@*)
  |(b(*@--@*)c)| > |(e(*@--@*)f)| :=
by
  euclid_intros
  euclid_apply (proposition_23' d e a b c f DE AB AC) as g'
  euclid_apply (line_from_points d g') as DG
  euclid_apply (extend_point_longer DG d g' (a(*@--@*)c)) as g''
  euclid_apply (proposition_3 d g'' a c DG AC) as g
  euclid_apply (line_from_points e g) as EG
  euclid_apply (line_from_points f g) as FG
  euclid_apply (proposition_4 a b c d e g AB BC AC DE EG DG)
  euclid_apply (proposition_5' d g f DG FG DF)
  by_cases (d.sameSide g EF)
  . euclid_assert ((*@$\angle$@*) d:f:g > (*@$\angle$@*) e:g:f)
    euclid_assert ((*@$\angle$@*) e:f:g > (*@$\angle$@*) e:g:f)
    euclid_apply (proposition_19 e f g EF FG EG)
    euclid_finish
  . -- Omitted by Euclid.
    by_cases g.onLine EF
    . euclid_finish
    . euclid_apply (extend_point FG g f) as h
      euclid_assert (*@$\lnot$@*)(g.onLine DF)
      euclid_assert (*@$\lnot$@*)(e.onLine DF)
      euclid_assert (g.opposingSides e DF)
      euclid_assert h.sameSide e DF
      euclid_apply (proposition_13 d f g h DF FG)
      euclid_apply (proposition_13 e f g h EF FG)
      euclid_apply (proposition_17 d g e DG EG DE)
      euclid_apply (proposition_17 d f e DF EF DE)
      euclid_assert ((*@$\angle$@*) d:g:e < (*@$\rightangle$@*) + (*@$\rightangle$@*))
      euclid_assert ((*@$\angle$@*) d:f:e < (*@$\rightangle$@*) + (*@$\rightangle$@*))
      euclid_assert ((*@$\angle$@*) e:f:g + (*@$\angle$@*) g:f:d + (*@$\angle$@*) d:f:e = (*@$\rightangle$@*) + (*@$\rightangle$@*) + (*@$\rightangle$@*) + (*@$\rightangle$@*))
      euclid_assert ((*@$\angle$@*) e:f:g > (*@$\angle$@*) e:g:f)
      euclid_apply (proposition_19 e f g EF FG EG)
      euclid_finish
\end{lstlisting}

\section{Comparison to Existing Formalization of \emph{Elements}}
\label{appendix:comparison}

We use Proposition~1 as an example to compare {\dataname} with existing formalizations of Euclid's \emph{Elements}~\citep{hernandezespiet2023e,beeson2019proof}. Below is Euclid's proof of Proposition~1 from \citet{heiberg2007euclid}:

\begin{figure*}[ht]
    \begin{center}
    \includegraphics[width=0.35\linewidth]{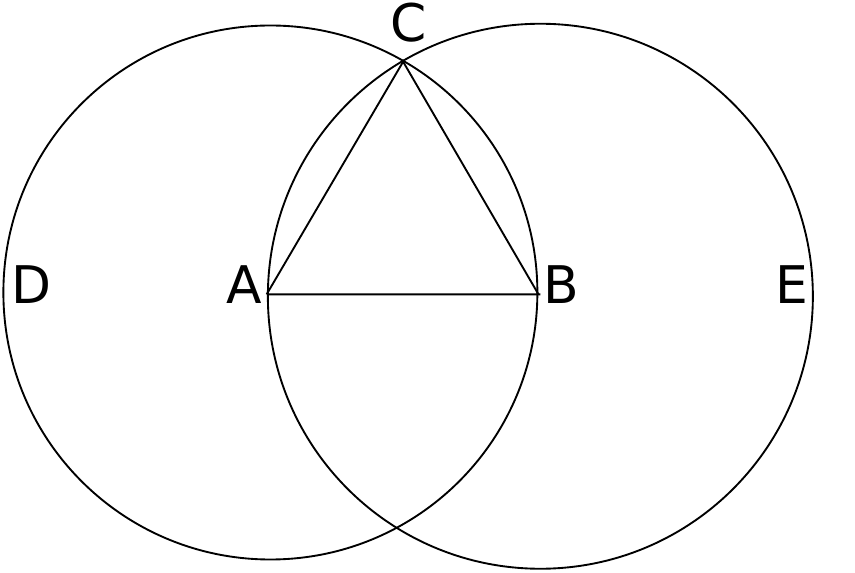}
    \label{fig:prop_1}
    \end{center}
\end{figure*}

\begin{displayquote}
To construct an equilateral triangle on a given finite straight-line.

Let $AB$ be the given finite straight-line. 

So it is required to construct an equilateral triangle on the straight-line $AB$.

Let the circle $BCD$ with center $A$ and radius $AB$ have been drawn [Post.3], and again let the circle $ACE$ with center $B$ and radius $BA$ have been drawn [Post.3].
And let the straight-lines $CA$ and $CB$ have been joined from the point $C$, where the circles cut one another, to the points $A$ and $B$ (respectively) [Post.1].

And since the point $A$ is the center of the circle $CDB$, $AC$ is equal to $AB$ [Def.5]. Again,
since the point $B$ is the center of the circle $CAE$, $BC$ is equal to $BA$ [Def.5]. But $CA$ 
was also shown (to be) equal to $AB$. Thus, $CA$ and $CB$ are each equal to $AB$. But things equal to the same thing are also equal to one another [C.N.1]. Thus, $CA$ is also equal to $CB$. Thus, the three (straight-lines) $CA$, $AB$, and $BC$ are equal to one another.

Thus, the triangle $ABC$ is equilateral, and has been constructed on the
given finite straight-line $AB$. (Which is) the very thing it was required to do.
\end{displayquote}

\smallsec{\dataname}
Below is our formalization of Proposition~1. Our proof is concise and follows Euclid's proof closely. It is straightforward to map the tactics in our proofs to sentences in Euclid's proof.
\begin{lstlisting}[language=lean,numbers=none]
theorem proposition_1 : (*@$\forall$@*) (a b : Point) (AB : Line),
  distinctPointsOnLine a b AB (*@$\rightarrow$@*)
  (*@$\exists$@*) c : Point, |(c(*@--@*)a)| = |(a(*@--@*)b)| (*@$\land$@*) |(c(*@--@*)b)| = |(a(*@--@*)b)| :=
by
  euclid_intros
  euclid_apply circle_from_points a b as BCD
  euclid_apply circle_from_points b a as ACE
  euclid_apply intersection_circles BCD ACE as c
  euclid_apply point_on_circle_onlyif a b c BCD
  euclid_apply point_on_circle_onlyif b a c ACE
  use c
  euclid_finish
\end{lstlisting}

\smallsec{\citet{hernandezespiet2023e}}
Below is \citet{hernandezespiet2023e}'s formalization. In this formalization, all diagrammatic inferences must be handled explicitly. For instance, in the proof of Proposition~1 (\texttt{iseqtri\_sameside\_of\_ne}), the proof that circles BCD and ACE intersect  
``\texttt{(circlesInter\_of\_inside\_on\_circle b$\alpha$ a$\beta$ (inside\_circle\_of\_center a$\alpha$) (inside\_circle\_of\_center b$\beta$))}''  must be supplied directly by the proof author, even though this assertion is left implicit in the text. In contrast, we use the symbolic engine to prove it automatically and implicitly. 
\begin{lstlisting}[language=lean,numbers=none]
theorem online_of_circlesinter (a(*@$\alpha$@*) : CenterCircle a (*@$\alpha$@*)) (b(*@$\beta$@*) : CenterCircle b (*@$\beta$@*))
    ((*@$\alpha$@*)(*@$\beta$@*) : CirclesInter (*@$\alpha$@*) (*@$\beta$@*)) : (*@$\exists$@*) c L, OnLine a L (*@$\land$@*) OnLine b L (*@$\land$@*) OnCircle c (*@$\alpha$@*) (*@$\land$@*)
    OnCircle c (*@$\beta$@*) (*@$\land$@*) (*@$\lnot$@*)OnLine c L := by
rcases line_of_pts a b with (*@$\langle$@*)L, aL, bL(*@$\rangle$@*); rcases not_online_of_line L with (*@$\langle$@*)d, dL(*@$\rangle$@*);
  rcases pt_sameSide_of_circlesInter aL bL dL a(*@$\alpha$@*) b(*@$\beta$@*) (*@$\alpha$@*)(*@$\beta$@*) with (*@$\langle$@*)c, cdL, c(*@$\alpha$@*), c(*@$\beta$@*)(*@$\rangle$@*);
  exact (*@$\langle$@*)c, L, aL, bL, c(*@$\alpha$@*), c(*@$\beta$@*), not_onLine_of_sameSide cdL(*@$\rangle$@*)

theorem DiffSide_of_sameside_DiffSide (abL : SameSide a b L) (acL : DiffSide a c L) :
    DiffSide b c L := by
by_contra h; unfold DiffSide at h; push_neg at h; exact acL.2.2
 (sameSide_trans (sameSide_symm abL) (h (not_onLine_of_sameSide (sameSide_symm abL)) acL.2.1))

theorem DiffSide_of_circlesinter (a(*@$\alpha$@*) : CenterCircle a (*@$\alpha$@*)) (b(*@$\beta$@*) : CenterCircle b (*@$\beta$@*))
    ((*@$\alpha$@*)(*@$\beta$@*) : CirclesInter (*@$\alpha$@*) (*@$\beta$@*)) : (*@$\exists$@*) c d L, OnLine a L (*@$\land$@*) OnLine b L (*@$\land$@*) OnCircle c (*@$\alpha$@*) (*@$\land$@*)
    OnCircle c (*@$\beta$@*) (*@$\land$@*) OnCircle d (*@$\alpha$@*) (*@$\land$@*) OnCircle d (*@$\beta$@*) (*@$\land$@*) DiffSide c d L := by
rcases online_of_circlesinter a(*@$\alpha$@*) b(*@$\beta$@*) (*@$\alpha$@*)(*@$\beta$@*) with (*@$\langle$@*)c, L, aL, bL, c(*@$\alpha$@*), c(*@$\beta$@*), cL(*@$\rangle$@*);
rcases diffSide_of_not_onLine cL with (*@$\langle$@*)e, eL, ceL(*@$\rangle$@*); rcases pt_sameSide_of_circlesInter aL bL eL
 a(*@$\alpha$@*) b(*@$\beta$@*) (*@$\alpha$@*)(*@$\beta$@*) with (*@$\langle$@*)d, deL, d(*@$\alpha$@*), d(*@$\beta$@*)(*@$\rangle$@*); exact (*@$\langle$@*)c, d, L, aL, bL, c(*@$\alpha$@*), c(*@$\beta$@*), d(*@$\alpha$@*), d(*@$\beta$@*), DiffSide_symm
 (DiffSide_of_sameside_DiffSide (sameSide_symm deL) (*@$\langle$@*)eL, cL, not_sameSide_symm ceL(*@$\rangle$@*))(*@$\rangle$@*)

/-- Euclid I.1, construction of two equilateral Triangles -/
theorem iseqtri_iseqtri_DiffSide_of_ne (ab : a (*@$\neq$@*) b) : (*@$\exists$@*) c d L, OnLine a L (*@$\land$@*)
    OnLine b L (*@$\land$@*) DiffSide c d L (*@$\land$@*) EqTri a b c (*@$\land$@*) EqTri a b d := by
rcases circle_of_ne ab with (*@$\langle$@*)(*@$\alpha$@*), a(*@$\alpha$@*), b(*@$\alpha$@*)(*@$\rangle$@*)
rcases circle_of_ne (Ne.symm ab) with (*@$\langle$@*)(*@$\beta$@*), b(*@$\beta$@*), a(*@$\beta$@*)(*@$\rangle$@*)
rcases DiffSide_of_circlesinter a(*@$\alpha$@*) b(*@$\beta$@*) (circlesInter_of_inside_on_circle b(*@$\alpha$@*) a(*@$\beta$@*)
  (inside_circle_of_center a(*@$\alpha$@*)) (inside_circle_of_center b(*@$\beta$@*))) with
  (*@$\langle$@*)c, d, L, aL, bL, c(*@$\alpha$@*), c(*@$\beta$@*), d(*@$\alpha$@*), d(*@$\beta$@*), cdL(*@$\rangle$@*)
have ab_ac := (on_circle_iff_length_eq a(*@$\alpha$@*) b(*@$\alpha$@*)).mpr c(*@$\alpha$@*)
have bc_ba := (on_circle_iff_length_eq b(*@$\beta$@*) c(*@$\beta$@*)).mpr a(*@$\beta$@*)
have ab_ad := (on_circle_iff_length_eq a(*@$\alpha$@*) b(*@$\alpha$@*)).mpr d(*@$\alpha$@*)
have bd_ba := (on_circle_iff_length_eq b(*@$\beta$@*) d(*@$\beta$@*)).mpr a(*@$\beta$@*)
exact (*@$\langle$@*)c, d, L, aL, bL, cdL, EqTri_of_length_online ab aL bL cdL.1 ab_ac bc_ba,
  EqTri_of_length_online ab aL bL cdL.2.1 ab_ad bd_ba(*@$\rangle$@*)

/-- Euclid I.1, construction of an equilateral Triangle on the sameside of a point -/
theorem iseqtri_sameside_of_ne (ab : a (*@$\neq$@*) b) (aL : OnLine a L) (bL : OnLine b L) (dL : (*@$\lnot$@*)OnLine d L):
    (*@$\exists$@*) c, (*@$\lnot$@*)OnLine c L (*@$\land$@*) SameSide c d L (*@$\land$@*) EqTri a b c := by
  rcases iseqtri_iseqtri_DiffSide_of_ne ab with (*@$\langle$@*)c1, c2, M, aM, bM, c1c2M, eqtri1, eqtri2(*@$\rangle$@*)
  rcases sameSide_or_of_diffSide' dL (by rwa [line_unique_of_pts ab aM bM aL bL] at c1c2M)
    with c1dL | c2dL
  refine (*@$\langle$@*)c1, not_onLine_of_sameSide c1dL, c1dL, eqtri1(*@$\rangle$@*)
  refine (*@$\langle$@*)c2, not_onLine_of_sameSide c2dL, c2dL, eqtri2(*@$\rangle$@*)

/-- Euclid I.1, construction of a single equilateral Triangle -/
theorem iseqtri_of_ne (ab : a (*@$\neq$@*) b) : (*@$\exists$@*) c, EqTri a b c :=
  by rcases iseqtri_iseqtri_DiffSide_of_ne ab with (*@$\langle$@*)c, -, -, -, -, -, eqtri, -(*@$\rangle$@*); exact (*@$\langle$@*)c, eqtri(*@$\rangle$@*)
\end{lstlisting}

\smallsec{\citet{beeson2019proof}}
Below is \citet{beeson2019proof}'s formalization in Coq. Similar to \citet{hernandezespiet2023e}, it formalizes diagrammatic reasoning manually. As a result, their formal proofs tend to be longer than ours, and it is harder discern the correspondence with Euclid's proofs.
\begin{lstlisting}[language=lean,numbers=none]
Lemma proposition_01 : 
   forall A B, 
   neq A B ->
   exists X, equilateral A B X /\ Triangle A B X.
Proof.
intros.
let Tf:=fresh in
assert (Tf:exists J, CI J A A B) by (conclude postulate_Euclid3);destruct Tf as [J];spliter.
assert (neq B A) by (conclude lemma_inequalitysymmetric).
let Tf:=fresh in
assert (Tf:exists K, CI K B B A) by (conclude postulate_Euclid3);destruct Tf as [K];spliter.
let Tf:=fresh in
assert (Tf:exists D, (BetS B A D /\ Cong A D A B)) by (conclude lemma_localextension);destruct Tf as [D];spliter.
assert (Cong B A B A) by (conclude cn_congruencereflexive).
assert (OutCirc D K) by (conclude_def OutCirc) .
assert (eq B B) by (conclude cn_equalityreflexive).
assert (InCirc B K) by (conclude_def InCirc ).
assert (Cong A B A B) by (conclude cn_congruencereflexive).
assert (OnCirc B J) by (conclude_def OnCirc ).
assert (OnCirc D J) by (conclude_def OnCirc ).
assert (eq A A) by (conclude cn_equalityreflexive).
assert (InCirc A J) by (conclude_def InCirc ).
let Tf:=fresh in
assert (Tf:exists C, (OnCirc C K /\ OnCirc C J)) by (conclude postulate_circle_circle);destruct Tf as [C];spliter.
assert (Cong A C A B) by (conclude axiom_circle_center_radius).
assert (Cong A B A C) by (conclude lemma_congruencesymmetric).
assert (Cong B C B A) by (conclude axiom_circle_center_radius). 
assert (Cong B C A B) by (forward_using lemma_congruenceflip). 
assert (Cong B C A C) by (conclude lemma_congruencetransitive).
assert (Cong A B B C) by (conclude lemma_congruencesymmetric).
assert (Cong A C C A) by (conclude cn_equalityreverse).
assert (Cong B C C A) by (conclude lemma_congruencetransitive).
assert (equilateral A B C) by (conclude_def equilateral ).
assert (neq B C) by (conclude axiom_nocollapse).
assert (neq C A) by (conclude axiom_nocollapse).
assert (~ BetS A C B).
 {
 intro.
 assert (~ Cong A C A B) by (conclude lemma_partnotequalwhole).
 assert (Cong C A A C) by (conclude cn_equalityreverse).
 assert (Cong C A A B) by (conclude lemma_congruencetransitive).
 assert (Cong A C C A) by (conclude cn_equalityreverse).
 assert (Cong A C A B) by (conclude lemma_congruencetransitive).
 contradict.
 }
assert (~ BetS A B C).
 {
 intro.
 assert (~ Cong A B A C) by (conclude lemma_partnotequalwhole).
 assert (Cong A B C A) by (conclude lemma_congruencetransitive).
 assert (Cong C A A C) by (conclude cn_equalityreverse).
 assert (Cong A B A C) by (conclude lemma_congruencetransitive).
 contradict.
 }
assert (~ BetS B A C).
 {
 intro.
 assert (~ Cong B A B C) by (conclude lemma_partnotequalwhole).
 assert (Cong B A A B) by (conclude cn_equalityreverse).
 assert (Cong B A B C) by (conclude lemma_congruencetransitive).
 contradict.
 }
assert (~ Col A B C).
 {
 intro.
 assert (neq A C) by (conclude lemma_inequalitysymmetric).
 assert ((eq A B \/ eq A C \/ eq B C \/ BetS B A C \/ BetS A B C \/ BetS A C B)) by (conclude_def Col ).
 contradict.
 }
assert (Triangle A B C) by (conclude_def Triangle ).
close.
Unshelve.
all: (exact A).
Qed.
\end{lstlisting}

\section{Autoformalization Prompts}
\label{appendix:prompts}
Here, we share the prompt templates used in our experiments. 

\subsection{Theorem Statement Formalization}
To autoformalize theorem statements from \emph{Elements} and UniGeo, we used the prompt shown below. We also prepend the instructions with a header describing the task (i.e., ``Your task is to take an English statement of a theorem from Euclidean Geometry and formalize it using Lean 4 programming language, adhering to the following structures and guidelines''.

One additional intricacy in formalizing theorem statements from \emph{Elements} is that, unlike in modern mathematics, Euclid does not always state a complete logical theorem statement before beginning a proof. He often begins with an imperative assertion (e.g., Proposition 9: ``To cut a given rectilinear angle in half'') whose proper logical interpretation is clarified at the conclusion of the proof (e.g., ``Thus, angle DAF is equal to
angle EAF. Thus, the given rectilinear angle BAC has been cut in
half by the straight-line AF''). In other words, to determine the proper logical statement of the theorem, both the beginning \emph{and} the end of the proof must sometimes be considered. During preliminary experiments, we attempted to autoformalize theorem statements by giving the model the entire natural language proof, but we found that the resulting theorem statements were often very poor. Instead, we manually removed the majority of the ``proof'' portion of the natural language text, leaving only the portions required to obtain a reasonable formalization of the theorem statement. This context was also included as part of the prompt header when formalizing theorems from \emph{Elements}. These abbreviated versions of \emph{Elements} proofs are also available as part of {\dataname}.
\begin{lstlisting}[language=lean, numbers=none]
/- Basic Geometric Sorts -/
axiom Point : Type
axiom Line : Type
axiom Circle : Type
/- Inductive Types for Geometric Entities -/
inductive Angle | right | ofPoints (A B C : Point)
inductive Segment | endpoints (a b : Point)
inductive Triangle | ofPoints (a b c : Point)
/- Notations and Macros for Geometric Entities -/
"|(a(*@--@*) b)|" means the length of the line segment between point a and point b.
"(*@$\angle$@*)  a:b:c" means the degree of the angle formed by three points a, b, and c.
"(*@$\rightangle$@*) " means the right angle.
"(*@$\triangle$@*)  a:b:c" means the triangle formed from points a, b and c.
"Triangle.area (*@$\triangle$@*)  a:b:c" means the area of the triangle formed by points a, b and c.
/- Relations and Axioms for Geometric Sorts -/
namespace Point
def onLine (a : Point) (L : Line) -- point a is on line L.
def sameSide (a b : Point) (L : Line) -- point a and b are on the same side of line L.
def opposingSides (a b : Point) (L : Line) -- point a and b are on the opposite sides of line L.
def collinear (a b c : Point) -- points a, b and c are collinear.
end Point
namespace Line
def intersectsLine (L M : Line) -- two lines L and M intersect at some point.
def intersectsCircle (L : Line) (C : Circle) -- line L and circle C intersect.
end Line
namespace Circle
def onCircle (a: Point) (C: Circle) -- point a is on circle C.
def insideCircle (a: Point) (C: Circle) -- point a is inside circle C.
def outsideCircle (a: Point) (C: Circle)-- point a is outside circle C.
def isCentre (a: Point) (C: Circle) -- point a is on the center circle C.
def intersectsCircle (C1 C2: Circle) -- circle C1 and C2 intersect.
end Circle
namespace Triangle
def congruent (T1 T2 : Triangle) -- triangle T1 and T2 are congruent.
def similar (T1 T2 : Triangle) -- triangle T1 and T2 are similar.
end Triangle
/- Geometric Relations -/
def distinctPointsOnLine (a b : Point) (L : Line) -- points a and b are distinct and on line L.
def twoLinesIntersectAtPoint (AB BC : Line) (b : Point) -- line AB and BC intersect at point b.
def between (a b c : Point) -- points a, b and c collinear and cyclically ordered.
def formTriangle (a b c : Point) (AB BC CA : Line) -- point a, b and c form a triangle, where point a and b are on line AB, point b and c are on line BC, point a and c are on line CA.
def formRectilinearAngle (a b c : Point) (AB BC : Line)  -- point a, b and c form a rectilinear angle where a(*@$\neq$@*)b and b(*@$\neq$@*)c.
def formParallelogram (a b c d : Point) (AB CD AC BD : Line) -- point a, b, d and c (in clockwise/counterclockwise order) form a parallelogram, where point a and b are on line AB,  where point c and d are on line CD,  where point a and c are on line AC,  where point b and d are on line BD.
def formQuadrilateral (a b c d : Point) (AB CD BC AD : Line)  -- point a, b, c and d (in clockwise/counterclockwise order) form a quadrilateral, where point a and b are on line AB,  where point c and d are on line CD,  where point b and c are on line BC,  where point a and d are on line AD.
/-  Guidelines: -/
1. Proposition Format: Your proposition must be of the form <<< (*@$\forall$@*)  (...) P_1 (*@$\land$@*)  P_2 ... (*@$\land$@*)  P_n (*@$\to$@*)  (*@$\exists$@*)  (...) Q_1 (*@$\land$@*) Q_2 ... (*@$\land$@*) Q_m >>> where where each P_i and Q_i is built from the above building blocks using conjunction ((*@$\land$@*)) disjunction ((*@$\lor$@*)) and negation ((*@$\neg$@*)). Note that there may be zero existentially quantified variables. 
2. Implication: There can be only a single implication in the formula; either side of the implication must be a conjunction of formulae.
3. Numeric Values Restrictions: Denote 90-degree angle by (*@$\rightangle$@*), 180-degree angle by (*@$\rightangle$@*) + (*@$\rightangle$@*), etc. Also, when referring to segments, we always mean its length (i.e. |(a(*@--@*)b)|).
4. Quantified Variables: Your quantified variables must be limited to primitive geometric types: points, lines, and circles. ALL bound variables must be mentioned at some point.
5. Intermediate Variables: You should never define an intermediate variable inside the proposition. For example, "let X := (something);" is NOT allowed.
6. Axioms: You should only use the provided axioms. For example, Line L is parallel to line M can be expressed as (*@$\neg$@*)(L.intersectsLine M). Do not use Line.Parallel L M.
7. Response Format: Present your formalized Lean expression within triple angle brackets (<<< Lean expression here >>>). Do not add any annotations/explanations, or markdown syntax.
\end{lstlisting}

\subsection{Proof Formalization (\emph{Elements})} 
To formalize proofs from \emph{Elements}, we included (1) all inference rules that were used in our manual formalization of \emph{Elements} and (2) all variations of all theorem statements that precede the target theorem.  We also provided 5 hardcoded examples (Propositions 2, 6, 12, 32, 42) which were chosen to showcase a variety of proof formats and tactic invocations. The token $\{{\tt PRECEDING\_THMS}\}$ is substituted at formalization time by the list of visible theorems allowed for the target proposition. For instance, when formalizing a proof of Proposition $n$ in \emph{Elements}, the model is allowed to use proofs of Propositions 1 through $n-1$, and all their variations.

\begin{lstlisting}[language=lean,numbers=none]
/- Basic Geometric Sorts -/
axiom Point : Type
axiom Line : Type
axiom Circle : Type

/- Notations and Macros for Geometric Entities -/
"|(a(*@--@*)b)|" means the length of the line segment between point a and point b.
"(*@$\angle$@*) a:b:c" means the degree of the angle formed by points a, b, and c.
"(*@$\rightangle$@*)" means the right angle.
"(*@$\triangle$@*) a:b:c" means the triangle formed from points a, b and c.
"Triangle.area (*@$\triangle$@*) a:b:c" means the area of the triangle formed by points a, b and c.
/- Relations and Axioms for Geometric Sorts -/
namespace Point
def onLine (a : Point) (L : Line) -- point a is on line L.
def sameSide (a b : Point) (L : Line) -- point a and b are on the same side of line L.
def opposingSides (a b : Point) (L : Line) -- point a and b are on the opposite sides of line L.
def collinear : (a b c : Point) -- points a, b and c are collinear.
end Point
namespace Line
def intersectsLine (L M : Line) -- two lines L and M intersect at some point.
def intersectsCircle (L : Line) (C : Circle) -- line L and circle C intersect.
end Line
namespace Circle
def onCircle (a: Point) (C: Circle) -- point a is on circle C.
def insideCircle (a: Point) (C: Circle) -- point a is inside circle C.
def outsideCircle (a: Point) (C: Circle)-- point a is outside circle C.
def isCentre (a: Point) (C: Circle) -- point a is on the center circle C.
def intersectsCircle (C1 C2: Circle) -- circle C1 and C2 intersect.
end Circle

-- Abbreviations --
abbrev distinctPointsOnLine (p q : Point) (L : Line) := 
   P.onLine L (*@$\land$@*) Q.onLine L (*@$\land$@*) P (*@$\neq$@*) Q


abbrev Point.opposingSides (a b : Point) (l : Line) :=
  (*@$\neg$@*) a.onLine l  (*@$\land$@*) (*@$\neg$@*) b.onLine l (*@$\land$@*) (*@$\neg$@*) sameSide a b l

abbrev Point.outsideCircle (p c : Point) :=
   (*@$\neg$@*) p.insideCircle c (*@$\land$@*) (*@$\neg$@*) p.onCircle c

abbrev formTriangle (a b c : Point) (AB BC CA : Line) : Prop :=
  distinctPointsOnLine a b AB (*@$\land$@*)
  b.onLine BC (*@$\land$@*) c.onLine BC (*@$\land$@*) c.onLine CA (*@$\land$@*) a.onLine CA (*@$\land$@*)
  AB (*@$\neq$@*) BC (*@$\land$@*) BC (*@$\neq$@*) CA (*@$\land$@*) CA (*@$\neq$@*) AB

abbrev formRectilinearAngle (a b c : Point) (AB BC : Line) :=
  distinctPointsOnLine a b AB (*@$\land$@*) distinctPointsOnLine b c BC

abbrev formParallelogram (a b c d : Point) (AB CD AC BD : Line) : Prop :=
    a.onLine AB (*@$\land$@*) b.onLine AB (*@$\land$@*) c.onLine CD (*@$\land$@*) d.onLine CD (*@$\land$@*) a.onLine AC (*@$\land$@*) c.onLine AC (*@$\land$@*) distinctPointsOnLine b d BD (*@$\land$@*)
    (a.sameSide c BD) (*@$\land$@*) (*@$\neg$@*)(AB.intersectsLine CD) (*@$\land$@*) (*@$\neg$@*)(AC.intersectsLine BD)

Your proofs can make use of the following construction rules and inference rules:
Construction Rules:
- Points -
axiom distinct_points : (*@$\forall$@*) p : Point, (*@$\exists$@*) q : Point, p (*@$\neq$@*) q
axiom line_nonempty : (*@$\forall$@*) l : Line, (*@$\exists$@*) p : Point, p.onLine l
axiom exists_distincts_points_on_line : (*@$\forall$@*) l : Line, (*@$\forall$@*) p : Point, (*@$\exists$@*) p' : Point, p (*@$\neq$@*) p' (*@$\land$@*) p'.onLine l
axiom exists_point_between_points_on_line : (*@$\forall$@*) (L : Line) (b c : Point), distinctPointsOnLine b c L (*@$\to$@*) (*@$\exists$@*) a : Point, (a.onLine L) (*@$\land$@*) (between b a c)
axiom point_between_points_shorter_than : (*@$\forall$@*) (L : Line) (b c : Point) (s : Segment), distinctPointsOnLine b c L (*@$\land$@*) (|s| > 0) (*@$\to$@*) (*@$\exists$@*) a : Point, (a.onLine L) (*@$\land$@*) (between b a c) (*@$\land$@*) (|(b(*@--@*)a)| < s)
axiom extend_point : (*@$\forall$@*) (L : Line) (b c : Point), distinctPointsOnLine b c L (*@$\to$@*) (*@$\exists$@*) a : Point, (a.onLine L) (*@$\land$@*) (between b c a)
axiom extend_point_not_on_line : (*@$\forall$@*) (L M : Line) (b c : Point), distinctPointsOnLine b c L (*@$\land$@*) L (*@$\neq$@*) M (*@$\to$@*) (*@$\exists$@*) a : Point, (a.onLine L) (*@$\land$@*) (between b c a) (*@$\land$@*) (*@$\neg$@*)(a.onLine M)
axiom extend_point_longer : (*@$\forall$@*) (L : Line) (b c : Point) (s : Segment), distinctPointsOnLine b c L (*@$\to$@*) (*@$\exists$@*) a : Point, (a.onLine L) (*@$\land$@*) (between b c a) (*@$\land$@*) (|(c(*@--@*)a)| > s)
axiom point_on_line_same_side : (*@$\forall$@*) (L M : Line) (b : Point), (*@$\neg$@*)(b.onLine L) (*@$\land$@*) (L.intersectsLine M) (*@$\to$@*) (*@$\exists$@*) a : Point, a.onLine M (*@$\land$@*) a.sameSide b L
axiom exists_point_opposite : (*@$\forall$@*) (L : Line) (b : Point), (*@$\neg$@*)(b.onLine L) (*@$\to$@*) (*@$\exists$@*) a : Point, a.opposingSides b L
- Lines and Circles -
axiom line_from_points : (*@$\forall$@*) (a b : Point), a (*@$\neq$@*) b (*@$\to$@*) (*@$\exists$@*) L : Line, (a.onLine L) (*@$\land$@*) (b.onLine L)
axiom circle_from_points : (*@$\forall$@*) (a b : Point), a (*@$\neq$@*) b (*@$\to$@*) (*@$\exists$@*) (*@$\alpha$@*) : Circle, (a.isCentre (*@$\alpha$@*)) (*@$\land$@*) (b.onCircle (*@$\alpha$@*))
- Intersections -
axiom intersection_lines : (*@$\forall$@*) (L M : Line), L.intersectsLine M (*@$\to$@*) (*@$\exists$@*) a : Point, (a.onLine L) (*@$\land$@*) (a.onLine M)
axiom intersection_circle_line : (*@$\forall$@*) ((*@$\alpha$@*) : Circle) (L : Line), L.intersectsCircle (*@$\alpha$@*) (*@$\to$@*) (*@$\exists$@*) a : Point, (a.onCircle (*@$\alpha$@*)) (*@$\land$@*) (a.onLine L)
axiom intersections_circle_line : (*@$\forall$@*) ((*@$\alpha$@*) : Circle) (L : Line), L.intersectsCircle (*@$\alpha$@*) (*@$\to$@*) (*@$\exists$@*) (a b : Point), (a.onCircle (*@$\alpha$@*)) (*@$\land$@*) (a.onLine L) (*@$\land$@*) (b.onCircle (*@$\alpha$@*)) (*@$\land$@*) (b.onLine L) (*@$\land$@*) a (*@$\neq$@*) b
axiom intersection_circle_line_between_points : (*@$\forall$@*) ((*@$\alpha$@*) : Circle) (L : Line) (b c :Point), (b.insideCircle (*@$\alpha$@*)) (*@$\land$@*) (b.onLine L) (*@$\land$@*) (c.outsideCircle (*@$\alpha$@*)) (*@$\land$@*) (c.onLine L) (*@$\to$@*) (*@$\exists$@*) a : Point, (a.onCircle (*@$\alpha$@*)) (*@$\land$@*) (a.onLine L) (*@$\land$@*) (between b a c)
axiom intersection_circle_line_extending_points : (*@$\forall$@*) ((*@$\alpha$@*) : Circle) (L : Line) (b c :Point), (b.insideCircle (*@$\alpha$@*)) (*@$\land$@*) distinctPointsOnLine b c L (*@$\to$@*) (*@$\exists$@*) a : Point, (a.onCircle (*@$\alpha$@*)) (*@$\land$@*) (a.onLine L) (*@$\land$@*) (between a b c)
axiom intersection_circles : (*@$\forall$@*) ((*@$\alpha$@*) (*@$\beta$@*) : Circle), (*@$\alpha$@*).intersectsCircle (*@$\beta$@*) (*@$\to$@*) (*@$\exists$@*) a : Point, (a.onCircle (*@$\alpha$@*)) (*@$\land$@*) (a.onCircle (*@$\beta$@*))
axiom intersection_same_side : (*@$\forall$@*) ((*@$\alpha$@*) (*@$\beta$@*) : Circle) (b c d : Point) (L : Line), ((*@$\alpha$@*).intersectsCircle (*@$\beta$@*)) (*@$\land$@*) (c.isCentre (*@$\alpha$@*)) (*@$\land$@*) (d.isCentre (*@$\beta$@*)) (*@$\land$@*) (c.onLine L) (*@$\land$@*) (d.onLine L) (*@$\land$@*) (*@$\neg$@*)(b.onLine L) (*@$\to$@*) (*@$\exists$@*) a : Point, (a.onCircle (*@$\alpha$@*)) (*@$\land$@*) (a.onCircle (*@$\beta$@*)) (*@$\land$@*) (a.sameSide b L)
axiom intersection_opposite_side : (*@$\forall$@*) ((*@$\alpha$@*) (*@$\beta$@*) : Circle) (b c d : Point) (L : Line), ((*@$\alpha$@*).intersectsCircle (*@$\beta$@*)) (*@$\land$@*) (c.isCentre (*@$\alpha$@*)) (*@$\land$@*) (d.isCentre (*@$\beta$@*)) (*@$\land$@*) (c.onLine L) (*@$\land$@*) (d.onLine L) (*@$\land$@*) (*@$\neg$@*)(b.onLine L) (*@$\to$@*) (*@$\exists$@*) a : Point, (a.onCircle (*@$\alpha$@*)) (*@$\land$@*) (a.onCircle (*@$\beta$@*)) (*@$\land$@*) a.opposingSides b L
--- Inference Rules ---
axiom two_points_determine_line : (*@$\forall$@*) (a b : Point) (L M : Line),distinctPointsOnLine a b L (*@$\land$@*) (a.onLine M) (*@$\land$@*) (b.onLine M) (*@$\to$@*) L = M
axiom superposition : (*@$\forall$@*) (a b c d g h : Point) (AB BC AC L : Line), formTriangle a b c AB BC AC (*@$\land$@*) distinctPointsOnLine d g L (*@$\land$@*) (*@$\neg$@*)(h.onLine L) (*@$\to$@*) (*@$\exists$@*) (b' c' : Point) (BC' AC' : Line), ((*@$\angle$@*) b:a:c ) = ((*@$\angle$@*) b':d:c') (*@$\land$@*) ((*@$\angle$@*) a:c:b) = ((*@$\angle$@*) d:c':b') (*@$\land$@*) ((*@$\angle$@*) c:b:a) = ((*@$\angle$@*) c':b':d) (*@$\land$@*) |(a(*@--@*)b)| = |(d(*@--@*)b')| (*@$\land$@*) |(b(*@--@*)c)| = |(b'(*@--@*)c')| (*@$\land$@*) |(c(*@--@*)a)| = |(c'(*@--@*)d)| (*@$\land$@*) b'.onLine L (*@$\land$@*) (*@$\neg$@*)(between b' d g) (*@$\land$@*) c'.sameSide h L (*@$\land$@*) distinctPointsOnLine b' c' BC' (*@$\land$@*) distinctPointsOnLine d c' AC'
axiom between_if : (*@$\forall$@*) (a b c : Point),between a b c (*@$\to$@*) |(a(*@--@*)b)| + |(b(*@--@*)c)| = |(a(*@--@*)c)|
axiom equal_circles : (*@$\forall$@*) (a b c : Point) ((*@$\alpha$@*) (*@$\beta$@*) : Circle),(a.isCentre (*@$\alpha$@*)) (*@$\land$@*) (a.isCentre (*@$\beta$@*)) (*@$\land$@*) (b.onCircle (*@$\alpha$@*)) (*@$\land$@*) (c.onCircle (*@$\beta$@*)) (*@$\land$@*) |(a(*@--@*)b)| = |(a(*@--@*)c)| (*@$\to$@*) (*@$\alpha$@*) = (*@$\beta$@*)
axiom point_on_circle_if : (*@$\forall$@*) (a b c : Point) ((*@$\alpha$@*) : Circle),(a.isCentre (*@$\alpha$@*)) (*@$\land$@*) (b.onCircle (*@$\alpha$@*)) (*@$\land$@*) |(a(*@--@*)c)| = |(a(*@--@*)b)| (*@$\to$@*) c.onCircle (*@$\alpha$@*)
axiom point_on_circle_onlyif : (*@$\forall$@*) (a b c : Point) ((*@$\alpha$@*) : Circle),(a.isCentre (*@$\alpha$@*)) (*@$\land$@*) (b.onCircle (*@$\alpha$@*)) (*@$\land$@*) (c.onCircle (*@$\alpha$@*)) (*@$\to$@*) |(a(*@--@*)c)| = |(a(*@--@*)b)|
axiom point_in_circle_if : (*@$\forall$@*) (a b c : Point) ((*@$\alpha$@*) : Circle),(a.isCentre (*@$\alpha$@*)) (*@$\land$@*) (b.onCircle (*@$\alpha$@*)) (*@$\land$@*) (a(*@--@*)c) < (a(*@--@*)b) (*@$\to$@*) c.insideCircle (*@$\alpha$@*)
axiom sum_angles_if : (*@$\forall$@*) (a b c d : Point) (L M : Line),(a.onLine L) (*@$\land$@*) (a.onLine M) (*@$\land$@*) (b.onLine L) (*@$\land$@*) (c.onLine M) (*@$\land$@*) (a (*@$\neq$@*) b) (*@$\land$@*) (a (*@$\neq$@*) c) (*@$\land$@*)(*@$\neg$@*)(d.onLine L) (*@$\land$@*) (*@$\neg$@*)(d.onLine M) (*@$\land$@*) (L (*@$\neq$@*) M) (*@$\land$@*) ((*@$\angle$@*) b:a:c) = ((*@$\angle$@*) b:a:d) + ((*@$\angle$@*) d:a:c) (*@$\to$@*)(b.sameSide d M) (*@$\land$@*) (c.sameSide d L)
axiom sum_angles_onlyif : (*@$\forall$@*) (a b c d : Point) (L M : Line), (a.onLine L) (*@$\land$@*) (a.onLine M) (*@$\land$@*) (b.onLine L) (*@$\land$@*) (c.onLine M) (*@$\land$@*) (a (*@$\neq$@*) b) (*@$\land$@*) (a (*@$\neq$@*) c) (*@$\land$@*) (d.onLine L) (*@$\land$@*) (*@$\neg$@*)(d.onLine M) (*@$\land$@*) (L (*@$\neq$@*) M) (*@$\land$@*) (b.sameSide d M) (*@$\land$@*) (c.sameSide d L) (*@$\to$@*)  ((*@$\angle$@*) b:a:c) = ((*@$\angle$@*) b:a:d) + ((*@$\angle$@*) d:a:c)
axiom perpendicular_if : (*@$\forall$@*) (a b c d : Point) (L : Line),(a.onLine L) (*@$\land$@*) (b.onLine L) (*@$\land$@*) (between a c b) (*@$\land$@*) (*@$\neg$@*)(d.onLine L) (*@$\land$@*) ((*@$\angle$@*) a:c:d = (*@$\angle$@*) d:c:b) (*@$\to$@*)((*@$\angle$@*) a:c:d) = (*@$\rightangle$@*)
axiom sum_areas_if : (*@$\forall$@*) (a b c d : Point) (L : Line), (a.onLine L) (*@$\land$@*) (b.onLine L) (*@$\land$@*) (c.onLine L) (*@$\land$@*) (a (*@$\neq$@*) b) (*@$\land$@*) (a (*@$\neq$@*) c) (*@$\land$@*) (b (*@$\neq$@*) c) (*@$\land$@*) (*@$\neg$@*)(d.onLine L) (*@$\land$@*) (between a c b) (*@$\to$@*) (Triangle.area (*@$\triangle$@*) a:c:d ) + (Triangle.area (*@$\triangle$@*) d:c:b) = (Triangle.area (*@$\triangle$@*) a:d:b)
axiom rectangle :  (a b c d : Point) (AB CD AC BD : Line),
  formParallelogram a b c d AB CD AC BD (*@$\land$@*) ((*@$\angle$@*) a:c:d = (*@$\rightangle$@*)) (*@$\to$@*)
  (Triangle.area (*@$\triangle$@*) a:c:d + Triangle.area (*@$\triangle$@*) a:b:d = |(a(*@--@*)b)| * |(a(*@--@*)c)|) (*@$\land$@*) (Triangle.area (*@$\triangle$@*)  b:a:c + Triangle.area (*@$\triangle$@*)  b:d:c) = |(a(*@--@*)b)| * |(a(*@--@*)c)|
{PRECEDING_THMS}

--- Proof DSL ---
Your response must be a tactic proof in the LeanEuclid proof DSL. This DSL is built from the following tactics (arguments shown in angle-brackets <> ):

* TACTIC: euclid_intros *
 Introduces universally quantified variables and premises of the current goal into the proof context. No names required.
* TACTIC: euclid_apply <rule> <args> *
 where <rule> is either a construction rule, inference rule, or other theorem.
 Given a rule <rule> with type of the form (*@$\forall$@*) (<args> : Types) ... P -> Q, this tactic instantiates <rule> with <args>, and attempts to prove premise P from the local proof context using an SMT solver. If successful, propsition Q is added to the proof context.
usage examples :
  euclid_apply between_if a b c
  eculid_apply same_side_rfl x y L
* TACTIC: euclid_apply <rule> <args> as X *
Given a rule <rule> with type of the form (*@$\forall$@*) (<args> : Types) ... P -> (*@$\exists$@*) x . Q(x), this tactic instantiates <rule> with <args>, and attempts to prove premise P from the local proof context using an SMT solver. If successful, object x and premise Q(x) are added to the proof context.
usage examples:
  euclid_apply point_same_side a L as b
  euclid_apply line_from_points p1 p2 as M

NOTE: You can only use `euclid_apply  <rule> <args> as <X>` if the rule produces an existential. You should not name any propsotions introduced using `euclid_apply` e,g, `euclid_apply <rule> <args> as H1`.
NOTE: It is very important that *all* non-propositional (i.e., universally quantified) arguments are provided to the rule when invoking `euclid_apply`. 

* euclid_finish * 
    Attempts to resolve the proof goal using the current proof context using an SMT solver.
* euclid_assert <P> *
    Attempts to prove proposition <P> from the current proof context using an SMT solver. Equivalent to "have : <P> := by euclid_finish"

If you are proving an existentially quantified proposition, you can use the standard Lean tactic
` use <X>`
to provide the witness <X> for the quantifier. DO NOT use the tactic `use` if you are not proving an existentially quantified proposition.
 
You can use standard Lean tactics such as <by_cases>, <cases>, <split_ands> and <constructor> to structure your proof. But, you should not use imperative Lean tactics, such as `rw` or `simp`. You should only use the above declarative tactics.

Your response should be exactly of the following form:

    <tac_1>
    <tac_2>
    ...
    <tac_n>

where each <tac_i> is a Lean tactic adhering to the above guidelines. 
 
DO *NOT* wrap your answer in markdown syntax, e.g. ```lean <contents> ```. It must be simply a Lean tactic script that can be inserted into a proof.

Here are some examples. NOTE: You may not necessarily use the propositoins shown in these proofs, unless they are also listed above.

Input: 
 Informal Proof: 
  To place a straight-line equal to a given straight-line at a given point (as an extremity). Let $A$ be the given point, and $BC$ the given straight-line. So  it is required to place a straight-line at point $A$ equal to the given straight-line $BC$.  For  let the straight-line $AB$ have been joined from point $A$ to point $B$ [Post.~1], and let the equilateral triangle $DAB$ have been been constructed upon it [Prop.~1.1].  And let the straight-lines $AE$ and $BF$ have been produced in a straight-line with $DA$ and $DB$  (respectively) [Post.~2]. And let the circle $CGH$ with center $B$ and radius $BC$ have been drawn [Post.~3], and again let the circle $GKL$ with center $D$ and radius $DG$ have been drawn [Post.~3].      Therefore, since the point $B$ is the center of (the circle) $CGH$, $BC$ is equal to  $BG$ [Def.~1.15]. Again, since the point $D$ is the center of the circle $GKL$, $DL$ is equal to $DG$ [Def.~1.15]. And within these,  $DA$ is equal to $DB$. Thus, the remainder $AL$ is equal to the remainder $BG$ [C.N.~3]. But $BC$ was also shown (to be)  equal to $BG$. Thus,  $AL$ and $BC$ are each equal to $BG$. But things equal to the same thing are also equal to one another [C.N.~1]. Thus, $AL$ is also equal to $BC$.  Thus, the straight-line $AL$, equal to the given straight-line $BC$, has been placed at the given point $A$. (Which is) the very thing it was required to do
 
 Proof Setup:
  def example : (*@$\forall$@*) (a b c : Point) (BC : Line), distinctPointsOnLine b c BC (*@$\land$@*) a (*@$\neq$@*) b (*@$\to$@*) (*@$\exists$@*) l : Point, |(a(*@--@*)l)| = |(b(*@--@*)c)| := 
  by 
 
 Your Response: 
  euclid_intros
  euclid_apply (line_from_points a b) as AB
  euclid_apply (proposition_1 a b AB) as d
  euclid_apply (line_from_points d a ) as AE
  euclid_apply (line_from_points d b ) as BF
  euclid_apply (circle_from_points b c) as CGH
  euclid_apply (intersection_circle_line_extending_points CGH BF b d) as g
  euclid_apply (circle_from_points d g) as GKL
  euclid_apply (intersection_circle_line_extending_points GKL AE a d) as l
  euclid_apply (point_on_circle_onlyif b c g CGH)
  euclid_apply (point_on_circle_onlyif d l g GKL)
  euclid_apply (between_if l a d )
  euclid_apply (between_if g b d )
  use l
  euclid_finish

Input: 
 Informal Proof:
  To draw a straight-line perpendicular to a given infinite straight-line from a given point which is not on it.   Let $AB$ be the given infinite straight-line  and $C$ the given point, which is not on ($AB$). So it is required to draw a  straight-line  perpendicular to the given infinite straight-line $AB$ from the given point $C$, which is not on ($AB$).  For let point $D$ have been taken at random on the other side (to $C$) of  the straight-line $AB$, and let the circle $EFG$ have been drawn with center $C$ and radius $CD$ [Post.~3], and let the straight-line $EG$ have been cut in half at (point) $H$ [Prop.~1.10], and let the straight-lines $CG$, $CH$, and $CE$ have been joined. I say that the  (straight-line) $CH$ has been drawn  perpendicular to the given infinite straight-line $AB$ from the given point $C$, which is not on ($AB$).  For since $GH$ is equal to $HE$, and $HC$ (is) common, the two (straight-lines) $GH$,  $HC$ are equal to the two (straight-lines) $EH$, $HC$, respectively, and the base $CG$ is equal to the base $CE$. Thus, the angle $CHG$ is equal to the angle $EHC$ [Prop.~1.8], and they are adjacent. But when a straight-line stood on a(nother) straight-line makes the adjacent angles equal to one another, each of the equal angles is a right-angle, and the former straight-line is called a perpendicular to that upon which it stands [Def.~1.10].  Thus, the (straight-line) $CH$ has been drawn perpendicular to the given infinite straight-line $AB$ from the given point $C$, which is not on  ($AB$). (Which is) the very thing it was required to do.
 
 Proof Setup: 
  def example : (*@$\forall$@*) (a b c : Point) (AB : Line), distinctPointsOnLine a b AB (*@$\land$@*) (*@$\neg$@*)(c.onLine AB) (*@$\to$@*) (*@$\exists$@*) h : Point, h.onLine AB (*@$\land$@*) ((*@$\angle$@*) a:h:c) = (*@$\rightangle$@*) (*@$\lor$@*) ((*@$\angle$@*) b:h:c) = (*@$\rightangle$@*) :=
 
 Your Response: 
  euclid_intros
  euclid_apply (exists_point_opposite AB c) as d
  euclid_apply (circle_from_points c d) as EFG
  euclid_apply (intersections_circle_line EFG AB) as (e, g)
  euclid_apply (proposition_10 e g AB) as h
  euclid_apply (line_from_points c g) as CG
  euclid_apply (line_from_points c h) as CH
  euclid_apply (line_from_points c e) as CE
  use h
  euclid_apply (proposition_8 h c g h c e CH CG AB CH CE AB)
  euclid_finish


Input: 
 Informal Proof:
 In any triangle,  (if) one of the sides (is) produced  (then) the external angle is equal to the (sum of the) two internal and opposite (angles), and the (sum of the) three internal angles of the triangle is equal to two right-angles. Let $ABC$ be a triangle, and let one of its sides $BC$ have been produced to $D$. I say that the external angle $ACD$ is equal to the (sum of the) two internal and opposite angles $CAB$ and $ABC$, and the (sum of the) three internal angles of the triangle---$ABC$, $BCA$, and $CAB$---is equal to two right-angles.  For let $CE$ have been drawn through point $C$ parallel to the straight-line $AB$ [Prop.~1.31].  And since $AB$ is parallel to $CE$, and $AC$ has fallen across them, the alternate angles $BAC$ and $ACE$ are equal to one another [Prop.~1.29]. Again, since $AB$ is parallel to $CE$, and the straight-line $BD$ has fallen across them,  the external angle $ECD$ is equal to the internal and opposite (angle) $ABC$ [Prop.~1.29]. But $ACE$ was also shown (to be) equal to $BAC$. Thus, the whole angle $ACD$ is equal to the (sum of the) two internal and opposite (angles) $BAC$ and $ABC$.  Let $ACB$ have been added to both. Thus, (the sum of) $ACD$ and $ACB$ is equal to the (sum of the) three (angles) $ABC$, $BCA$, and $CAB$. But, (the sum of) $ACD$ and $ACB$ is equal to two right-angles [Prop.~1.13]. Thus, (the sum of) $ACB$, $CBA$, and $CAB$ is also equal to two right-angles.  Thus, in any triangle,  (if) one of the sides  (is) produced (then) the external angle is equal to the (sum of the) two internal and opposite (angles), and the (sum of the) three internal angles of the triangle is equal to two right-angles. (Which is) the very thing it was required to show.
 
 Proof Setup: 
  def example: (*@$\forall$@*) (a b c d : Point) (AB BC AC : Line), formTriangle a b c AB BC AC (*@$\land$@*) (between b c d) (*@$\to$@*) (*@$\angle$@*) a:c:d = (*@$\angle$@*) c:a:b + (*@$\angle$@*) a:b:c (*@$\land$@*) (*@$\angle$@*) a:b:c + (*@$\angle$@*) b:c:a + (*@$\angle$@*) c:a:b = (*@$\rightangle$@*) + (*@$\rightangle$@*) 
  by 

 Your Response: 
  euclid_intros
  have : ((*@$\angle$@*) a:c:d) = ((*@$\angle$@*) c:a:b) + ((*@$\angle$@*) a:b:c) := by
    euclid_apply (proposition_31 c a b AB ) as CE
    euclid_apply (point_on_line_same_side BC CE a ) as e
    euclid_apply (proposition_29''' b e a c AB CE AC)
    euclid_apply (proposition_29'''' e a d c b CE AB BC)
    euclid_finish
  constructor
  . assumption
  . euclid_apply (proposition_13 a c b d AC BC)
    euclid_finish



Input: 
 Informal Proof:
  To construct a parallelogram equal to a given triangle in a given rectilinear angle. Let $ABC$ be the given triangle, and $D$ the given rectilinear angle. So it is required to construct a parallelogram equal to triangle $ABC$ in the rectilinear angle $D$. Let $BC$ have been cut in half at $E$ [Prop.~1.10], and let $AE$ have been joined. And let (angle) $CEF$, equal to angle $D$,  have been constructed at the point $E$ on the straight-line $EC$ [Prop.~1.23]. And let $AG$ have been drawn through $A$ parallel to $EC$ [Prop.~1.31], and let $CG$ have been drawn through $C$ parallel to $EF$ [Prop.~1.31]. Thus, $FECG$ is a parallelogram. And since $BE$ is equal to $EC$, triangle $ABE$ is also equal to triangle $AEC$. For they are on the equal bases, $BE$ and $EC$, and between the same parallels, $BC$ and $AG$ [Prop.~1.38]. Thus, triangle $ABC$ is double (the area) of triangle $AEC$. And parallelogram $FECG$ is also double (the area) of triangle $AEC$. For it has the same base as ($AEC$), and is between the same parallels  as ($AEC$) [Prop.~1.41]. Thus, parallelogram $FECG$ is equal to triangle $ABC$.  ($FECG$) also has the angle $CEF$ equal to the given (angle) $D$. Thus, parallelogram $FECG$,  equal to the given triangle $ABC$, has been constructed in the angle $CEF$, which is equal to $D$. (Which is) the very thing it was required to do.
 
 Theorem Statement: 
  def example : (*@$\forall$@*) (a b c d1 d2 d3 : Point) (AB BC AC D12 D23: Line), formTriangle a b c AB BC AC (*@$\land$@*) formRectilinearAngle d1 d2 d3 D12 D23 (*@$\land$@*) ((*@$\angle$@*) d1:d2:d3) > 0 (*@$\land$@*) ((*@$\angle$@*) d1:d2:d3) < (*@$\rightangle$@*) + (*@$\rightangle$@*) (*@$\to$@*) (*@$\exists$@*) (f g e c' : Point) (FG EC EF CG : Line), formParallelogram f g e c' FG EC EF CG (*@$\land$@*) ((*@$\angle$@*) c':e:f = (*@$\angle$@*) d1:d2:d3) (*@$\land$@*) (Triangle.area (*@$\triangle$@*) f:e:c' + Triangle.area (*@$\triangle$@*) f:c':g = Triangle.area (*@$\triangle$@*) a:b:c) := 
  by 
 
 Your Response: 
  euclid_intros
  euclid_apply (proposition_10 b c BC) as e
  euclid_apply (line_from_points a e) as AE
  euclid_apply (proposition_23''' e c d2 d1 d3 a BC D12 D23) as f'
  euclid_apply (line_from_points e f') as EF
  euclid_apply (proposition_31 a b c BC) as AG
  euclid_apply (intersection_lines AG EF) as f
  euclid_apply (proposition_31 c e f EF) as CG
  euclid_apply (intersection_lines CG AG) as g
  euclid_assert (formParallelogram f g e c AG BC EF CG)
  euclid_apply (proposition_38 a b e a e c AG BC AB AE AE AC)
  euclid_apply (proposition_41 f e c g a AG BC EF CG AE AC)
  use f, g, e, c, AG, BC, EF, CG
  euclid_finish

**NOTICE: in the above example, there is exactly one argument to `use` for every existentially quantified variables in the theoerem statement.**

Input : 
  Informal Proof: 
    If a triangle has two angles equal to one another then the sides subtending the equal angles will also be equal to one another. Let $ABC$ be a triangle having the angle $ABC$ equal to the angle $ACB$. I say that side $AB$ is also equal to side $AC$.  For if $AB$ is unequal to $AC$ then one of them is greater. Let $AB$ be greater. And let $DB$, equal to the lesser $AC$, have been cut off from the greater $AB$ [Prop.~1.3].  And let $DC$ have been joined [Post.~1].  Therefore, since $DB$ is equal to $AC$, and $BC$ (is) common, the two sides $DB$, $BC$ are equal to the two sides $AC$, $CB$, respectively, and the angle $DBC$ is equal to the angle $ACB$. Thus, the base $DC$ is equal to the base $AB$, and the triangle $DBC$ will be equal to the triangle $ACB$ [Prop.~1.4], the lesser to the greater. The very notion (is) absurd [C.N.~5]. Thus, $AB$ is not unequal to $AC$. Thus, (it is) equal.  Thus, if a triangle has two angles equal to one another then the sides subtending the equal angles will also be equal to one another. (Which is) the very thing it was required to show.
  Proof Setup:
    theorem proposition_6 : (*@$\forall$@*) (a b c : Point) (AB BC AC : Line), formTriangle a b c AB BC AC (*@$\land$@*) ((*@$\angle$@*) a:b:c = (*@$\angle$@*) a:c:b) (*@$\to$@*)(|(a(*@--@*)b)| = |(a(*@--@*)c)|) :=
  Your Response: 
    euclid_intros
    by_contra
    by_cases |(a(*@--@*)b)| > |(a(*@--@*)c)|
    . euclid_apply (proposition_3 b a a c AB AC) as d
      euclid_apply (line_from_points d c) as DC
      euclid_apply proposition_4 b d c c a b AB DC BC AC AB BC
      euclid_finish
    . euclid_apply (proposition_3 c a a b AC AB) as d
      euclid_apply (line_from_points d b) as DB
      euclid_apply (proposition_4 c d b b a c AC DB BC AB AC BC)
      euclid_finish
--- End of Instruction ---
\end{lstlisting}

\subsection{Proof Formalization (UniGeo)}
To formalize UniGeo proofs, we likewise include those inference rules which were used in our manual formalization, as well as the relevant theorems from \emph{Elements} . Unlike for proofs from \emph{Elements}, we randomly select shots from a separate test set of UniGeo proofs. The guidelines are slightly different from \emph{Elements}, to accommodate the different style of input for UniGeo proofs. 

\begin{lstlisting}[language=lean, numbers=none]
    --- Definitions ---
- Basic Geometric Sorts -
axiom Point : Type
axiom Line : Type
axiom Circle : Type

- Inductive Types for Geometric Entities -
inductive Angle | right | ofPoints (A B C : Point)
inductive Segment | endpoints (a b : Point)
inductive Triangle | ofPoints (a b c : Point)

- Notations and Macros for Geometric Entities -
"|(a(*@--@*)b)|" means the length of the line segment between point a and point b.
"(*@$\angle$@*) a:b:c" means the degree of the angle formed by points a, b, and c.
"(*@$\rightangle$@*)" means the right angle.
"(*@$\triangle$@*) a:b:c" means the triangle formed from points a, b and c.
"Triangle.area (*@$\triangle$@*) a:b:c" means the area of the triangle formed by points a, b and c.

- Relations and Axioms for Geometric Sorts -
namespace Point
def onLine (a : Point) (L : Line) -- point a is on line L.
def sameSide (a b : Point) (L : Line) -- point a and b are on the same side of line L.
def opposingSides (a b : Point) (L : Line) -- point a and b are on the opposite sides of line L.
def collinear : (a b c : Point) -- points a, b and c are collinear.
end Point

namespace Line
def intersectsLine (L M : Line) -- two lines L and M intersect at some point.
def intersectsCircle (L : Line) (C : Circle) -- line L and circle C intersect.
end Line

namespace Circle
def onCircle (a: Point) (C: Circle) -- point a is on circle C.
def insideCircle (a: Point) (C: Circle) -- point a is inside circle C.
def outsideCircle (a: Point) (C: Circle)-- point a is outside circle C.
def isCentre (a: Point) (C: Circle) -- point a is on the center circle C.
def intersectsCircle (C1 C2: Circle) -- circle C1 and C2 intersect.
end Circle

namespace Triangle
-- triangle T1 and T2 are congruent (i.e satisfies the SAS, AAS, ASA or SSS condition)
def congruent (T1 T2 : Triangle) 

-- triangle T1 and T2 are similar.
def similar (T1 T2 : Triangle) 

-- If T1 and T2 are congruent, then the corresponding angles and sides are equal.
theorem congruent_if (T1 T2: Triangle): congruent T1 T2 (*@$\to$@*) match T1,T2 with
  | (Triangle.ofPoints A B C) ,(Triangle.ofPoints D E F) =>
    |(A(*@--@*)B)| = |(D(*@--@*)E)| (*@$\land$@*) |(B(*@--@*)C)| = |(E(*@--@*)F)| (*@$\land$@*) |(A(*@--@*)C)| = |(D(*@--@*)F)| (*@$\land$@*) (*@$\angle$@*) A:B:C = (*@$\angle$@*) D:E:F (*@$\land$@*) (*@$\angle$@*) A:C:B = (*@$\angle$@*) D:F:E (*@$\land$@*) (*@$\angle$@*) B:A:C = (*@$\angle$@*) E:D:F  

-- If T1 and T2 are similar, then the corresponding angles and the ratio of corresponding sides are equal.
theorem similar_if (T1 T2: Triangle): similar T1 T2 (*@$\to$@*) match T1,T2 with
  | (Triangle.ofPoints A B C) ,(Triangle.ofPoints D E F) =>
    |(A(*@--@*)B)| / |(D(*@--@*)E)| = |(B(*@--@*)C)| / |(E(*@--@*)F)| (*@$\land$@*) |(A(*@--@*)B)| / |(D(*@--@*)E)| = |(B(*@--@*)C)| / |(E(*@--@*)F)|
    (*@$\land$@*) |(C(*@--@*)A)| / |(F(*@--@*)D)| = |(A(*@--@*)B)| / |(D(*@--@*)E)| (*@$\land$@*) (*@$\angle$@*) A:B:C = (*@$\angle$@*) D:E:F
    (*@$\land$@*) (*@$\angle$@*) A:C:B = (*@$\angle$@*) D:F:E (*@$\land$@*) (*@$\angle$@*) B:A:C = (*@$\angle$@*) E:D:F 

- Geometric Relations -
-- points a and b are distinct and on line L.
def distinctPointsOnLine (a b : Point) (L : Line) 

-- line AB and BC intersect at point b.
def twoLinesIntersectAtPoint (AB BC : Line) (b : Point) 

-- points a, b and c collinear and cyclically ordered.
def between (a b c : Point) 

-- point a, b and c form a triangle, where point a and b are on line AB, point b and c are on line BC, point a and c are on line CA.
def formTriangle (a b c : Point) (AB BC CA : Line) 

-- point a, b and c form a rectilinear angle where a(*@$\neq$@*)b and b(*@$\neq$@*)c.
def formRectilinearAngle (a b c : Point) (AB BC : Line)  

-- point a, b, d, c (in clockwise/counterclockwise order) form a parallelogram, a and b are on line AB, c and d are on line CD, a and c are on line AC, b and d are on line BD
def formParallelogram (a b c d : Point) (AB CD AC BD : Line)

-- point a, b, d, c (in clockwise/counterclockwise order) form a quadrilateral, a and b are on line AB, c and d are on line CD, a and c are on line AC, b and d are on line BD
def formQuadrilateral (a b c d : Point) (AB CD AC BD : Line) 

--- Axioms ---
- Construction Rules-
axiom extend_point : (*@$\forall$@*) (L : Line) (b c : Point), distinctPointsOnLine b c L (*@$\to$@*) (*@$\exists$@*) a : Point, (a.onLine L) (*@$\land$@*) (between b c a)

- Useful Lemmas from Euclid's Elements Book I-
-- In isosceles triangles the angles at the base equal one another, and, if the equal straight lines are produced further, then the angles under the base equal one another.
theorem proposition_5 : (*@$\forall$@*) (a b c d e : Point) (AB BC AC : Line), formTriangle a b c AB BC AC (*@$\land$@*) (|(a(*@--@*)b)| = |(a(*@--@*)c)|) (*@$\land$@*) (between a b d) (*@$\land$@*) (between a c e) (*@$\to$@*) ((*@$\angle$@*) a:b:c) = ((*@$\angle$@*) a:c:b) (*@$\land$@*) ((*@$\angle$@*) c:b:d) = ((*@$\angle$@*) b:c:e)
theorem proposition_5' : (*@$\forall$@*) (a b c : Point) (AB BC AC : Line), formTriangle a b c AB BC AC (*@$\land$@*) (|(a(*@--@*)b)| = |(a(*@--@*)c)|) (*@$\to$@*) (((*@$\angle$@*) a:b:c ) = (*@$\angle$@*) a:c:b)

-- If a straight line stands on a straight line, then it makes either two right angles or angles whose sum equals two right angles.
theorem proposition_13 : (*@$\forall$@*) (a b c d : Point) (AB CD : Line), AB (*@$\neq$@*) CD (*@$\land$@*) distinctPointsOnLine a b AB (*@$\land$@*) distinctPointsOnLine c d CD (*@$\land$@*) between d b c (*@$\to$@*) (*@$\angle$@*) c:b:a + (*@$\angle$@*) a:b:d = (*@$\rightangle$@*) + (*@$\rightangle$@*)

-- If two straight lines cut one another, then they make the vertical angles equal to one another.
theorem proposition_15 : (*@$\forall$@*) (a b c d e : Point) (AB CD : Line), distinctPointsOnLine a b AB (*@$\land$@*) distinctPointsOnLine c d CD (*@$\land$@*) e.onLine AB (*@$\land$@*) e.onLine CD (*@$\land$@*) CD (*@$\neq$@*) AB (*@$\land$@*) (between d e c) (*@$\land$@*) (between a e b) (*@$\to$@*) ((*@$\angle$@*) a:e:c ) = ((*@$\angle$@*) d:e:b) (*@$\land$@*) ((*@$\angle$@*) c:e:b ) = ((*@$\angle$@*) a:e:d)

-- If a straight line falling on two straight lines makes the alternate angles equal to one another, then the straight lines are parallel to one another.
theorem proposition_27 : (*@$\forall$@*) (a b c d e f : Point) (AB CD EF : Line), distinctPointsOnLine a b AB (*@$\land$@*) distinctPointsOnLine c d CD (*@$\land$@*) distinctPointsOnLine e f EF (*@$\land$@*) (between a e b) (*@$\land$@*) (between c f d) (*@$\land$@*) (b.sameSide d EF) (*@$\land$@*) (*@$\angle$@*) a:e:f  = ((*@$\angle$@*) e:f:d) (*@$\to$@*) (*@$\neg$@*)(AB.intersectsLine CD)
theorem proposition_27' : (*@$\forall$@*) (a d e f : Point) (AB CD EF : Line), distinctPointsOnLine a e AB (*@$\land$@*) distinctPointsOnLine f d CD (*@$\land$@*) distinctPointsOnLine e f EF (*@$\land$@*) a.opposingSides d EF (*@$\land$@*) (*@$\angle$@*) a:e:f = ((*@$\angle$@*) e:f:d) (*@$\to$@*) (*@$\neg$@*)(AB.intersectsLine CD)

-- If a straight line falling on two straight lines makes the exterior angle equal to the interior and opposite angle on the same side, or the sum of the interior angles on the same side equal to two right angles, then the straight lines are parallel to one another.
theorem proposition_28 : (*@$\forall$@*) (a b c d e f g h : Point) (AB CD EF : Line), distinctPointsOnLine a b AB (*@$\land$@*) distinctPointsOnLine c d CD (*@$\land$@*) distinctPointsOnLine e f EF (*@$\land$@*) (between a g b) (*@$\land$@*) (between c h d) (*@$\land$@*) (between e g h) (*@$\land$@*) (between g h f) (*@$\land$@*) (b.sameSide d EF) (*@$\land$@*) ((*@$\angle$@*) e:g:b = (*@$\angle$@*) g:h:d (*@$\lor$@*) (*@$\angle$@*) b:g:h + (*@$\angle$@*) g:h:d = (*@$\rightangle$@*) + (*@$\rightangle$@*)) (*@$\to$@*) (*@$\neg$@*)(AB.intersectsLine CD)

-- A straight line falling on parallel straight lines makes the alternate angles equal to one another, the exterior angle equal to the interior and opposite angle, and the sum of the interior angles on the same side equal to two right angles.
theorem proposition_29 : (*@$\forall$@*) (a b c d e f g h : Point) (AB CD EF : Line), distinctPointsOnLine a b AB (*@$\land$@*) distinctPointsOnLine c d CD (*@$\land$@*) distinctPointsOnLine e f EF (*@$\land$@*) (between a g b) (*@$\land$@*) (between c h d) (*@$\land$@*) (between e g h) (*@$\land$@*) (between g h f) (*@$\land$@*) (b.sameSide d EF) (*@$\land$@*) (*@$\neg$@*)(AB.intersectsLine CD) (*@$\to$@*) (*@$\angle$@*) a:g:h = (*@$\angle$@*) g:h:d (*@$\land$@*) (*@$\angle$@*) e:g:b = (*@$\angle$@*) g:h:d (*@$\land$@*) (*@$\angle$@*) b:g:h + (*@$\angle$@*) g:h:d = (*@$\rightangle$@*) + (*@$\rightangle$@*)
theorem proposition_29' : (*@$\forall$@*) (a b c d e g h : Point) (AB CD EF : Line), distinctPointsOnLine a b AB (*@$\land$@*) distinctPointsOnLine c d CD (*@$\land$@*) distinctPointsOnLine g h EF (*@$\land$@*) (between a g b) (*@$\land$@*) (between c h d) (*@$\land$@*) (between e g h) (*@$\land$@*) (b.sameSide d EF) (*@$\land$@*) (*@$\neg$@*)(AB.intersectsLine CD) (*@$\to$@*) (*@$\angle$@*) a:g:h = (*@$\angle$@*) g:h:d (*@$\land$@*) (*@$\angle$@*) e:g:b = (*@$\angle$@*) g:h:d (*@$\land$@*) (*@$\angle$@*) b:g:h + (*@$\angle$@*) g:h:d = (*@$\rightangle$@*) + (*@$\rightangle$@*)
theorem proposition_29'' : (*@$\forall$@*) (a b d g h : Point) (AB CD GH : Line), distinctPointsOnLine a b AB (*@$\land$@*) distinctPointsOnLine h d CD (*@$\land$@*) distinctPointsOnLine g h GH (*@$\land$@*) (between a g b) (*@$\land$@*) (b.sameSide d GH) (*@$\land$@*) (*@$\neg$@*)(AB.intersectsLine CD) (*@$\to$@*) (*@$\angle$@*) a:g:h = (*@$\angle$@*) g:h:d (*@$\land$@*) (*@$\angle$@*) b:g:h + (*@$\angle$@*) g:h:d = (*@$\rightangle$@*) + (*@$\rightangle$@*)
theorem proposition_29''' : (*@$\forall$@*) (a d g h : Point) (AB CD GH : Line),distinctPointsOnLine a g AB (*@$\land$@*) distinctPointsOnLine h d CD (*@$\land$@*) distinctPointsOnLine g h GH (*@$\land$@*)a.opposingSides d GH (*@$\land$@*) (*@$\neg$@*)(AB.intersectsLine CD) (*@$\to$@*) (*@$\angle$@*) a:g:h = (*@$\angle$@*) g:h:d
theorem proposition_29'''' : (*@$\forall$@*) (b d e g h : Point) (AB CD EF : Line), distinctPointsOnLine g b AB (*@$\land$@*) distinctPointsOnLine h d CD (*@$\land$@*) distinctPointsOnLine e h EF (*@$\land$@*) between e g h  (*@$\land$@*) b.sameSide d EF (*@$\land$@*) (*@$\neg$@*)(AB.intersectsLine CD) (*@$\to$@*) (*@$\angle$@*) e:g:b = (*@$\angle$@*) g:h:d
theorem proposition_29''''' : (*@$\forall$@*) (b d g h : Point) (AB CD EF : Line), distinctPointsOnLine g b AB (*@$\land$@*) distinctPointsOnLine h d CD (*@$\land$@*) distinctPointsOnLine g h EF (*@$\land$@*) b.sameSide d EF (*@$\land$@*) (*@$\neg$@*)(AB.intersectsLine CD) (*@$\to$@*) (*@$\angle$@*) b:g:h + (*@$\angle$@*) g:h:d = (*@$\rightangle$@*) + (*@$\rightangle$@*)

-- In any triangle, if one of the sides is produced, then the exterior angle equals the sum of the two interior and opposite angles, and the sum of the three interior angles of the triangle equals two right angles.
theorem proposition_32 : (*@$\forall$@*) (a b c d : Point) (AB BC AC : Line), formTriangle a b c AB BC AC (*@$\land$@*) (between b c d) (*@$\to$@*) (*@$\angle$@*) a:c:d = (*@$\angle$@*) c:a:b + (*@$\angle$@*) a:b:c (*@$\land$@*) (*@$\angle$@*) a:b:c + (*@$\angle$@*) b:c:a + (*@$\angle$@*) c:a:b = (*@$\rightangle$@*) + (*@$\rightangle$@*)
-- In parallelogrammic areas the opposite sides and angles equal one another, and the diameter bisects the areas.
theorem proposition_34 : (*@$\forall$@*) (a b c d : Point) (AB CD AC BD BC : Line), formParallelogram a b c d AB CD AC BD (*@$\land$@*) distinctPointsOnLine b c BC (*@$\to$@*) |(a(*@--@*)b)| = |(c(*@--@*)d)| (*@$\land$@*) |(a(*@--@*)c)| = |(b(*@--@*)d)| (*@$\land$@*) (*@$\angle$@*) a:b:d = (*@$\angle$@*) a:c:d (*@$\land$@*) (*@$\angle$@*) b:a:c  = (*@$\angle$@*) c:d:b (*@$\land$@*) Triangle.area (*@$\triangle$@*) a:b:c = Triangle.area (*@$\triangle$@*) d:c:b
theorem proposition_34' : (*@$\forall$@*) (a b c d : Point) (AB CD AC BD : Line), formParallelogram a b c d AB CD AC BD (*@$\to$@*) |(a(*@--@*)b)| = |(c(*@--@*)d)| (*@$\land$@*) |(a(*@--@*)c)| = |(b(*@--@*)d)| (*@$\land$@*) (*@$\angle$@*) a:b:d = (*@$\angle$@*) a:c:d (*@$\land$@*) (*@$\angle$@*) b:a:c = (*@$\angle$@*) c:d:b

--- Proof DSL ---
Your response must be a tactic proof in the LeanEuclid proof DSL. This DSL is built from the following tactics:
You need to formalize each key step (lemma) in the given proof and prove it by 'have <name> : <claim> := by <proof>'

1. euclid_intros
It introduces universally quantified variables and premises of the current goal into the proof context. No names required.

2. euclid_apply <rule> <args>
If <rule> has the form (*@$\forall$@*) (<args> : Types) ... P -> Q, it instantiates <rule> with <args>, and attempts to prove premise P from the local proof context using the above axioms. If successful, propsition Q is added to the proof context.
usage examples :
  euclid_apply proposition_13 A B C D AB CD -- This proves that (*@$\angle$@*) C:B:A + (*@$\angle$@*) A:B:D = (*@$\rightangle$@*) + (*@$\rightangle$@*)

If <rule> has the form (*@$\forall$@*) (<args> : Types) ... P -> (*@$\exists$@*) x . Q(x), this tactic instantiates <rule> with <args>, and attempts to prove premise P from the local proof context using the above axioms. If successful, object x and premise Q(x) are added to the proof context.
usage examples:
  euclid_apply extend_point L a b as c -- This creates a point c on line L such that b is between a and c.

3. euclid_assert <P>
It attempts to prove proposition <P> from the current proof context and the above axioms.

4. euclid_finish
It attempts to resolve the proof goal using the current proof context and the above axioms.

5. use <X>
This tactic proves an existentially quantified proposition by providing the witness <X> for the quantifier.

6. have <name> : <claim> := by <proof>
This tactic proves a lemma <claim>. The <name> is optional, and you need to use euclid_finish at the end of <proof>. If <claim> is already in premise, you can use directly use.
In case the <proof> only contains euclid_finish, you can directly use euclid_assert <claim> for simplicity.

--- Guidelines ---
1. Response Format: Present your proof within THREE angle brackets by <<<Your Proof>>>. Do not add any annotations/explanations in your proof. Do not use markdown syntax.
2. You should not use other imperative Lean tactics, such as 'rw' or 'simp'. You should only use the above declarative tactics.
--- End of Instruction ---
\end{lstlisting}

\section{Theorem Statement Autoformalization}

\label{appendix:props}
\subsection{Successful Formalizations}
As mentioned in the paper, some of the successful predictions proven correct by {\toolName} were actually syntactic matches with their ground truth formalizations; these cases are not interesting, since existing syntactic evaluation techniques can identify these are correct. What is interesting are cases where the prediction is \emph{not} a syntactic match with its ground truth. Even a small deviation (e.g., writing $\angle ABC$ rather than $\angle BAC$) may or may not affect the semantics of the result, depending on the context. It is these cases for which {\toolName} offers an improvement over  syntactic or manual evaluation methods. We highlight some such cases here.

\smallsec{Proposition 8, GPT-4V, 5-shot}
Consider Proposition 8 of \emph{Elements}, whose abridged version used for autoformalization of theorem statements reads as follows:
\begin{center}
    If two triangles have two sides equal to two sides, respectively,  and also have the base equal to the base, then they will also have equal the angles  encompassed by the equal straight-lines. Let $ABC$ and $DEF$ be two triangles having the two sides $AB$ and $AC$ equal to the two sides $DE$ and $DF$, respectively. (That is) $AB$ to $DE$, and $AC$ to $DF$. Let them also have the base $BC$ equal to the base $EF$. I say that the angle $BAC$ is also equal to the angle $EDF$ ...   So the angle $BAC$ will also coincide with angle $EDF$, and will be equal to it [C.N.~4]. Thus, if two triangles have  two  sides equal to two side, respectively, and  have the base equal to the base, then they will also have equal the angles  encompassed by the equal straight-lines
\end{center}

The formalization generated by GPT-4 from this text is as follows, alongside our ground truth formalization:

\begin{lstlisting}[language=lean, numbers=none]
prediction :  
    (*@$\forall$@*) (a b c d e f : Point) (AB AC BC DE DF EF : Line),
    formTriangle a b c AB AC BC (*@$\land$@*) formTriangle d e f DE DF EF (*@$\land$@*) 
    |(a(*@--@*)b)| = |(d(*@--@*)e)| (*@$\land$@*) |(a(*@--@*)c)| = |(d(*@--@*)f)| (*@$\land$@*)  |(b(*@--@*)c)| = |(e(*@--@*)f)| (*@$\to$@*) 
    ((*@$\angle$@*) a:b:c = (*@$\angle$@*) d:e:f) 

ground : 
    (*@$\forall$@*) (a b c d e f : Point) (AB BC AC DE EF DF : Line), 
    formTriangle a b c AB BC AC (*@$\land$@*) formTriangle d e f DE EF DF (*@$\land$@*) 
    |(a(*@--@*)b)| = |(d(*@--@*)e)| (*@$\land$@*) |(a(*@--@*)c)| = |(d(*@--@*)f)| (*@$\land$@*) |(b(*@--@*)c)| = |(e(*@--@*)f)| (*@$\to$@*) 
    ((*@$\angle$@*) b:a:c) = ((*@$\angle$@*) e:d:f)    
\end{lstlisting}

Syntactically, there are four differences between the two propositions: in each of the ${\tt formTriangle}$ clauses, we have swapped two sets of lines $BC \mapsto AC$ and $EF \mapsto DF$, and in the postcondition, we have also swapped points $b \mapsto a$ and $e \mapsto d$. This is a good example of a set of simple perturbations whose validity may not be immediately obvious to a human grader. However, {\toolName} is easily able to prove that these propositions are indeed equivalent.

\smallsec{Proposition 10, GPT-4, 1-shot}
Consider Proposition 10 of \emph{Elements}:
\begin{center}
    ``To cut a given finite straight-line in half. Let $AB$ be the given finite straight-line. So it is required to cut the finite straight-line $AB$ in half... Thus, the base $AD$ is equal to the base $BD$ [Prop.~1.4]. Thus, the given finite straight-line $AB$ has been cut in half at (point) $D$.''
\end{center}

When formalizing this statement, one has to specify that point $D$ lies halfway between $A$ and $B$. Our ground truth formalization formalizes this by asserting that $D$ lies between $A$ and $B$, and the length $|AD|$ is equal to the length $|BD|$.

\begin{lstlisting}[language=lean,numbers=none]
ground : 
    (*@$\forall$@*) (a b : Point) (AB : Line), 
    distinctPointsOnLine a b AB (*@$\to$@*) 
    (*@$\exists$@*) d : Point, between a d b (*@$\land$@*) |(a(*@--@*)d)| = |(d(*@--@*)b)|
\end{lstlisting}

However, there is another way to specify that $D$ lies exactly halfway between $A$ and $B$, as suggested by GPT-4 -- namely, by specifying that, in addition to $|AD| = |BD|$, we have $|AD| + |BD| = |AB|$:
\begin{lstlisting}[language=lean, numbers=none]
prediction:
    (*@$\forall$@*) (a b : Point) (AB : Line), 
    distinctPointsOnLine a b AB (*@$\to$@*) 
    (*@$\exists$@*) d : Point, |(a(*@--@*)d)| = |(b(*@--@*)d)| (*@$\land$@*) |(d(*@--@*)b)| + |(a(*@--@*)d)| = |(a(*@--@*)b|
\end{lstlisting}
{\toolName} is able to fill the gap between the two formulations.

\smallsec{Proposition 20, GPT-4, 1-shot}
Consider Proposition 20 of \emph{Elements}:

\begin{center}
    In any triangle, (the sum of) two sides taken together in any (possible way) is greater than the remaining (side). For let $ABC$ be a triangle. I say that in triangle $ABC$ (the sum of) two sides taken together in any (possible way) is greater than the remaining (side)... (So), (the sum of) $BA$ and $AC$ (is greater) than $BC$, (the sum of) $AB$ and $BC$ than $AC$, and (the sum of) $BC$ and $CA$ than $AB$.  Thus, (the sum of) $BA$ and $AC$ is greater than $BC$. Similarly, we can show that (the sum of) $AB$ and $BC$ is also greater than $CA$, and (the sum of) $BC$ and $CA$ than $AB$. Thus, in any triangle, (the sum of) two sides taken together in any (possible way) is greater than the remaining (side)
\end{center}

In the text, Euclid explicitly mentions each combination of distinct line segments explicitly. However, when formalizing the theorem statement, it is not necessary to do so. All combinations are covered by the following concise formulation: 

\begin{lstlisting}[language=lean, numbers=none]
ground : 
    (*@$\forall$@*) (a b c : Point) (AB BC AC : Line),
    formTriangle a b c AB BC AC (*@$\to$@*) 
    |(b(*@--@*)a)| + |(a(*@--@*)c)| > |(b(*@--@*)c)|)
\end{lstlisting}

Since this simplification is not obvious to the model, it will instead attempt to enumerate each for the combinations mentioned:

\begin{lstlisting}[language=lean, numbers=none]
prediction : 
    (*@$\forall$@*) (a b c : Point) (AB BC CA : Line), 
    formTriangle a b c AB BC CA (*@$\to$@*) 
    (|(a(*@--@*)b)| + |(b(*@--@*)c)| > |(a(*@--@*)c)|) (*@$\land$@*) 
    (|(a(*@--@*)b)| + |(a(*@--@*)c)| > |(b(*@--@*)c)|) (*@$\land$@*) 
    (|(b(*@--@*)c)| + |(a(*@--@*)c)| > |(a(*@--@*)b)|)
\end{lstlisting}
Using only a syntactic similarity metric, one might judge the prediction poorly since it nearly doubles the character length of the ground truth formalization. However, {\toolName} correctly identifies them as semantically equivalent.

\subsection{False Negatives}
\label{appendix:FalseNeg}
As mentioned in the paper, {\toolName} is occasionally unable to verify reasonable predictions. Based on our sample size, we estimate the false negative rate to be approximately 15\%. Here, we share the false negatives we found while manually evaluating the results from GPT-4 and GPT-4V ($k=5$) in Experiment \#1.

\smallsec{Proposition 1, Both Models, 5-shot}
Two of our false negatives were the same formalization of Proposition 1, whose text reads:
\begin{center}
    ``To construct an equilateral triangle on a given finite straight-line. Let $AB$ be the given finite straight-line... Thus, the three (straight-lines) $CA$, $AB$, and $BC$ are equal to one another. Thus, the triangle $ABC$ is equilateral, and has been constructed on the given finite straight-line $AB$''.
\end{center}

Our ground truth formalization is as follows:
\begin{lstlisting}[language=lean, numbers=none]
ground:     
    (*@$\forall$@*) (a b : Point) (AB : Line), distinctPointsOnLine a b AB (*@$\to$@*) 
    (*@$\exists$@*) c : Point, |(c(*@--@*)a)| = |(a(*@--@*)b)| (*@$\land$@*) |(c(*@--@*)b)| = |(a(*@--@*)b)|
\end{lstlisting}

The prediction which {\toolName} is unable to prove equivalent is
\begin{lstlisting}[language=lean]
prediction:
    (*@$\forall$@*)  (a b : Point) (AB : Line), distinctPointsOnLine a b AB (*@$\to$@*)  
    (*@$\exists$@*)  (c : Point) (AC BC : Line), formTriangle a b c AB BC AC (*@$\land$@*)  
        |(a(*@--@*) b)| = |(b(*@--@*) c)| (*@$\land$@*) |(a(*@--@*) c)| = |(a(*@--@*) b)|
\end{lstlisting}
During equivalence checking, {\toolName} is unable to apply the construction rules required to form lines $AC$ and $BC$.

\smallsec{Proposition 11, Text-based, 5-shot}
Another false negative came from Proposition 11, which reads:
\begin{center}
    ``To draw a straight-line at right-angles to a given straight-line from a given point on it. Let $AB$ be the given straight-line, and $C$ the given point on it. So it is required to draw a straight-line from the point $C$ at right-angles to the straight-line $AB$.  Thus, the straight-line $CF$ has been drawn at right-angles to the given straight-line $AB$ from the given point $C$ on it.''
\end{center}

Our ground truth formalization is as follows:

\begin{lstlisting}[language=lean, numbers=none]
ground : 
    (*@$\forall$@*)  (a b c : Point) (AB : Line), distinctPointsOnLine a b AB (*@$\land$@*)  between a c b (*@$\to$@*)  
    (*@$\exists$@*)  f : Point, (*@$\neg$@*)(f.onLine AB) (*@$\land$@*)  (*@$\angle$@*) a:c:f = (*@$\rightangle$@*) 
\end{lstlisting}
This formalization reflects the diagram in the souce material, in which the point $C$ around which the right angle is constructed indeed lies between $A$ and $B$. However, the only real requirement in the formalization is to ensure that there is a right angle against line $AB$. 

The prediction which is rejected by {\toolName} is as follows:
\begin{lstlisting}[language=lean, numbers=none]
prediction:
    (*@$\forall$@*) (a b c : Point) (AB : Line), distinctPointsOnLine a b AB (*@$\land$@*) c.onLine AB 
    (*@$\to$@*) (*@$\exists$@*) (f : Point), ((*@$\angle$@*) a:c:f) = (*@$\rightangle$@*) (*@$\land$@*) ((*@$\angle$@*) b:c:f) = (*@$\rightangle$@*)
\end{lstlisting}
Which captures the essence of the theorem. Both theorems are actually true, and are each reasonable formalizations of the given statement, but {\toolName} is unable to prove their equivalence.  In addition to revealing the limitations of the tool, this also reflects the difficulty in building a dataset which can anticipate all possible ways of formalizing a natural language theorem.

\subsection{False Positive}
As mentioned in the paper, we did not encounter indications of soundness bugs in our experiment results. However, we did encounter one instance of a proposition which was proven correct by {\toolName}, despite being an ``incorrect'' formalization of the given proposition. The instance is from the UniGeo/Congruent dataset, which amounts to proving that, with respect to the diagram in Fig.~\ref{fig:UniGeo2}, if $|TU| = |VW|$ and $TU$ is parallel to $VW$, then the triangles formed by points $TUV$ and $TVW$ are congruent.

\begin{figure}[t]
    \centering
    \includegraphics[scale=0.5]{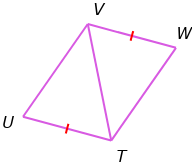}
    \caption{Diagrammatic input for UniGeo proof/Congruent Thm02}
    \label{fig:UniGeo2}
\end{figure}
Our ground truth formalization is as follows:
\begin{lstlisting}[language=lean, numbers=none]
ground : 
    (*@$\forall$@*) (T U V W : Point) (TU UV TV VW TW : Line), 
    formTriangle T U V TU UV TV (*@$\land$@*) formTriangle T V W TV VW TW (*@$\land$@*) 
    U.opposingSides W TV (*@$\land$@*) |(T(*@--@*)U)| = |(V(*@--@*)W)| (*@$\land$@*)(*@$\neg$@*) TU.intersectsLine VW (*@$\to$@*)
    (*@$\triangle$@*) T:U:V).congruent ((*@$\triangle$@*) V:W:T)
\end{lstlisting}

One of the predictions generated by GPT-4v is given as follows:
\begin{lstlisting}[language=lean]
prediction : 
    (*@$\forall$@*) (U V W T : Point) (UV VT TW WT : Line), 
    formQuadrilateral U V W T UV VT TW WT (*@$\land$@*)
    |(T(*@--@*)U)| = |(V(*@--@*)W)| (*@$\land$@*) (*@$\neg$@*)(UV.intersectsLine WT) (*@$\to$@*) 
    ((*@$\triangle$@*) T:U:V).congruent ((*@$\triangle$@*) V:W:T)
\end{lstlisting}
In this case, the formalization does not quite line up with the text of the proposition; the statement assumes a quadrilateral rather than two triangles, and the premises refer to $UV$ and $WT$ being parallel rather than $TU$ and $VW$, despite still assuming $TU = VW$. Of course, the statement is still true, and in fact {\toolName} is able to prove they are equivalent; however, this is not quite a ``faithful'' formalization of the text.

\subsection{Unsatisfiable Statements}
{\toolName} did not identify any unsatisfiable predictions during our final rounds of experiments, but we did encounter a few such cases during preliminary experiments and testing; we include them here for the interested reader, as well as to emphasize the importance of rigorous semantic validation.

\smallsec{Proposition 11, GPT-4, 5-shot}
Consider the following:

\begin{lstlisting}[language=lean, numbers=none]
prediction : 
    (*@$\forall$@*)  (a b c : Point) (AB : Line), 
    c.onLine AB (*@$\to$@*)  
    (*@$\exists$@*)  (CF : Line),  c.onLine CF (*@$\land$@*)  (*@$\neg$@*) (AB.intersectsLine CF) (*@$\land$@*)  (*@$\angle$@*)  a:c:b = (*@$\rightangle$@*)  
\end{lstlisting}
{\toolName} correctly identifies this proposition as unsatisfiable, since, if $c$ is on $CF$, and $c$ is on $AB$, then obviously these lines must intersect.

\smallsec{Proposition 46, GPT-4, 5-shot}
Consider the following:

\begin{lstlisting}[language=lean, numbers=none]
prediction: 
(*@$\forall$@*)  (a b : Point) (AB : Line), distinctPointsOnLine a b AB (*@$\to$@*)  (*@$\exists$@*)  (c d : Point) (AC BD : Line), formQuadrilateral a d c b AB BD AC BD (*@$\land$@*) |(a(*@--@*) b)| = |(b(*@--@*) c)| (*@$\land$@*)  |(c(*@--@*) d)| = |(d(*@--@*) a)| (*@$\land$@*) (*@$\angle$@*)  a:b:c = (*@$\rightangle$@*)  (*@$\land$@*) (*@$\angle$@*)  b:c:d = (*@$\rightangle$@*)  (*@$\land$@*)  (*@$\angle$@*)  c:d:a = (*@$\rightangle$@*)  (*@$\land$@*)  (*@$\angle$@*)  d:a:b = (*@$\rightangle$@*) 
\end{lstlisting}

{\toolName} correctly identifies this proposition as unsatisfiable. The cause is that, in the application of ${\tt formQuadrilateral}$, the line $BD$ is given as an argument twice. However, ${\tt formQuadrilateral}$ requires that all its inputs are distinct objects.

\smallsec{Proposition 48, GPT-4, 5-shot}
Consider the following prediction, which is stating the converse direction of the Pythagorean theorem:

\begin{lstlisting}[language=lean,numbers=none]
prediction: 
    (*@$\forall$@*) (a b c : Point) (AB AC BC : Line), 
    formTriangle a b c AB AC BC (*@$\land$@*) 
    |(b(*@--@*)c)| * |(b(*@--@*)c)| = |(a(*@--@*)b)| * |(a(*@--@*)b)| + |(a(*@--@*)c)| * |(a(*@--@*)c)| (*@$\to$@*) 
    (*@$\angle$@*) a:b:c = (*@$\rightangle$@*)
\end{lstlisting}
Obviously, if $|BC|^2 = |AB|^2 + |AC|^2$, then $BC$ must be the hypotenuse of the triangle, so the right angle cannot be around point $B$. {\toolName} is able to identify this impossibility.

\subsection{The Approximate Logical Equivalence Procedure}
As shown in Fig.~\ref{fig:approxDraft}, the approximate equivalence checker attempts to quantify how ``semantically close'' one theroem statement is to another. The intention is to allow us to distinguish ``close'' formalizations from ``junk'' ones. We provide a more detailed description of how this procedure works. 

We assume we are given a ground truth formula $T_{gt}$ and a prediction $T_{gt}$ such that the two formulas agree on the quantity and types of their bound variables. The goal is to find a unification $\rho$ between bound variables in $T_{gt}$ and $T_{gt}$ with which we can prove the equivalence of the preconditions and postconditions of the two formulae. Obviously, the number of potential unifications is factorial in the number of variables, so we fix some upper bound $n$ on the number of unifications we consider, and rank candidates using a string similarity heuristic.

Having fixed a unification $\rho$ of variables in $T_{pred}$ with variables in $T_{gt}$, let $T_{pred}^\prime$ be the result of instantiating the variables in $T_{pred}$ with variables in $T_{gt}$ according to $\rho$. Assuming $T_{pred}^\prime$ and $T_{gt}$ are both well-formed statements in E, we can extract the sets $\{p_i\}$ and $\{q_i\}$ containing the pre/post-conditions of $T_{gt}$, respectively, and likewise the set $\{r_i\}$ and $\{s_i\}$ containing the pre/postconditions of $T_{pred}^\prime$. We then perform a four-step process: (1) Assuming $\bigwedge_i p_i$ try to prove each $r_j$ (2) Assuming $\bigwedge_i r_i$, try to prove each $p_j$ (3) Assuming $\bigwedge_i q_i$, try to prove each $s_j$ (4) Assuming $\bigwedge_i s_i$, try to prove each $q_j$.

If steps (1) and (2) are both successful (meaning that the preconditions of the formulae were proven equivalent), then the preconditions are included as additional assumptions for steps (3) and (4). For each step, {\toolName} records how many obligations were able to be proved. The results are then aggregated for each of the $n$ unifications of bound variables, which can then be evaluated by any number of metrics. For instance, Fig.~\ref{fig:approxDraft} compares the formalization instances using the ratio of clauses solved across steps (1)--(4).

\section{Proof Autoformalization}

\label{appendix:proofs}

\subsection{Quantitative Results}
As mentioned in the paper, only two proofs from \emph{Elements} were able to be formalized correctly out-of-the-box. For the remaining 43 proofs, we wished to quantify how ``good'' the proofs are by computing the degree of modification introduced during the repair process, as measured by the Levenshtein ratio. The results of this process for the (incomplete) proofs autoformalized by GPT-4 and GPT-4V are shown in Fig.~\ref{fig:bookProofHistText} and Fig.~\ref{fig:bookProofHistVision}, respectively. Obviously, this is not a perfect metric -- for instance, a tactic invoking ${\tt  proposition\_6}$ may only differ in a single character with one invoking ${\tt  proposition\_7}$. However, we find that these kinds of major conceptual errors, where the model chooses an entirely irrelevant theorem or rule, are quite rare. The majority of modifications necessary to repair proofs (as shown by the exemplars below) are simply rearranging arguments to tactics or modifying one variant of a theorem for another (e.g., ${\tt proposition\_9}$ versus the variant ${\tt proposition\_9'}$).

\begin{figure}[ht]
    \centering
    \includegraphics[width=0.8\textwidth]{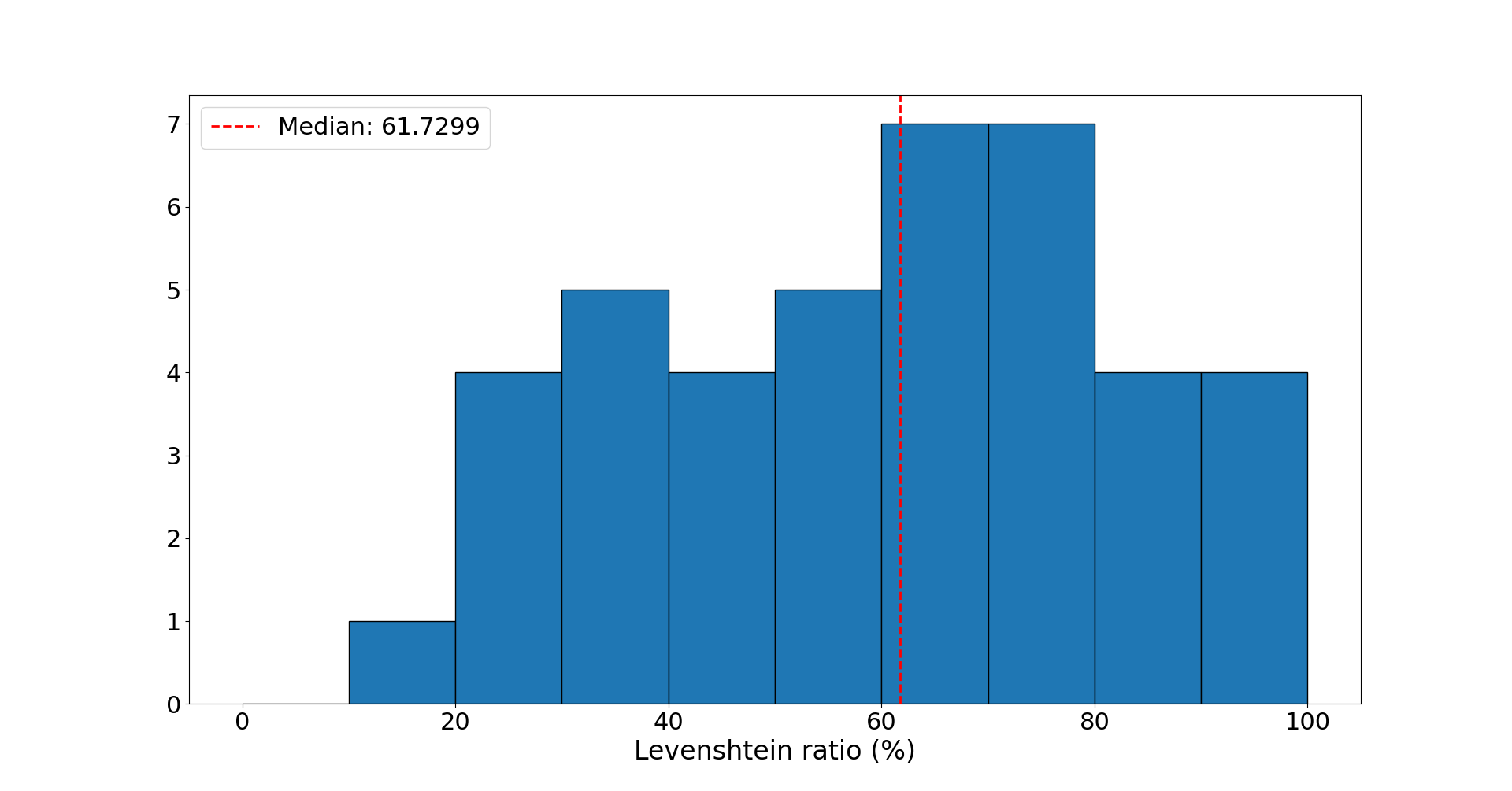}
    \caption{Levenshtein ratio between autoformalized and repaired \emph{Elements} proofs generated by GPT-4 (5-shot).}
    \label{fig:bookProofHistText}
\end{figure}
\begin{figure}[ht]
    \centering
    \includegraphics[width=0.8\textwidth]{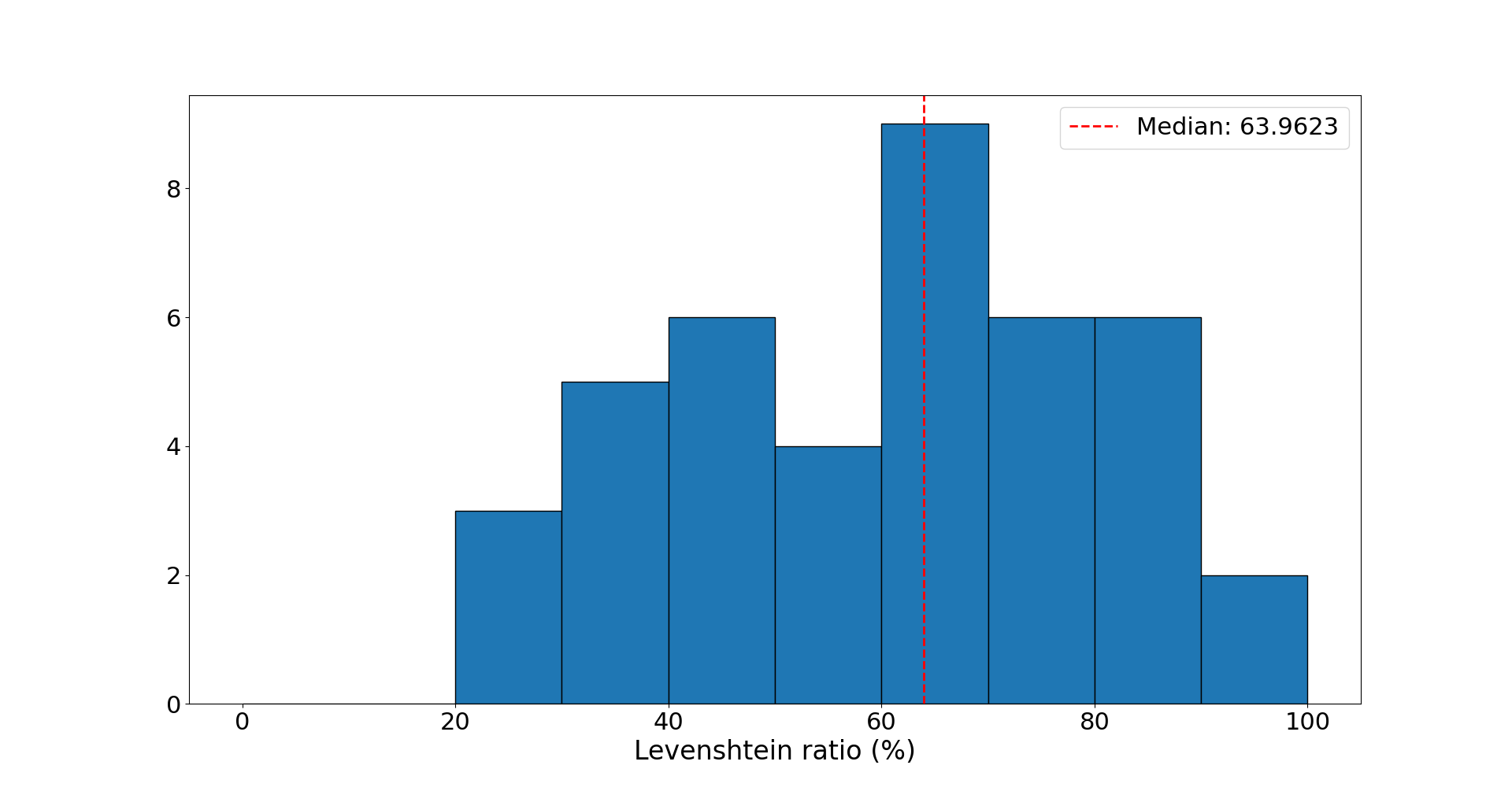}
    \caption{Levenshtein ratio between autoformalized and repaired \emph{Elements} proofs generated by GPT-4V (5-shot).}
    \label{fig:bookProofHistVision}
\end{figure}

\subsection{Correctly Formalized Proofs from \emph{Elements}}
The same two proofs -- Proposition 1 and Proposition 17 -- were correctly formalized out-of-the-box by GPT-4 and GPT4-V. Both of these are relatively simple proofs. Nevertheless, they serve as good exemplars of how {\dataname}'s proof language supports simple, elegant and faithful formalizations of the natural language proof.

\smallsec{Proposition 1, GPT-4V}
Natural language proof:
\begin{center}
    ``To construct an equilateral triangle on a given finite straight-line. Let $AB$ be the given finite straight-line. So it is required to construct an equilateral triangle on the straight-line $AB$. Let the circle $BCD$ with center $A$ and radius $AB$ have been drawn [Post.~3], and again let the circle $ACE$ with center $B$ and radius $BA$ have been drawn [Post.~3].And let the straight-lines $CA$ and $CB$ have been joined from the point $C$, where the circles cut one another, to the points $A$ and $B$ (respectively) [Post.~1].And since the point $A$ is the center of the circle $CDB$, $AC$ is equal to $AB$ [Def.~1.15]. Again,since the point $B$ is the center of the circle $CAE$, $BC$ is equal to $BA$ [Def.~1.15]. But $CA$ was also shown (to be) equal to $AB$. Thus, $CA$ and $CB$ are each equal to $AB$. But things equal to the same thing are also equal to one another [C.N.~1]. Thus, $CA$ is also equal to $CB$. Thus, the three (straight-lines) $CA$, $AB$, and $BC$ are equal to one another.Thus, the triangle $ABC$ is equilateral, and has been constructed on the given finite straight-line $AB$. (Which is) the very thing it was required to do.''
\end{center}
Autoformalized proof:
\begin{lstlisting}[language=lean, numbers=none]
theorem prop1_prediction : (*@$\forall$@*) (a b : Point) (AB : Line), distinctPointsOnLine a b AB (*@$\to$@*) (*@$\exists$@*) c : Point, |(c(*@--@*)a)| = |(a(*@--@*)b)| (*@$\land$@*) |(c(*@--@*)b)| = |(a(*@--@*)b)|:=
by  euclid_intros
    euclid_apply circle_from_points a b as BCD
    euclid_apply circle_from_points b a as ACE
    euclid_apply intersection_circles BCD ACE as c
    euclid_apply line_from_points c a as CA
    euclid_apply line_from_points c b as CB
    euclid_apply point_on_circle_if a b c BCD
    euclid_apply point_on_circle_if b a c ACE
    use c
    euclid_finish
\end{lstlisting}

\smallsec{Proposition 17, GPT-4V}

Natural language proof:

\begin{center}
    ``For any triangle,  (the sum of) two angles taken together in any (possible way) is less than two right-angles. Let $ABC$ be a triangle. I say that (the sum of) two angles of triangle $ABC$ taken together in any (possible way) is less than two right-angles.  For let $BC$ have been produced to $D$.  And since the angle $ACD$ is external to triangle $ABC$, it is greater than the internal and opposite angle $ABC$ [Prop.~1.16]. 
\end{center}

\begin{center}
Let $ACB$ have been added to both. Thus, the (sum of the angles) $ACD$ and $ACB$ is greater than the  (sum of the angles) $ABC$ and $BCA$. But, (the sum of) $ACD$ and $ACB$ is equal to two right-angles [Prop.~1.13]. Thus, (the sum of) $ABC$ and $BCA$ is less than two right-angles. Similarly, we can show that (the sum of) $BAC$ and $ACB$ is also less than two right-angles, and further (that the sum of) $CAB$ and $ABC$ (is less than two right-angles).  Thus, for any triangle,  (the sum of) two angles taken together in any (possible way) is less than two right-angles. (Which is) the very thing it was required to show.''
\end{center}

Autoformalized proof:
\begin{lstlisting}[language=lean,numbers=none]
theorem prop_17_prediction : (*@$\forall$@*) (a b c : Point) (AB BC AC : Line), formTriangle a b c AB BC AC (*@$\to$@*) ((*@$\angle$@*) a:b:c) + ((*@$\angle$@*) b:c:a) < (*@$\rightangle$@*) + (*@$\rightangle$@*):= 
by  euclid_intros
    euclid_apply extend_point BC b c as d
    euclid_apply proposition_16 a b c d AB BC AC
    euclid_apply proposition_13 a c b d AC BC
    euclid_finish
\end{lstlisting}

\subsection{Partiallly Correct Autoformalized Proofs from \emph{Elements}}
Here, we showcase some attempts at autoformalizing proofs from \emph{Elements} which, despite being incorrect, require only minor modifications to be repaired. These exemplars demonstrate that, in many cases, the steps required to repair the proof do not require creative insight. Moreover, we believe this process is quite amenable to automation.

\smallsec{Proposition 3, GPT-4}

Natural language proof: 
\begin{center}
    ``For two given unequal straight-lines, to cut off from the greater a straight-line equal to the lesser.  Let $AB$ and $C$ be the two given unequal straight-lines, of which let the greater be $AB$. So it is required to cut off a straight-line equal to the lesser $C$ from the greater $AB$.  Let the line $AD$, equal to the straight-line $C$, have been placed at point $A$ [Prop.~1.2]. And let the circle $DEF$ have been drawn with center $A$ and radius $AD$ [Post.~3].  And since  point $A$ is the center of  circle $DEF$, $AE$ is equal to $AD$ [Def.~1.15]. But, $C$ is also equal to $AD$. Thus, $AE$ and $C$ are each equal to $AD$. So $AE$ is also equal to $C$ [C.N.~1]. Thus, for two given unequal straight-lines, $AB$ and $C$, the (straight-line) $AE$, equal to the lesser $C$, has been cut off from the greater $AB$. (Which is) the very thing it was required to do."
\end{center}

Autoformalized Proof:
\begin{lstlisting}[language=lean,numbers=none]
    def prop3_prediction :  (*@$\forall$@*) (a b c0 c1 : Point) (AB C : Line), 
        distinctPointsOnLine a b AB (*@$\land$@*) distinctPointsOnLine c0 c0 C  (*@$\land$@*) a (*@$\neq$@*) c0 (*@$\land$@*) 
        |(a(*@--@*)b)| > |(c0(*@--@*)c1)| (*@$\to$@*) 
            (*@$\exists$@*) e, between a e b (*@$\land$@*) |(a(*@--@*)e)| = |(c0(*@--@*)c1)| := 
    by
        euclid_intros
        euclid_apply (proposition_2 a c0 c1 C) as d
        euclid_apply (circle_from_points a d) as DEF
        euclid_apply (intersection_circle_line DEF AB) as (e, f)
        euclid_apply (point_on_circle_onlyif a d e DEF)
        euclid_apply between_if a e b
        use e
        euclid_finish
\end{lstlisting}

To prove the precondition for ${\tt between\_if}$, we need the fact ${\tt between\; a \;e\; b}$, which is not provable from the preceding tactics. To repair this proof, we can use a stronger version of ${\tt intersection\_circle\_line}$ on Line 9.

Repaired Proof:
\begin{lstlisting}[language=lean,numbers=none]
def prop3_repaired :  (*@$\forall$@*) (a b c0 c1 : Point) (AB C : Line), 
    distinctPointsOnLine a b AB (*@$\land$@*) distinctPointsOnLine c0 c0 C  (*@$\land$@*) a (*@$\neq$@*) c0 (*@$\land$@*) 
    |(a(*@--@*)b)| > |(c0(*@--@*)c1)| (*@$\to$@*) 
        (*@$\exists$@*) e, between a e b (*@$\land$@*) |(a(*@--@*)e)| = |(c0(*@--@*)c1)| := 
by
    euclid_intros
    euclid_apply (proposition_2 a c0 c1 C) as d
    euclid_apply (circle_from_points a d) as DEF
    euclid_apply (intersection_circle_line_between_points DEF AB a b) as e
    euclid_apply (point_on_circle_onlyif a d e DEF)
    euclid_apply between_if a e b
    use e
    euclid_finish
\end{lstlisting}

\smallsec{Proposition 5, GPT-4}
Natural language proof: 
\begin{center}
    ``For isosceles triangles, the angles at the base are equal to one another, and if the equal sides are produced then the angles under the base will be equal to one another. Let $ABC$ be an isosceles triangle having the side $AB$ equal to the side $AC$, and let the straight-lines $BD$ and $CE$ have been produced in a straight-line with $AB$ and $AC$ (respectively) [Post.~2]. I say that the angle $ABC$ is equal to $ACB$, and (angle) $CBD$  to $BCE$.  For let the point $F$ have been taken at random on  $BD$, and let $AG$ have been cut off from the greater $AE$, equal to the lesser $AF$ [Prop.~1.3]. Also, let the straight-lines $FC$ and $GB$ have been joined [Post.~1].  In fact, since $AF$ is equal to $AG$, and $AB$ to $AC$, the two (straight-lines) $FA$, $AC$ are equal to the two (straight-lines) $GA$, $AB$, respectively. They also encompass a common angle, $FAG$. Thus, the base $FC$ is equal to the base $GB$, and the triangle $AFC$ will be equal to the triangle $AGB$, and the remaining angles subtended by the equal sides will be equal to the corresponding remaining angles [Prop.~1.4].  (That is) $ACF$ to $ABG$, and $AFC$ to $AGB$. And since the whole of $AF$ is equal to the whole of $AG$, within which $AB$ is equal to $AC$, the remainder $BF$ is thus equal to the remainder $CG$ [C.N.~3]. But $FC$ was also shown (to be) equal to $GB$. So the two (straight-lines) $BF$, $FC$ are equal to the two (straight-lines) $CG$, $GB$, respectively, and the angle $BFC$ (is) equal to the angle $CGB$, and the base $BC$ is common to them. Thus, the triangle $BFC$ will be equal to the triangle $CGB$, and the remaining angles subtended by the equal sides will be equal to the corresponding remaining angles [Prop.~1.4]. Thus, $FBC$ is equal to $GCB$, and $BCF$ to $CBG$. Therefore, since the whole angle $ABG$ was shown (to be) equal to the whole angle $ACF$, within which $CBG$ is equal to $BCF$, the remainder $ABC$ is thus equal to the remainder $ACB$ [C.N.~3]. And they are at the base of triangle $ABC$. And $FBC$ was also shown (to be) equal to $GCB$. And they are under the base.  Thus, for isosceles triangles, the angles at the base are equal to one another, and if the equal sides are produced then the angles under the base will be equal to one another. (Which is) the very thing it was required to show.''
\end{center}

Autoformalized Proof:
\begin{lstlisting}[language=lean, numbers = none]
def prop5_prediction : (*@$\forall$@*) (a b c d e : Point) (AB BC AC : Line), formTriangle a b c AB BC AC (*@$\land$@*) |(a(*@--@*)b)| = |(a(*@--@*)c)| (*@$\land$@*) between a b d (*@$\land$@*) between a c e (*@$\to$@*) (*@$\angle$@*)a:b:c = (*@$\angle$@*)a:c:b (*@$\land$@*) (*@$\angle$@*)c:b:d = (*@$\angle$@*)b:c:e :=
by  euclid_intros
    euclid_apply (exists_point_between_points_on_line AB b d) as f
    euclid_apply (proposition_3 e a a f AC AB) as g
    euclid_apply (line_from_points f c) as FC
    euclid_apply (line_from_points g b) as GB
    euclid_apply proposition_4 f a c g a b FC AC GB AB AC AB
    euclid_apply (between_if f b a)
    euclid_apply (between_if g c a)
    euclid_apply (proposition_4 b f c c g b AB FC BC GB BC AB)
    use f, g, FC, GB
    euclid_finish
\end{lstlisting}
In order to prove the preconditions for ${\tt proposition\_3}$, we need to know $|ba| < |ce|$, which can be obtained via  a stronger version of  ${\tt exists\_point\_between\_points\_on\_line}$ on Line 2. We also need to rearrange some of the arguments on in the invocations of ${\tt proposition\_3}$ and ${\tt proposition\_4}$. Finally, we can remove the invocation of tactic ${\tt use}$ entirely, since it is inapplicable in this proof context.

Repaired Proof:
\begin{lstlisting}[language=lean, numbers=none]
def prop5_repaired : (*@$\forall$@*) (a b c d e : Point) (AB BC AC : Line), formTriangle a b c AB BC AC (*@$\land$@*) |(a(*@--@*)b)| = |(a(*@--@*)c)| (*@$\land$@*) between a b d (*@$\land$@*) between a c e (*@$\to$@*) (*@$\angle$@*)a:b:c = (*@$\angle$@*)a:c:b (*@$\land$@*) (*@$\angle$@*)c:b:d = (*@$\angle$@*)b:c:e  :=
by  euclid_intros
    euclid_apply (point_between_points_shorter_than AB b d (c(*@--@*)e)) as f
    euclid_apply (proposition_3 a e f a AC AB) as g
    euclid_apply (line_from_points f c) as FC
    euclid_apply (line_from_points b g) as GB
    euclid_apply proposition_4 a f c a g b AB FC AC AC GB AB
    euclid_apply (between_if f b a)
    euclid_apply (between_if g c a)
    euclid_apply (proposition_4 f b c g c b AB BC FC AC BC GB)
    euclid_finish
\end{lstlisting}

\smallsec{Proposition 7, GPT-4}

Natural language proof: 
\begin{center}
    On the same straight-line, two other straight-lines  equal, respectively, to  two (given) straight-lines (which meet) cannot be constructed (meeting) at  a different point on the same side (of the straight-line), but having the same ends as the given straight-lines.      For, if possible, let the two straight-lines $AC$, $CB$, equal to two other straight-lines $AD$, $DB$, respectively, have been constructed on the same straight-line $AB$, meeting at different points, $C$ and $D$, on the same side (of $AB$), and having the same ends (on $AB$). So $CA$ is equal to $DA$, having the same end $A$ as it, and $CB$ is equal to $DB$, having the same end $B$ as it. And let $CD$ have been joined [Post.~1].  Therefore, since $AC$ is equal to $AD$,  the angle $ACD$ is also equal to angle $ADC$ [Prop.~1.5]. Thus, $ADC$ (is) greater than $DCB$ [C.N.~5]. Thus, $CDB$ is much greater than $DCB$ [C.N.~5]. Again, since  $CB$ is equal to $DB$, the angle $CDB$ is also equal to angle $DCB$ [Prop.~1.5]. But it was shown that the former (angle) is also much greater (than the latter). The very thing is impossible.  Thus, on the same straight-line, two other straight-lines equal, respectively, to   two (given) straight-lines  (which meet) cannot be constructed (meeting) at a different point on the same side (of the straight-line), but having the same ends as the given straight-lines. (Which is) the very thing it was required to show.
\end{center}

Autoformalized Proof:
\begin{lstlisting}[language=lean,numbers=none]
def prop7_prediction : (*@$\forall$@*) (a b c d : Point) (AB AC CB AD DB CD : Line), distinctPointsOnLine a b AB (*@$\land$@*) distinctPointsOnLine a c AC (*@$\land$@*) distinctPointsOnLine c b CB (*@$\land$@*) distinctPointsOnLine a d AD (*@$\land$@*) distinctPointsOnLine d b DB (*@$\land$@*) Point.sameSide c d AB (*@$\land$@*) distinctPointsOnLine c d CD (*@$\land$@*) Point.sameSide a b CD (*@$\land$@*) Point.sameSide d b AC (*@$\land$@*) |(a(*@--@*)c)| = |(a(*@--@*)d)| (*@$\land$@*) |(c(*@--@*)b)| = |(d(*@--@*)b)| (*@$\to$@*) False :=
by  euclid_intros
    euclid_apply (line_from_points c d) as CD
    euclid_apply proposition_5 c a d AC AD CD
    euclid_apply proposition_5 c b d CB DB CD
    by_cases ((*@$\angle$@*)c:a:d) > ((*@$\angle$@*) d:c:b)
    . euclid_apply proposition_5' c a d b AC AD CD AB
      euclid_finish
    . by_cases ((*@$\angle$@*) c:b:d) > ((*@$\angle$@*) d:c:b)
      . euclid_apply proposition_5' c b d a CB DB CD AB
        euclid_finish
      . euclid_finish
\end{lstlisting}
The rule ${\tt proposition\_5}$ takes five points and three lines as argument, whereas the variant ${\tt proposition\_5'}$ takes three of each. It is easiest to fix the fourth and fifth tactics by instead invoking the latter theorem (and tweaking the arguments). The ${\tt euclid\_apply}$ within the case analysis also need repairing, since they are using too many arguments. These tactics are actually redundant, since ${\tt euclid\_finish}$ can complete the proof. Nevertheless, simply ommitting the last point and line from each tactic is sufficient to make the proof go through.

Repaired Proof:
\begin{lstlisting}[language=lean,numbers=none]
def prop7_repaired : (*@$\forall$@*) (a b c d : Point) (AB AC CB AD DB CD : Line), distinctPointsOnLine a b AB (*@$\land$@*) distinctPointsOnLine a c AC (*@$\land$@*) distinctPointsOnLine c b CB (*@$\land$@*) distinctPointsOnLine a d AD (*@$\land$@*) distinctPointsOnLine d b DB (*@$\land$@*) Point.sameSide c d AB (*@$\land$@*) distinctPointsOnLine c d CD (*@$\land$@*) Point.sameSide a b CD (*@$\land$@*) Point.sameSide d b AC (*@$\land$@*) |(a(*@--@*)c)| = |(a(*@--@*)d)| (*@$\land$@*) |(c(*@--@*)b)| = |(d(*@--@*)b)| (*@$\to$@*) False := 
by 
    euclid_intros
    euclid_apply line_from_points c d as CD'
    euclid_apply proposition_5' a c d AC CD AD
    euclid_apply proposition_5' b c d CB CD DB
    by_cases ((*@$\angle$@*)  c:a:d) > ((*@$\angle$@*)  d:c:b)
    . euclid_apply proposition_5' c a d AC AD CD
      euclid_finish
    . by_cases ((*@$\angle$@*)  c:b:d) > ((*@$\angle$@*)  d:c:b)
        . euclid_apply proposition_5' c b d CB DB CD
          euclid_finish
        . euclid_finish
\end{lstlisting}

\smallsec{Proposition 18, GPT-4V}

Natural language proof: 
\begin{center}
    ``In any triangle, the greater side subtends the greater angle. For let $ABC$ be a triangle having side $AC$ greater than $AB$. I say that angle $ABC$ is also greater than $BCA$.  For since $AC$ is greater than $AB$, let $AD$ be made equal to $AB$  [Prop.~1.3], and let $BD$ have been joined.  And since angle $ADB$ is external to triangle $BCD$, it is greater than the internal and opposite (angle) $DCB$ [Prop.~1.16]. But $ADB$ (is) equal to $ABD$, since side $AB$ is also equal to side $AD$ [Prop.~1.5]. Thus, $ABD$ is also greater than $ACB$. Thus, $ABC$ is much greater than  $ACB$.  Thus, in any triangle, the greater side subtends the greater angle. (Which is) the very thing  it was required to show.''
\end{center}

Autoformalized Proof:
\begin{lstlisting}[language=lean,numbers=none]
def prop18_prediction : (*@$\forall$@*) (a b c : Point) (AB BC AC : Line), formTriangle a b c AB BC AC (*@$\land$@*) (|(a (*@--@*)c)| > |(a(*@--@*)b)|) (*@$\to$@*) ((*@$\angle$@*) a:b:c > (*@$\angle$@*) b:c:a):=
by  euclid_intros
    euclid_apply proposition_3 a c a b AC AB as d 
    euclid_apply line_from_points b d as BD
    euclid_apply proposition_16 a b d c AB BD AC 
    euclid_apply proposition_5 a b d AB BD as H1
    euclid_apply proposition_6 a b d AB BD AC
    euclid_finish
\end{lstlisting}
The second tactic does not provide the right number of arguments to ${\tt proposition\_3}$, but it does for ${\tt proposition\_3'}$. The preconditions of the invocation of ${\tt proposition\_16}$ are not provable from the preceding steps, but we can permute points and $a,b,c$ to recover a valid tactic. The invocation of ${\tt proposition\_5}$ has the same problem as  ${\tt proposition\_3}$ , and is replaced with ${\tt proposition\_5}^\prime$. Also, the keyword ``${\tt as}$'' can only be used for constructive theorems, which ${\tt proposition\_5}$ is not. We can simply drop ``${\tt as\;H1}$'' and the proof goes through. 

Repaired Proof:
\begin{lstlisting}[language=lean,numbers=none]
def prop18_repaired : (*@$\forall$@*) (a b c : Point) (AB BC AC : Line), formTriangle a b c AB BC AC (*@$\land$@*) (|(a (*@--@*)c)| > |(a(*@--@*)b)|) (*@$\to$@*) (((*@$\angle$@*) a:b:c) > (*@$\angle$@*) b:c:a):=
    euclid_intros
    euclid_apply proposition_3' a c a b AC AB as d
    euclid_apply line_from_points b d as BD
    euclid_apply proposition_16 b c d a BC AC BD
    euclid_apply proposition_5' a b d AB BD AC
    euclid_apply proposition_6 a b d AB BD AC
    euclid_finish
\end{lstlisting}

\smallsec{Proposition 34, GPT-4V}

Natural language proof: 
\begin{center}
    ``In parallelogrammic figures the opposite sides and angles are equal to one another, and a diagonal cuts them in half.  Let $ACDB$ be a parallelogrammic figure, and $BC$ its diagonal. I say that for parallelogram $ACDB$, the opposite sides and angles are equal to one another, and the diagonal $BC$ cuts it in half.  For since $AB$ is parallel to $CD$, and the straight-line $BC$ has fallen across  them, the alternate angles $ABC$ and $BCD$ are equal to one another [Prop.~1.29].  Again, since $AC$ is parallel to $BD$, and $BC$ has fallen across them, the alternate angles $ACB$ and $CBD$ are equal to one another [Prop.~1.29].  So $ABC$ and $BCD$ are two triangles having the two angles $ABC$ and $BCA$ equal to the two (angles) $BCD$ and $CBD$, respectively, and one side equal to one side---the (one) by the equal angles and common to them, (namely) $BC$. Thus, they will also  have the remaining sides  equal to the corresponding remaining (sides), and the remaining angle (equal) to the remaining angle [Prop.~1.26]. Thus, side $AB$ is equal to $CD$, and $AC$ to $BD$. Furthermore, angle $BAC$ is  equal to $CDB$. And since angle $ABC$ is equal to $BCD$, and $CBD$ to $ACB$, the whole (angle) $ABD$ is thus equal to the whole (angle) $ACD$. And  $BAC$ was also shown  (to be) equal to $CDB$.  Thus, in parallelogrammic figures the opposite sides and angles are equal to one another.  And, I also say that a diagonal cuts them in half. For since $AB$ is equal to $CD$, and $BC$ (is) common, the two (straight-lines) $AB$, $BC$ are equal to the two (straight-lines) $DC$, $CB$, respectively. And angle $ABC$ is equal to angle $BCD$. Thus, the base $AC$ (is) also equal to $DB$, and triangle $ABC$ is equal to triangle $BCD$ [Prop.~1.4].  Thus, the diagonal $BC$ cuts the parallelogram $ACDB$ in half. (Which is) the very thing it was required to show.''
\end{center}

Autoformalized Proof:
\begin{lstlisting}[language=lean,numbers=none]
def prop34_prediction : (*@$\forall$@*) (a b c d : Point) (AB CD AC BD BC : Line), formParallelogram a b c d AB CD AC BD (*@$\land$@*) distinctPointsOnLine b c BC (*@$\to$@*) 
    |(a(*@--@*)b)| = |(c(*@--@*)d)| (*@$\land$@*) |(a(*@--@*)c)| = |(b(*@--@*)d)| (*@$\land$@*) 
    (*@$\angle$@*) a:b:d = (*@$\angle$@*) a:c:d (*@$\land$@*) (*@$\angle$@*) b:a:c = (*@$\angle$@*) c:d:b (*@$\land$@*) 
    Triangle.area (*@$\triangle$@*) a:b:c = Triangle.area (*@$\triangle$@*) d:c:b := 
by  euclid_intros
    euclid_apply proposition_29 a b c d AB CD AC BD
    euclid_apply proposition_29 c d a b CD AB BD AC
    euclid_apply proposition_26 a b c d a c b d AB BC AC BD AB CD
    euclid_apply proposition_4 a b c d c b AB BC AC BD CD BC
    euclid_finish
\end{lstlisting}

Similar to previous cases, the first invocation of ${\tt proposition\_29}$ on  is not well-formed, since ${\tt proposition\_29}$ requires more arguments than the model has supplied. However, the variant ${\tt proposition\_29'''}$ accepts the number and quantity of variables provided, so we can swap the rule and permute its arguments to make the tactic go through. The next invocation is repaired in the same fashion. Too few arguments are given to ${\tt proposition\_26}$5, but there are no variants of this theorem, so we must infer which arguments can be supplied from the proof context. The preconditions of the tactic using ${\tt proposition\_4}$ cannot be proved as-is. As a matter of fact, this tactic is unnecessary, since, the proof can be completed already by ${\tt euclid\_finish}$. However, if we want to repair it, we can simply permute a few of its arguments.

Repaired Proof:
\begin{lstlisting}[language=lean, numbers=none]
def prop34_prediction :  (*@$\forall$@*) (a b c d : Point) (AB CD AC BD BC : Line), formParallelogram a b c d AB CD AC BD (*@$\land$@*) distinctPointsOnLine b c BC (*@$\to$@*) 
    |(a(*@--@*)b)| = |(c(*@--@*)d)| (*@$\land$@*) |(a(*@--@*)c)| = |(b(*@--@*)d)| (*@$\land$@*) 
    (*@$\angle$@*) a:b:d = (*@$\angle$@*) a:c:d (*@$\land$@*) (*@$\angle$@*) b:a:c = (*@$\angle$@*) c:d:b (*@$\land$@*) 
    Triangle.area (*@$\triangle$@*) a:b:c = Triangle.area (*@$\triangle$@*) d:c:b := 
by 
    euclid_intros
    euclid_apply proposition_29''' a d b c AB CD BC
    euclid_apply proposition_29''' a d c b AC BD BC
    euclid_apply proposition_26 a b c d c b AB BC AC CD BC BD
    euclid_apply (proposition_4 b a c c d b AB AC BC CD BD BC)
    euclid_finish
\end{lstlisting}

\subsection{Autoformalized Proofs from UniGeo}
Here, we showcase two examples of correctly autoformalized proofs from the UniGeo dataset. 
\smallsec{Congruent/Thm07, GPT-4V, 5-shot}
The diagrammatic input for this theorem is shown in Fig.~\ref{fig:uniCong07}. The theorem asserts that if $|WX|$ = $|ZY|$ and $WX$ is parallel to $ZY$, then triangle $WYZ$ is congruent to triangle $YWX$. The UniGeo proof makes the following inferences:
\begin{align*}
 & WX \parallel
 XY
 \land |WX|=|YZ|\\
 \therefore\;\;& \angle WYZ = \angle XWY \\
 & |WY| = |WY|\\
 \therefore & \triangle WYZ \cong \triangle YWX
\end{align*}
\begin{figure}[ht]
    \centering
    \includegraphics[scale=0.6]{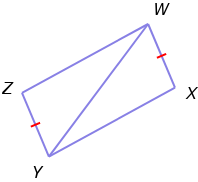}
    \caption{Diagrammatic input for UniGeo/Congruent/Thm07}
    \label{fig:uniCong07}
\end{figure}
GPT-4V correctly translates this semiformal proof to {\dataname} as follows:

\begin{lstlisting}[language=lean, numbers=none]
theorem prediction_thm07: 
     (*@$\forall$@*) (W X Y Z : Point) (WX XY WY YZ WZ : Line), 
    formTriangle W X Y WX XY WY  (*@$\land$@*) 
    formTriangle W Y Z WY YZ WZ  (*@$\land$@*) 
    X.opposingSides Z WY  (*@$\land$@*) |(W (*@$--$@*)X)| = |(Y (*@$--$@*)Z)|  (*@$\land$@*) 
     (*@$\neg$@*) WX.intersectsLine YZ  (*@$\to$@*) ( (*@$\triangle$@*) W:Y:Z).congruent ( (*@$\triangle$@*) Y:W:X):=
by
    euclid_intros
    have :  (*@$\angle$@*)W:Y:Z =  (*@$\angle$@*)X:W:Y := by
        euclid_apply proposition_29''' X Z W Y WX YZ WY
        euclid_finish
    have : |(W (*@$--$@*)Y)| = |(W (*@$--$@*)Y)| := by euclid_finish
    have :  ((*@$\triangle$@*)W:Y:Z).congruent  (*@$\triangle$@*)Y:W:X := by euclid_finish
    euclid_finish
\end{lstlisting}
\smallsec{Parallel/Thm06, GPT-4V, 5-shot}
The diagrammatic input for this proposition is shown in Fig.~\ref{fig:uniPar06}.  The theorem asserts that if $\angle S T W$ and $\angle T W V$ are supplementary, then $SU$ is parallel to $VX$. The UniGeo proof makes the following inferences:
\begin{align*}
& \angle STW {\textrm{ and } } \angle TWV {\textrm{ are supplementary } }\\
\therefore \;& \angle STW+ \angle TWV=180^\circ\\
\therefore \;&\angle RTS+ \angle STW=180^\circ\\
\therefore \;&\angle STW+ \angle TWV= \angle RTS+ \angle STW\\
\therefore \;& \angle TWV= \angle RTS\\
\therefore \;& SU\parallel VX
\end{align*}

\begin{figure}[ht]
    \centering
    \includegraphics[scale=0.6]{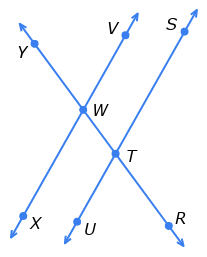}
    \caption{Diagrammatic input for UniGeo/Parallel/Thm06}
    \label{fig:uniPar06}
\end{figure}

GPT-4V correctly translates this semiformal proof to {\dataname} as follows:
\begin{lstlisting}[language=lean, numbers=none]
theorem prediction_thm06: 
     (*@$\forall$@*) (S U V X R Y T W : Point) (SU VX RY : Line), distinctPointsOnLine S U SU  (*@$\land$@*) 
    distinctPointsOnLine V X VX  (*@$\land$@*) 
    distinctPointsOnLine R Y RY  (*@$\land$@*) 
    twoLinesIntersectAtPoint SU RY T  (*@$\land$@*) 
    between S T U  (*@$\land$@*) between R T W  (*@$\land$@*) 
    twoLinesIntersectAtPoint VX RY W  (*@$\land$@*) 
    between V W X  (*@$\land$@*) between T W Y  (*@$\land$@*) 
    V.sameSide S RY  (*@$\land$@*) U.sameSide X RY  (*@$\land$@*) 
     (*@$\angle$@*) S:T:W +  (*@$\angle$@*) T:W:V =  (*@$\rightangle$@*) +  (*@$\rightangle$@*)  (*@$\to$@*)  (*@$\neg$@*) VX.intersectsLine SU:= 
by
  euclid_intros
  have :  (*@$\angle$@*) R:T:S +  (*@$\angle$@*) S:T:W  = (*@$\rightangle$@*)+(*@$\rightangle$@*):=
  by
    euclid_apply proposition_13 S T R W SU RY
    euclid_finish
  have :  (*@$\angle$@*)S:T:W + (*@$\angle$@*)T:W:V = (*@$\angle$@*) R:T:S +  (*@$\angle$@*) S:T:W := by euclid_finish
  have :  (*@$\angle$@*)T:W:V =  (*@$\angle$@*) R:T:S  := by euclid_finish
  have :  (*@$\neg$@*) VX.intersectsLine SU := by
    euclid_apply proposition_15 R W S U T RY SU
    euclid_apply proposition_27' U V T W SU VX RY
    euclid_finish
  euclid_finish
\end{lstlisting}

\end{document}